\documentclass[lettersize,journal]{IEEEtran}
\IEEEoverridecommandlockouts
\usepackage{amsmath,graphicx}
\usepackage[utf8]{inputenc}
\usepackage{orcidlink}
\usepackage{amssymb}
\usepackage{stackengine}

\usepackage[T1]{fontenc}
\usepackage{gensymb}
\usepackage{multirow}

 \usepackage{subcaption}
\usepackage{tikz}
\usepackage{tcolorbox}
\usepackage{breqn}
\definecolor{arrowblue}{rgb}{0,0,0.8}
\usepackage{acronym}
\usepackage{array}
\usepackage{algorithm}
\usepackage{algpseudocode}
\usepackage{url}

\usepackage{cite} 
\usepackage[export]{adjustbox}
\acrodef{GSF}{glare spread function}
\acrodef{HDR}{high dynamic range}
\acrodef{CV}{computer vision}
\acrodef{TF}{transfer function}
\usepackage{hyperref}
\begin{document}
\title{How to deal with glare for improved perception of Autonomous Vehicles }
\author{
   \hspace{-0.5cm}Muhammad Z. Alam\orcidlink{https://orcid.org/0000-0002-0114-8248}, \thanks{Muhammad Z. Alam, Department of Computer science, Brandon University, (e-mail: alamz@brandonu.ca)}
  \and
  Zeeshan Kaleem\orcidlink{https://orcid.org/0000-0002-7163-0443}, Senior Member, IEEE, \thanks{Zeeshan Kaleem, Department of Electrical and Computer Engineering, COMSATS University,  (e-mail: zeeshankaleem@ciitwah.edu.pk)}
    \and
      Sousso Kelouwani\orcidlink{https://orcid.org/0000-0002-1163-9147}, Senior Member, IEEE,\thanks{Sousso Kelouwani, Department of Mechanical Engineering, UQTR,  (e-mail: Sousso.Kelouwani@uqtr.ca)}   
}

\maketitle

\begin{abstract}
Vision sensors are versatile and can capture a wide range of visual cues, such as color, texture, shape, and depth. This versatility, along with the relatively inexpensive availability of machine vision cameras,  played an important role in adopting vision-based environment perception systems in autonomous vehicles (AVs). However, vision-based perception systems can be easily affected by glare in the presence of a bright source of light, such as the sun or the headlights of the oncoming vehicle at night or simply by light reflecting off snow or ice-covered surfaces; scenarios encountered frequently during driving. In this paper, we investigate various glare reduction techniques, including the proposed saturated pixel-aware glare reduction technique for improved performance of the computer vision (CV) tasks employed by the perception layer of AVs. We evaluate these glare reduction methods based on various performance metrics of the CV algorithms used by the perception layer. Specifically, we considered object detection, object recognition, object tracking, depth estimation, and lane detection which are crucial for autonomous driving. The experimental findings validate the efficacy of the proposed glare reduction approach, showcasing enhanced performance across diverse perception tasks and remarkable resilience against varying levels of glare.
\end{abstract}
\section{Introduction}
An accurate and robust environmental perception system is crucial for the advancement of intelligent transportation, especially in the case of self-driving vehicles \cite{TVT}. Meeting the requirements of level 5 autonomy, as specified in the J3016 \cite{hopkins2021talking} international standard, entails the ability to operate out of the so-called operational design domain. Instead of
a carefully managed (usually urban) environment with lots of dedicated infrastructure. Autonomous vehicles (AVs) should be able to operate in uncontrollable environments, including challenging weather, glare, haze, and fog causing illumination variation, poorly marked roads, and unpredictable road users \cite{10115323}.\\
The perception layer in the AV software stack is responsible for timely perceiving the changes happening in the vehicle's environment, through various computer vision (CV) tasks, such as object detection and recognition, depth estimation, lane detection, and more. 
In recent times, the vision-based perception layer in AVs has gained immense popularity \cite{TVT1}. Several automotive companies, such as Tesla, BMW, and Mobileye have created their vision-based perception systems. \\
This trend can be attributed to several factors, including the fact that road markings and signs are human-centric, the decreasing costs of machine vision cameras, and the rapid progress in deep neural networks used in vision-related tasks. However, the phenomenon of intense glare, encountered frequently while driving through the tunnels, bright sunlight, oncoming headlights, snow or fog, and the light reflected from the water droplets in the air, significantly degrades the performance of vision-based detection systems.\\
Typically, due to cost and, more importantly, size constraints, machine vision cameras deployed on AVs use simpler lens systems that lack anti-glare coatings, introducing much stronger glare than the high-end mirrorless cameras \cite{AlamG}. Additionally, state-of-the-art vision-based AVs perception algorithms that use deep neural networks are pre-trained on image data captured under normal conditions and, therefore, suffer loss in performance when encountering images with significant glare.\\
Glare that is caused by scattering (reflecting or refracting) of the light from the surface of the lens is a major limitation of any camera system as it can substantially reduce the dynamic range of captured images or when a portion of an image is overexposed due to glare, it can cause the loss of important details and features. It can produce artifacts such as halos, flares, and ghosting, which can distort or obscure the image and may also cause color distortion, resulting in inaccurate or unrealistic colors in the image.\\
In an ideal scenario, a point light source that is in focus should only illuminate a single pixel. However, in reality, due to the scattering and reflections of light inside the camera lens and body, the light also reaches other pixels on the sensor. To reduce the amount of additional light that reaches the sensor, several glare reduction techniques have been proposed. The majority of the current approaches for reducing glare can be classified into three primary categories: enhancing optics, computational or post-processing techniques, and techniques that utilize occlusion masks. All of the techniques from these different categories either involve expensive hardware modification or complex computational models designed for a specific imaging system. Even the more advanced deep learning-based approaches are typically trained on a dataset captured with a single source and would require extensively large datasets for fine-tuning to another imaging system. \\
This specificity to a single camera model limits their application in the AV's, because advanced imaging components e.g. telephoto lens, designed for high-speed and distant object detection struggle with close-range glare. This limitation arises because telephoto lenses have a narrower field of view and are optimized for distant focus, reducing effectiveness against glare from nearby sources. In contrast, cameras, featuring high-resolution and light-sensitive technology, face difficulties with distant glare due to their emphasis on detailed recording in urban settings. These cameras are optimized for clarity and detail in close to medium-range scenarios. Consequently, have limited focal range or dynamic range to effectively handle glare from distant light sources, which can be more diffused and less distinct, leading to reduced visibility and glare management at longer distances.\\
In this paper, we introduce a novel glare mitigation approach based on a joint Glare Spread Function (GSF). This method employs high dynamic range image of a point light source and refines a radially symmetric function in the joint GSF estimation process. This refinement is tailored to assimilate the glare characteristics of new imaging systems while preserving efficacy for previously utilized cameras. The focus of this technique is on adaptability and robustness, making it suitable for the diverse and challenging conditions encountered by AVs.\\
The proposed method aims to recover a glare-free image by using deconvolution of the observed image with a pre-estimated GSF. However, deconvolution-based methods require knowledge of the pixel intensities of the glare source to perform well. In practice, some pixels may get saturated, such as the headlights of a vehicle at night  introducing non-linearities and information loss in an image, violating the assumptions underlying the linear deconvolution process hence limiting its application to glare mitigation in the presence of saturation. This issue is addressed by accurately estimating the saturated regions in the input image using the dark channel prior \cite{he2010single}. Replacing the saturated regions by the estimated radiance, thereby restoring the assumption of linearity. Following this, deconvolution is performed to effectively restore the degraded input.\\
 \textit{The effectiveness of the proposed method is notably improved when the source of glare is within the camera's field of view (FoV). It operates under the assumption that sources outside the FoV can be adequately managed with simpler solutions, such as the use of a camera hood, to limit image degradation.} The proposed technique is briefly presented in \cite{AlamG}, lacks the joint GSF estimation, restricting its application to human perception, a task entirely different from AVs perception, which involves diverse cameras.
The major contributions of this paper are summarized as follows:
\begin{itemize}
    \item This paper proposes a glare reduction method that leverages joint GSF estimation in overcoming two prevalent issues in deconvolution-based glare reduction techniques: saturation handling and GSF specificity. 
   \item The paper presents the broadest range of perception tasks studied in the context of glare reduction research to date.
    \item  It contributes a thorough evaluation of various glare reduction techniques, based directly on autonomous vehicle perception algorithms performance metric.
\end{itemize}
The paper is organized as follows. In Section II, we present related work on glare reduction strategies. We explain the overall testing framework adopted in this paper in section III. The proposed joint GSF estimation framework is presented in section IV. The proposed glare reduction method is described in section V. In section VI we present the image encoding methods tested in this paper. AV perception layer algorithms along with the evaluation metrics are discussed in section VII. Experimental results and a detailed discussion of the results are provided in section VIII. Finally, we conclude the paper in section IX.
\section{Related Work}
Glare poses a well-recognized challenge in high dynamic range imaging. Previous research on mitigating glare can be broadly classified into three main categories: advancements in optics, computational or post-processing techniques, and techniques involving occlusion masks. The subsequent paragraphs provide a review of each of these groups. \\
To address glare in optics, high-quality anti-glare lens coatings are used to significantly reduce reflectivity. However, internal camera components like plastic interiors and painted surfaces can still contribute to glare. In \cite{Boynton}, a novel method using a liquid-filled charge-coupled device (CCD) is proposed to minimize stray light. This camera, called the simulated-eye-design (SED), features a glass lens with a liquid layer extending to the CCD, reducing reflective interfaces. Another approach in \cite{Hara} uses an electronically controlled optical shutter array in front of the camera lens. By selectively activating elements of this array, glare sources can be effectively blocked. While effective, these methods involve camera modifications and may not be universally applicable. While these approaches are effective, they necessitate certain modifications to the camera or optics and therefore may not be readily applicable without adaptations.\\
In previous studies, deconvolution has been commonly used as a post-processing technique to remove glare. One method proposed by Reinhard et al. in \cite{REINHARD} is a blind deconvolution-based approach that estimates the glare spread function (GSF) by fitting a radially-symmetric polynomial to the decrease of light around bright pixels. In \cite{Talvala}, the glare spread function (GSF) is characterized by a delta function along with a low-valued, low-frequency component that represents unwanted glare. The process involves removing glare from an image by deconvolving the glared image with the modeled GSF. However, it is important to note that deconvolution-based methods can only be effective when the brightest objects in the scene, which contribute to glare the most, can be captured without being saturated. Unfortunately, even high dynamic range (HDR) cameras may struggle to capture the full dynamic range of the scene in many scenarios, which limits the performance of deconvolution-based glare reduction methods.\\
In \cite{Nayar}, Nayar et al. propose an alternative method for removing veiling glare. Their approach involves separating the radiance of the scene into two components: direct radiance, resulting from the direct illumination of a point by a light source, and global radiance, which arises from the illumination of a point by other points in the scene. This separation is achieved using either structured illumination or an occlusion mask. Another variant of this method is presented in \cite{Talvala} where the global component of light in the image is limited through the insertion of high-frequency occlusion mask between the camera and the scene.  Although glare appears as an additive low-frequency bias in 2D, Raskar et al. \cite{Raskar} showed that a significant part of the glare is high-frequency noise in 4D ray-space. To remove that high-frequency part of glare, a high-frequency mask is introduced near the camera sensor to act like a sieve that separates spurious rays in ray-space. Again, these methods require either some additional hardware or modification of the camera design. \\
Similar to glare, atmospheric haze is a phenomenon that diminishes the quality of an image by introducing a reflected component of light to the overall radiance of a scene. In recent times, several highly effective de-hazing methods based on deep learning have been proposed \cite{Cai,Yang, Zhang, Chen}. Given that haze removal aims to reduce undesirable additive light components, these techniques may also be applicable to addressing the issue of glare reduction. For example, in \cite{Cai} a convolution neural network (CNN) based, end-to-end haze removal system, called the DehazeNet is proposed. DehazeNet takes a hazy image as an input and outputs the medium transmission map for the input image. The transmission map is then used to recover a haze free image by atmospheric scattering model. Yang et al. in \cite{Yang} proposed the proximal dehaze-net, which combined the haze model, the dark channel, and the transmission prior as the energy terms used in an optimization problem. \\
In \cite{Zhang} a densely connected pyramid dehazing network (DCPDN) is presented. DCPDN jointly learns the transmission map, atmospheric light and dehazing. The conventional patch-based haze removal algorithms (e.g. the dark channel prior) use a fixed patch size, which causes problems such as over-saturation and color distortion. To address those issues, a CNN model called the Patch Map Selection network (PMS-Net) is proposed for an adaptive and automatic patch size selection to each pixel \cite{Chen}.\\
Blur is another kind of image degradation which like glare is modeled as the convolution of a latent blur-free image with the blur kernel a.k.a point spread function (PSF). Image deblurring has been successfully addressed by several different methods in the \cite{10047966, 10124841, ALAM2019, J.sun, S.Nah, ALAM2022116845}. Most of the blind and non-blind deconvolution methods designed for image deblurring do not account for the saturation in the input image which makes their application to glare reduction problem less effective.\\
 The proposed method adopts a straightforward approach of deconvolution, we adopted the Fourier domain methods. We use Wiener deconvolution with the known GSF for glare removal because it is robust for frequencies that has a poor signal-to-noise ratio and apply it as a post-processing step. It tackles a significant limitation observed in previous deconvolution-based glare reduction methods, which is the presence of saturated regions in the image. To overcome this challenge, the method estimates the true radiance of the saturated regions by leveraging the dark channel prior, enabling more accurate glare reduction. 
\section{ Testing framework} 
\label{sec:testing frame work} 
\begin{figure}[h]
\centering
\includegraphics[trim={0 230 55 220},clip,width=3.5in,height=3.0in,keepaspectratio]{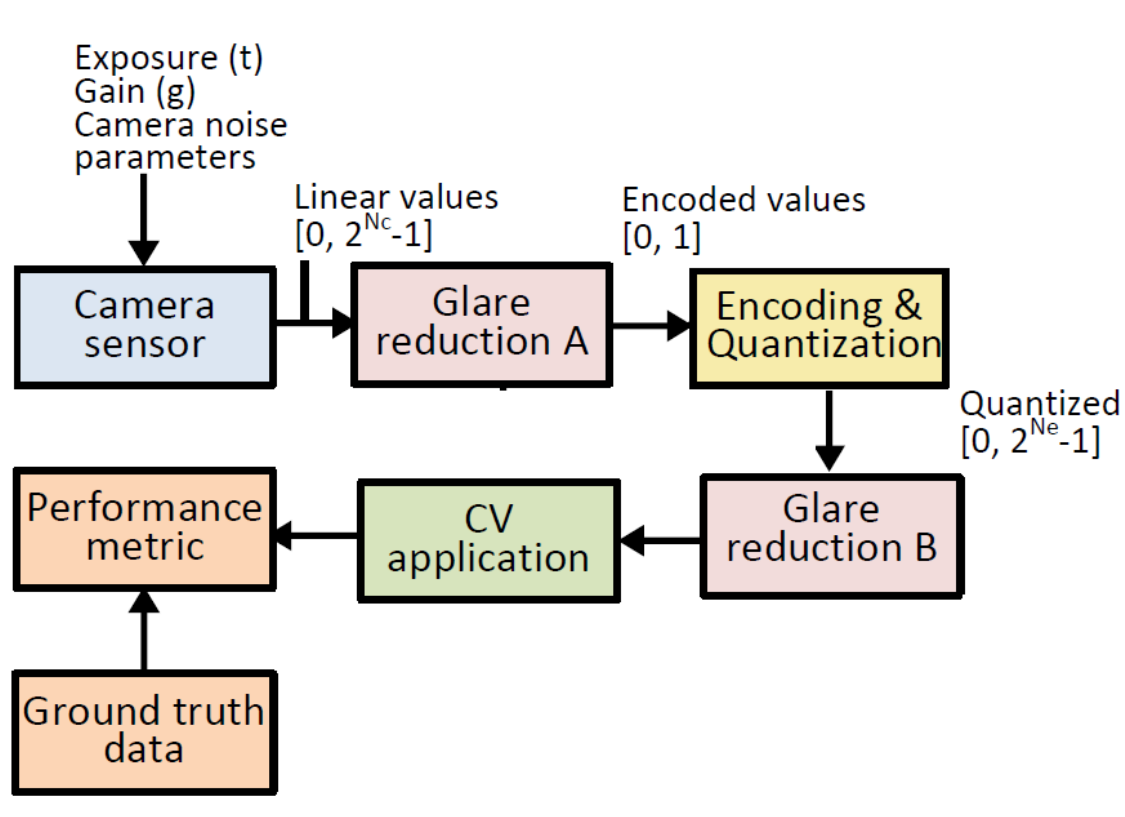}
\caption{Simulated camera pipeline and testing framework.}
\label{fig:testing-framework}
\end{figure}

The overview of our testing framework is shown in Figure \ref{fig:testing-framework}. The camera used for data acquisition is an IDS UI-3860CP-C-HQ computer vision camera, which has a Sony IMX290 1/2.8’’ CMOS sensor with a resolution of 1936× 1096 pixels and pixel size of 2.9 µm. A wide-angle lens, similar to those used in dash cameras, specifically  "Wide-angle Navitar MVL4WA lens with focal length 3.5 mm and the effective field of view of 85\degree × 62.9" is utilized. The testing dataset in this paper consists of the Tunnel scene described in \cite{hanji2021hdr4cv} and available at \footnote{\url{https://doi.org/10.17863/CAM.71285}}. Additionally, the evaluation is extended to a dataset captured by a real autonomous vehicle as is described in the Experimental results section.
The input to the testing pipeline is linear color camera frames. Prior to further processing, the RAW frames were demosaiced using DDFAPD \cite{menon2006demosaicing} and then subjected to white balancing. For white balancing, the RGB values were adjusted to ensure equal intensities across all color channels for the white patch present in the color checker chart. The final 16-bit images were generated by selecting an appropriate exposure from the captured exposure stacks.\\
The glare reduction is performed at two stages of the pipeline. The methods that need to operate on physical light quantities (linear values) reduce glare before it is encoded using a \ac{TF}. This is shown as \emph{Glare reduction A} block in the diagram. However, some glare reduction methods need to operate on (gamma) encoded values. This is shown as \emph{Glare reduction B} block in the diagram. We run the tests with two \acp{TF} (\emph{Encoding \& quantization} block): logarithmic and gamma, along with linear inputs. Logarithmic \ac{TF} is tested because it was one of the best performing \acp{TF} in our previous tests. Gamma \ac{TF} is tested because it is the dominant method of encoding images for CV. \\
Finally, the \ac{TF}-encoded images with reduced effect of glare are passed to the CV algorithms and the results are compared with the ground truth solution. The ground truth is obtained for each CV algorithm by simply testing all the images of the same scene under uniform lighting conditions. All the ground truth images were gamma-encoded to match the encoding of the training data typically used by state-of-the-art CV methods. Finally, the performance metrics used to evaluate each glare reduction method are described in detail in section \ref{sec:metrics}. 
\section{Joint gsf estimation for diverse camera systems}
\label{Joint GSF Estimation}
\begin{figure}[h]
\centering
\includegraphics[width=1.5in,height=1.8in,keepaspectratio]{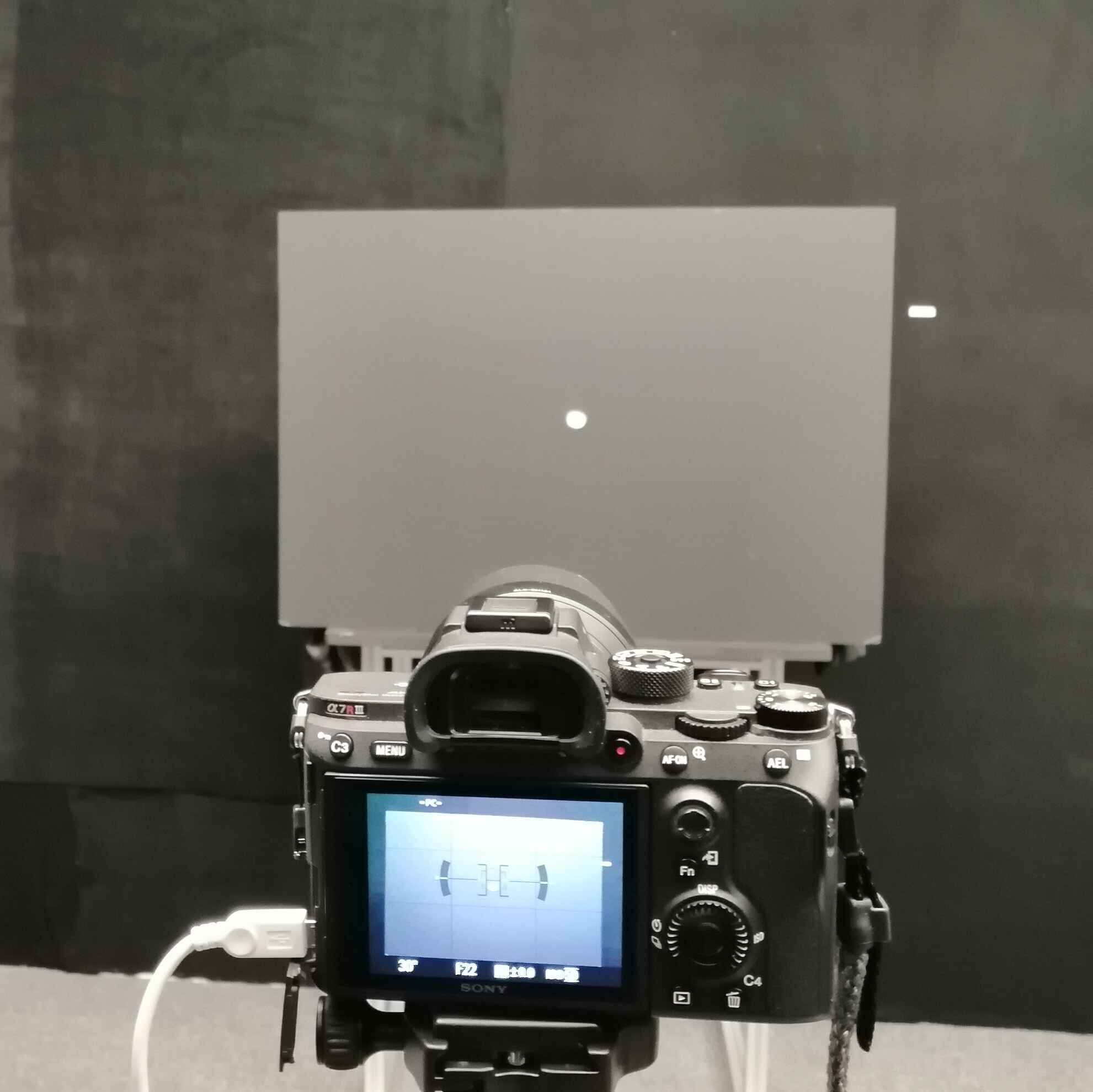}
\caption{Experimental setup to capture high dynamic range image for glare estimation.}
\label{fig:glare-measurement}
\end{figure}

The offline calibration step involves forming a combined dataset from multiple cameras, encompassing a diverse range of glare scenarios. we consider the following camera-lens combinations for the joint GSF estimation: GoPro HERO5, Canon EOS 2000D / Rebel T7 DSLR camera, full-size sensor mirror-less camera (Sony~$\alpha$7R1), and IDS UI-3140CP-M-GL with the following lens combinations: 55\,mm prime lens (Sony SEL55F18Z), Canon EF 50mm f/1.8 STM Lens, and Wide-angle Navitar MVL4WA lens with focal length 3.5 mm, with 25\,mm C-mount lens (Fujifilm HF25HA-1B). \\
 We capture \ac{HDR} images using a small light source that is partially covered by cardboard. The cardboard covered all parts of the light source except for a circular aperture in the center, with a diameter of 2 mm. The setup is illustrated in Figure~\ref{fig:glare-measurement}. Let's denote the intensity of the light source as \( \phi \), the diameter of the aperture as \( d \), and the exposure time for each image as \( t_i \) where \( i \) varies depending on the number of images captured. The total light captured in each image can be represented as a function of these variables.\\
The intensity of light passing through the aperture is expressed as follows: 
\begin{equation}
\phi_a = \phi \cdot \left( \frac{\pi d^2}{4} \right),
\end{equation}
where \( \frac{\pi d^2}{4} \) represents the area of the circular aperture. For each image captured at a different exposure time, the total captured light \( l_i \) can be represented as 
\begin{equation}
l_i = \phi_a \cdot t_i.
\end{equation}
So, the overall capture process where each image captures different levels of brightness, e.g. shorter exposures will capture information in the brightest areas, preventing them from being washed out, while longer exposures will limit noise in darker areas constituting a diverse range of glare scenarios and can be expressed as a series of individual captured light intensities for each exposure time:
\begin{equation}
\begin{split}
l = \{l_1, l_2, \ldots, l_n\} = \{ \phi \cdot ( \frac{\pi d^2}{4} ) \cdot t_1, \phi \cdot ( \frac{\pi d^2}{4} ) \cdot t_2, \\
\ldots, \phi \cdot \left( \frac{\pi d^2}{4} \right) \cdot t_n \}
\end{split}
\end{equation}
where \( n \) is the number of images taken with different exposure times.\\
These images are then merged using \emph{pfstools} \footnote{\url{http://pfstools.sourceforge.net/}} software. To ensure optimal image quality, the exposure times are adjusted accordingly to prevent any pixel saturation in the image captured with the shortest exposure time.
\subsection{Unified Parametric GSF Optimization}
Glare is a physical phenomenon, it is necessary to represent it in terms of linear radiance values. In many cameras, glare can be approximated through a spatially invariant convolution using a GSF. The radiance projected onto the sensor, denoted as $l_s$, can be approximated as follows:
\begin{equation}
    l_s(x,y) = l_{in}(x,y) \circledast g(x,y),
\end{equation}
where, $g(x,y)$ represents the GSF, and $l_{in}$ denotes the incoming radiance projected onto the pixel at coordinates $(x,y)$. Modeling glare in cameras presents a challenge due to the requirement of a kernel, $g(x,y)$, to be twice the size of the image to accurately simulate glare. This is because a bright pixel in one corner of the image can influence the pixel in the opposite corner. However, performing convolution with such large kernels directly is computationally expensive. Therefore, in practice, the convolution is typically performed in the Fourier domain:
\begin{equation}
    L_s(\omega,\phi) = L_{in}(\omega,\phi)\, G(\omega,\phi)\,,
\end{equation}
where $L_s$, $L_{in}$ and $G$ are the Fourier transforms of $l_s$, $l_{in}$ and $g$. To ensure the accurate simulation of glare, we assume that the original radiance map, $l_{in}$, is first expanded to twice its original resolution and then padded with zeros. This serves two purposes: (a) it ensures that the image size is consistent with the size of the convolution kernel, which is twice the size of the image, and (b) it helps to avoid incorrect simulation due to the circular symmetry of the Fourier transform. Without zero padding, a bright light source near one edge of the image may cause a strong glare on the opposing edge. By padding $L_{in}$ with zeros, we can simulate the effect of a well-designed lens hood on the camera, which blocks out all the light that cannot be focused on the sensor.\\ 
Glare in an image arises from the scattering and reflections of light within the camera lens and its body. Therefore, the imaging system's (lens and camera body combined) scattering and reflection patterns needed to be modeled accurately in the glare spread function estimation. As these patterns vary across different lens-camera combinations, we employ a joint GSF estimation strategy as follows: \\
As the ground truth glare-free image $l_{in}$ is available (which is all black except for the circular light source), it is possible to obtain the GSF through deconvolution. However, due to its unreliability and sensitivity to noise, an alternate approach is employed to estimate the GSF. The method involved optimizing the parameters of a radially symmetric function:
\begin{equation}
    g(r) = p_1\,\delta(r) + p_2 \exp\left( -p_3\,r^{p_4} \right),
    \label{eq:param-gsf}
\end{equation}
where $\delta(\cdot)$ is the Dirac delta function, $p_1$, ..., $p_4$ are the parameters of the model and $r$ is distance in the units of sensor pixels ($r=\sqrt{x^2+y^2}$). 
\( p_1 \delta(r) \) involves the Dirac delta function \(\delta(\cdot)\) and represents an instantaneous point of light or glare, modulated by the parameter \( p_1 \). \( p_2 \exp(-p_3 r^{p_4}) \) models the spread of glare away from the central point. It's an exponential decay function where \( p_2 \) scales the intensity, \( p_3 \) influences the rate of decay, and \( p_4 \) adjusts the non-linear response with respect to the distance \( r \). These parameters are estimated by minimizing the following objective function:
\begin{equation}
\begin{aligned}
\min_{p_1, p_2, p_3, p_4} \Bigg( & \sum_{i=1}^{n} \alpha_i \cdot ||\log(l_{si}) - \log(l_{\text{capt}i})||^2 \\ 
& + \lambda \cdot R(p_1, p_2, p_3, p_4) \Bigg)
\end{aligned}
\end{equation}
where, \( l_{\text{capt}i} \) denote the captured HDR image from the \( i \)-th camera. \( l_{si} \) represent the results of convolving the ground truth radiance maps with the parametric GSF. The parameter $\alpha \in [0,1]$ is a weighting factor, and $\lambda$ is the regularization coefficient. $R(p_1, p_2, p_3, p_4)$ is the regularization term, which we chose as the L2 norm of the parameters to prevent over-fitting to one camera type.\\
We employ a non-linear optimization algorithm to minimize the above objective function. The regularization term, controlled by $\lambda$, balances the model's flexibility and generalization capabilities across multiple camera types. A higher value of $\lambda$ ensures that the model does not overly conform to the glare characteristics of one specific camera, thus maintaining its applicability to a wider range of systems.\\
Post-optimization, the GSF model is validated using a separate set of images from multiple cameras. This step is crucial for assessing the model's performance and generalizability. Based on the validation results, we adjust $\alpha$ and $\lambda$ to fine-tune the model, ensuring optimal performance across diverse cameras.
\section{Saturated pixel aware glare reduction}
\label{sec:glare-reduction}
Bright lights in a scene are sometimes difficult to capture even by an HDR camera, for example, the headlight of a vehicle at night or a bright sun shining at the end of a tunnel, which results in some saturated regions $S$ in the image. Such an image $Y$  can be represented as a combination of both saturated $S$ and unsaturated regions $U$.  
\begin{equation}
         Y  = S \cup U \\
         \label{union}
     \end{equation}
The saturated regions are responsible for the maximum glare in the image and deconvolution without true intensities in the saturated regions fails in significant glare reduction. In the proposed  method, outlined in Algorithm \ref{algo1}, the true radiance of the saturated regions is estimated using dark channels prior and the joint GSF, followed by the deconvolution with known GSF.\\
 \begin{algorithm}
\caption{Saturated Pixel-Aware Deconvolution}
\begin{algorithmic}[1]
\State Initialize: 
\State $Y \gets \text{Observed Image}$
\State $G \gets \text{Glare Spread Function}$
\State Decompose $Y$ into $Y_s$ and $Y_u$:
\State $Y_s, Y_u = \text{Decompose}(Y)$
\State Estimate stray light in $Y_u$ using Dark Channel Prior:
\State $Y_uD = \text{EstimateStrayLight}(Y_u)$
\State Minimize objective function for $X_s$ and $G \ast X_u(D)$:
\State $X_s, (G\ast X_u)(D) = \text{MinObjective}(Yu_D, X_s,X_u, G)$
\State Apply constraints for positive pixel values:
\State Ensure $X_u(D) \geq 0$ and $((G \ast X_s) + Y_u \oslash G \geq 0$
\State Substitute saturated areas in $Y$ with estimated $X_s$:
\State $Y_{\text{modified}} = \text{SubstituteSaturatedAreas}(Y, X_s, G)$
\State Perform deconvolution to get glare-reduced image $Y'$:
\State $Y' = \text{Deconvolution}(Y_{\text{modified}}, G)$
\end{algorithmic}
\label{algo1}
\end{algorithm}\\
In the typical\textit{Image formation model:} Glare in an image is represented as the convolution of the GSF $G$ with the latent image $X$.  as shown by the following equation:
\begin{equation}
         Y = G \ast X + n, \\
         \label{model}
     \end{equation}
where $Y$ is the observed image, $G$ is the glare spread function, $X$ is the latent image and $n$ is the additive noise. 
Writing equation \ref{model} to represent separated saturated and unsaturated regions as in equation \ref{union} we get,
     \begin{equation}
     Y_s + Y_u = G \ast X_s  +G \ast X_u + n.
     \label{Combination}
     \end{equation}
To estimate the true radiance of the saturated regions in (\ref{union}), \textit{Dark channel prior} is utilized. According to the dark channel prior’s key observation, there are some pixels $D$, where $D \in U $, in a glare-free image  $X_u(D)$  with very low-intensity values, approaching zero, in at least one color channel and therefore, intensity values at these pixels, in a glared version of the image $Y_U(D)$, can be considered as the approximated amount of stray light at that pixel.  However, the presence of noise in the dark regions limits the performance of this method. To lessen the effect of noise, we apply a Gaussian blur to the image before finding the minimum values. Based on aforementioned assumption we can write (\ref{Combination}) in terms of dark pixels $D$ as:
  \begin{equation}
       Y_u(D) = (G \ast X_s)(D) +( G \ast X_u)(D)  + n.
   \end{equation} 
We minimize the following objective function subject to the constraints given in (\ref{const1}), (\ref{const2}), and (\ref{const3}), to estimate $X_S$ and $(G \ast X_U)(D)$.
\begin{equation}
\begin{split}
\textrm{argmin}_{X_s,({G\ast X_u})(D)} \left \|Y_u(D) - (G \ast X_s)(D)-   {(G \ast X_u)(D)} \right \|_2 \\
+  \lambda \left \| {G \ast X_u}(D) \right \|_1.     
\end{split}
\end{equation}
\begin{equation}
    Y_u - \left( G * (X_s+\tilde{X}_u))(u)  \right) \geq 0,
    \label{const1}
\end{equation}
the constraint in (\ref{const1}) is akin to the primary data term but is contingent on all pixels without
saturation, serving the purpose of preventing negative pixel values. It aims to prevent negative pixel values. Here, $Y_u$ denotes the unsaturated pixels, and $\tilde{X}_u$ is obtained by deconvolving $Y_u$ with the glare spread function $G$.
\begin{equation}
    {X}_u(D) \geq 0,
    \label{const2}
\end{equation}
where (\ref{const2}) ensures that all estimates for dark pixels in the image remain positive, thus preventing non-physical values in darker regions.
\begin{equation}
    {{((G*X_s)+Y_u) \oslash G}},  \geq 0
    \label{const3}
\end{equation}
where (\ref{const3}) applies to the combined image of estimated saturated pixels ($X_s$) and captured unsaturated ones ($Y_u$). The operation $\oslash$ denotes deconvolution, and this ensures that the deconvolution outcome remains strictly positive in each iteration.\\
Given that the GSF extends to double the image's size, performing deconvolution in the spatial domain becomes impractical. Consequently, we have opted for the Fourier domain method. For glare removal, \textit{Wiener deconvolution} is utilized in conjunction with the known GSF. This approach is particularly effective due to its robustness in handling frequencies with a low signal-to-noise ratio. The glare-reduced image $Y'$ is ultimately derived by substituting the saturated areas in the observed image $Y$ with the estimated radiance values $X_s$, followed by Wiener filtering with the known GSF, and is represented as:.
\begin{equation}
    Y'={{((G*X_s)+Y_u) \oslash G}}\,.
\end{equation}
It is important to emphasize that the successful recovery of the final image $Y'$ relies on two primary factors: the availability of ample dark regions in the input image degraded by glare, which enables accurate estimation of the stray light despite the presence of noise, and the accuracy of the adopted glare model in measuring the GSF $G$.\\
\vspace{-0.10in}
\section{Transfer functions for glare-induced luminance variations}
\label{sec:transfer-functions}
Glare significantly impacts the dynamic range of a scene, often leading to overexposed or underexposed areas in an image. Therefore, the choice of encoding becomes critical as different transfer functions offer unique approaches to mitigate these challenges. Encoding methods tailor their response to the luminance affected by glare, ensuring optimal detail preservation and dynamic range handling for improved computer vision performance.\\
Most camera sensors, including CCD and CMOS, exhibit a linear response to light, counting incoming photons and registering values linearly related to radiometric quantities (irradiance). While linear color values are perceptually non-uniform and require high bit-depths for storage, they are often tone-mapped into gamma-corrected representations for visually appealing content. However, for computer vision (CV) tasks, where perceptual uniformity is not as critical as detailed information capture, especially under glare, transfer functions that: a) can encode a large dynamic range in the presence of extreme glare; b) are efficient in terms of required bit-depth becomes imperative. 
 \subsection{Gamma encoding}
Gamma encoding is a non-linear operation that compresses linear light intensities from the sensor by raising the values to an exponent less than 1, typically in the range from $1/2.4$ to $1/1.7$. It makes the encoded values more perceptually uniform and it partially accounts for the non-uniform variance of the camera noise. Gamma encoding is expressed by the following equation:
\begin{equation}
V_{\text{out}} = V_{\text{in}} ^ {(1 / \gamma)},
\end{equation}
where $V_{\text{in}}$ is the input pixel value, $V_{\text{out}}$ is the output pixel value after gamma encoding, and gamma is the gamma value. The power-law relationship in gamma encoding intensifies the lower luminance range, crucial in counteracting glare's tendency to overexpose and wash out details.
\subsection{Lograthmic encoding}
Logarithmic encoding is used to transform pixel values in an image to better represent a wide range of intensity levels. It is particularly useful in applications where a large dynamic range needs to be captured such as the presence of glare source in the scene.  We define our logarithmic TF as:
\begin{equation}\text{if}
    E(Y)=\begin{cases}
    \frac{\log_2{Y}}{\log_2(2^N-1)} & \textrm{if}\quad Y>=1 \\
    0 & \textrm{otherwise} \\
    \end{cases}
    \label{log}
\end{equation}
where $N$ is the number of bits used to represent the linear digital signal $Y$. The encoded values $E$ range between 0 and 1 and can be represented using the desired number of bits. In (\ref{log}) the upper end of the luminance spectrum is adeptly compressed. This is vital as glare often results in high luminance values that can overpower subtle image details. 
\subsection{Linear encoding} The straightforward linearity expressed by \ref{linear} ensures minimal alteration of the original luminance values. In scenarios where glare's effect is less pronounced, maintaining the integrity of the original data is beneficial for accurate image analysis.
\begin{equation}
V_{\text{out}} = m * V_{\text{in}} + c,
\label{linear}
\end{equation}
where \( V_{\text{in}} \)  input pixel value, \( V_{\text{out}} \)  is transformed to the output, \( m \) is the gain factor and \( c \) is the offset.\\
These transfer functions, grounded in their distinct mathematical properties, provide tailored solutions to the various challenges posed by glare, and their effective application in autonomous vehicle perception for handling high dynamic range scenes encountered in the presence of glare is inevitable. 
\section{ av perception algorithms} 
\label{sec:metrics}
Machine vision plays a crucial role in autonomous driving perception by enabling the vehicle to understand and interpret its environment. Some of the key machine vision applications involved in autonomous driving and included in this paper are object detection (OD), object recognition (OR), lane detection (LD), object tracking (OT), and depth estimation (DE). Due to the inherent vulnerabilities of these perception algorithms, as they are predominantly designed for the visible light spectrum, when faced with glare encounter significant performance degradation. \vspace{-0.10in}
\subsection{Object detection:} We have used state-of-the-art deep learning algorithms, Yolov5 \cite{yolov5} to detect various objects in the scene, such as vehicles, pedestrians, traffic signs, and traffic lights. Our test scene "Tunnel" contains a large variety of objects encountered on and around the road, and some of the images of the test scene are presented in Fig. \ref{fig:TFVIZ} and \ref{fig:Cv_compviz}. Timely and precise detection of the aforementioned and all other possible road objects under all possible conditions, including the presence of strong glare, is imperative for the autonomous vehicle to make informed decisions and navigate safely. However, glare reduces the contrast between objects and their backgrounds which is crucial for object detection causing extreme vulnerability to glare.
Therefore, to evaluate the performance of glare reduction methods for object detection, ground truth is generated by applying the same detection algorithm on the dataset captured under uniform lighting without the presence of a major source of glare, as shown in Fig. \ref{fig:Cv_compviz}; and all the detected object bounding box (BB) coordinates are stored as ground truth.\\
The Mean Intersection over Union (MIoU) metric, given below was selected for evaluating glare reduction in object detection due to its effectiveness in assessing precision and recall, measuring the overlap accuracy between predicted and actual bounding boxes.
\begin{equation}
{\text{MIoU}}=\frac{area(R_{pred} \bigcap R_{ref})}{area(R_{pred}\bigcup R_{ref})},
\end{equation}
where $R_{pred} = \bigcup_{i=1}^{N} B_{pred}(i)$ is the union of all $N$ bounding boxes predicted and $R_{ref} = \bigcup_{j=1}^{M} B_{ref}(j)$ is the union of all $M$ reference bounding boxes.
\subsection{Object Recognition:}
Object recognition plays a crucial role in autonomous vehicles for perception and understanding of the surrounding environment.  Once objects are detected, they need to be classified into specific categories or classes for example, pedestrians, cyclists, cars, or traffic signs. We have used state-of-the-art Detic \cite{detic}  object recognition method to quantitatively evaluate the performance of the selected glare reduction methods in object recognition. Like OD, the dataset for testing recognition is the tunnel scene which contains a vast range of objects encountered during driving. \\
In addition to classification, Detic \cite{detic} also performs segmentation that classifies every pixel with a corresponding object label. This fine-grained understanding of the scene allows the vehicle to differentiate between regions and objects, enabling more precise decision-making and allowing for safer navigation, and obstacle avoidance in complex traffic scenarios.
Like OD  the ground truth for the evaluation of OR is generated by running the same algorithm on a dataset of the same scene except without the major source of glare and with uniform lighting. The evaluation metric of choice is mean average precision (MAP) which measures the quality of ranked retrieval results or the accuracy of object detection and localization. MAP is an extension of Average Precision (AP) that computes the average precision for each class and then takes the mean over all classes:
\begin{equation}
\text{AP} = \sum_n (R_n - R_{n-1}) P_n,
\end{equation}
where $R_n$ and $P_n$ are the precision and recall at the $n$th threshold.
\begin{equation}
\text{MAP} = (\text{AP_1} + \text{AP_2} + ... + \text{AP_N}) / N,
\end{equation}
$N$ is the total number of classes or categories in the dataset, and AP_1, AP_2, ..., and  AP_N are the Average Precision scores for each class.
\subsection{Object Tracking:}
Object tracking helps the vehicle understand the movement and behavior of objects in its surroundings. OT identifies and monitors objects in the environment, predicting their future positions, and understanding how they interact with the vehicle. Therefore, it contributes to the vehicle's overall situational awareness. By continuously updating its knowledge of the surrounding environment, the vehicle can adapt to changing road conditions, traffic dynamics, and unexpected obstacles. The object tracking algorithm estimates the current position and motion state of the vehicle which involves determining the vehicle's position, velocity, acceleration, and orientation. \\
The tracking algorithm used in this work \cite {wojke2017simple} is a deep learning-based method called DeepSORT, that takes into account the object's current state and historical data to forecast how it will move in the coming seconds.
Like other perception tasks, ground truth for OT is generated using the dataset of the scene with uniform lighting and in the absence of a major source of glare and the evaluation metrics used or (Multiple Object Tracking Accuracy) MOTA and (Multiple Object Tracking Precision) MOTP. MOTA evaluates the performance of detection, misses, and ID switches. The accuracy of the tracker, MOTA  is calculated by:
\begin{equation}
{\text{MOTA}}= 1- \frac{\sum_{t}  \text{FN}_{t} + \text{FP}_{t} + \text{IDS}_{t})}{\sum_{t}\text{GT}_{t}},
\end{equation}
where FN is the number of false negatives, FP is the number of false positives, IDS is the number of identity switches at time $t$ and GT is the ground truth. MOTA can also be negative.
MOTP is be calculated with the following formula,
\begin{equation}
{\text{MOTP}}= 1- \frac{\sum_{t,i}  d_{t,i}}{\sum_{t}c_{t}},
\end{equation}
where $d_{t,i}$ is the bounding box overlap of target $i$ with its assigned ground truth object and $c_t$ are total matches made between the ground truth and the detection output.
\subsection{Lane detection:}
Lane detection is an integral part of autonomous vehicles' perception systems. when an AV intends to change lanes or merge onto a highway. It helps the vehicle identify gaps in traffic and determine when it's safe to change lanes. Lane change maneuvers are coordinated with object detection systems to avoid collisions with other vehicles. Deep learning-based lane detection algorithms offer advantages in terms of their ability to handle a wide range of road conditions and adapt to different environments. They can also learn complex features that may be challenging to design manually. 
In this work, we adopted a deep learning model \cite{pan2018spatial}  to evaluate the effectiveness of various glare reduction techniques in lane detection. Evaluation of the regression model’s performance can best be described in terms of error values. For this purpose, we calculated the commonly used root mean square error metric (RMSE) [52] and \cite{alam2022learning}. The following equation presents the formula to calculate RMSE.
\begin{equation}
\text{RMSE} =\sqrt {\frac {\sum^N _{i=1} (\text{Actual}_{i} - \text{Predicted}_{i})^2}{N_{{images}}}}
\end{equation}
where Predicted$_{i}$ represents the predicted lane line point coordinates value in pixels units, and Actual$_{i}$ is the ground truth value for the same point coordinates. In practice, Actual and Predicted values are given as two-dimensional vectors with x and y coordinates of the selected points on the lane lines. So the smaller RMSE will indicate that the predicted lane line is closer to the ground truth than a predicted line with a large RMSE.
\subsection{Depth estimation:}
Depth estimation provides essential information about the three-dimensional structure of the environment. This information is critical for safe and effective navigation and interaction with the surroundings. Depth data aids in path planning and control algorithms. It allows the vehicle to determine the road's topography and make decisions on acceleration, deceleration, and steering that are appropriate for the terrain. Understanding the depth of the road and the surrounding environment is crucial for executing safe maneuvers like merging onto highways, changing lanes, or navigating complex intersections.\\
Deep learning-based monocular depth estimation methods aim to predict the depth or distance information of objects in a scene from a single input image, We adopted the state-of-the-art monocular depth estimation method MIDAS \cite{01341} in this work.
Similar to LD we have utilized RMSE to evaluate the performance of various glare reduction methods for depth estimation. The RMSE is estimated directly from the depth map estimated by the ground truth image and the depth map of the resultant image after applying the glare reduction technique on the input glared image.
\section{Experimental results}
\begin{figure}[!h]
\centering
{\includegraphics[trim={7 10 16 20 },clip,width=3.5in,height=3.5in,keepaspectratio]{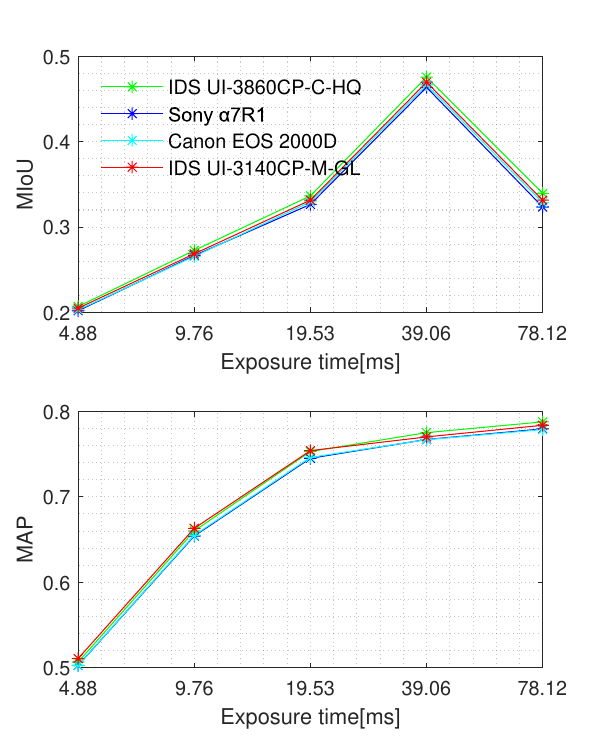}}
\caption{Impact of joint GSF from different camera and lens combinations, described in \ref{Joint GSF Estimation} on object detection and recognition tasks involved in autonomous vehicle perception module. From top to bottom (in legend) the optimized  GSF  represents all the camera types upto the mentioned type e.g red line represents the joint GSF for IDS UI-3860CP-C-HQ, Sony $\alpha$7R1, Canon EOS 2000D, and IDS UI-3140CP-M-GL.}
\label{fig:Rcamercomp}
\end{figure}

\begin{figure}[!h]
\centering
{\includegraphics[trim={48 0 23 1},clip,width=3.8in,height=2.6in, keepaspectratio]{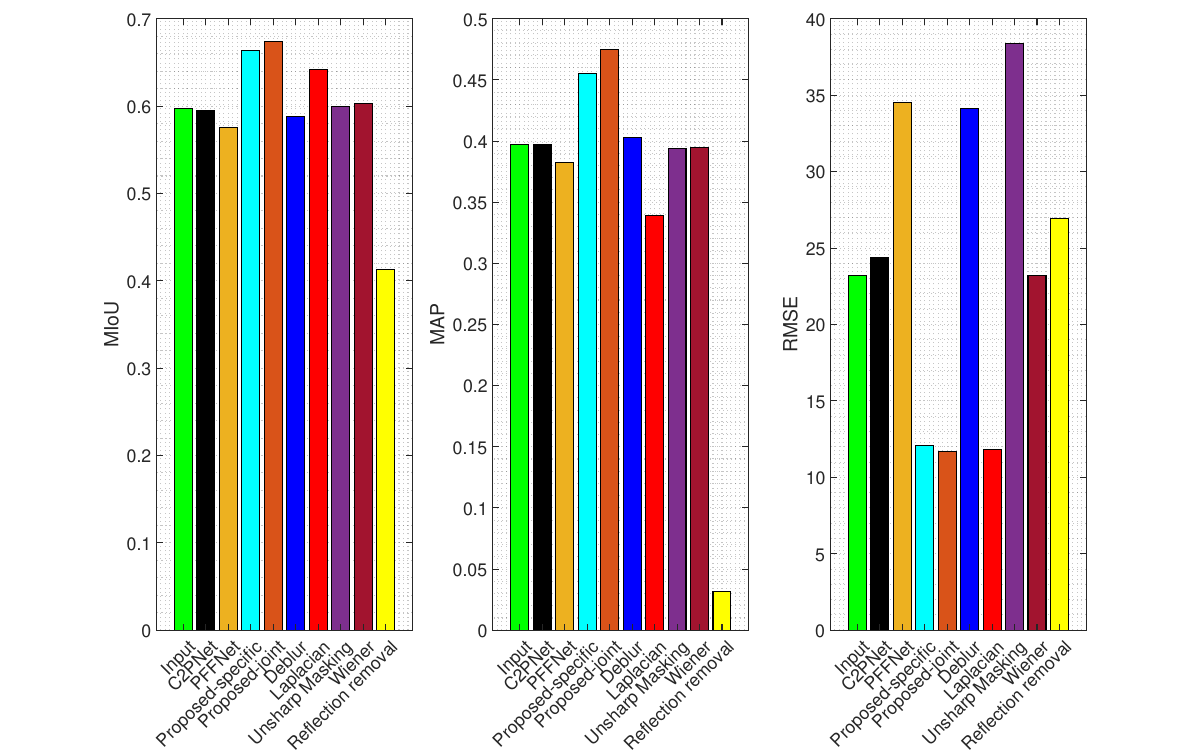}}
\caption{Comparison of glare reduction methods based on the performance of various CV applications on a real dataset \cite{boisclair2022attention}. (Top: Left to Right) Object detection, Object recognition, Object tracking. (Bottom: Left to Right) Object tracking, Lane detection, and Depth estimation.}
\label{fig:Cv_comp_real}
\end{figure}

\begin{figure*}[!h]
\centering
{\includegraphics[trim={7 33 10 13},clip,width=5.8in,height=4.6in, keepaspectratio]{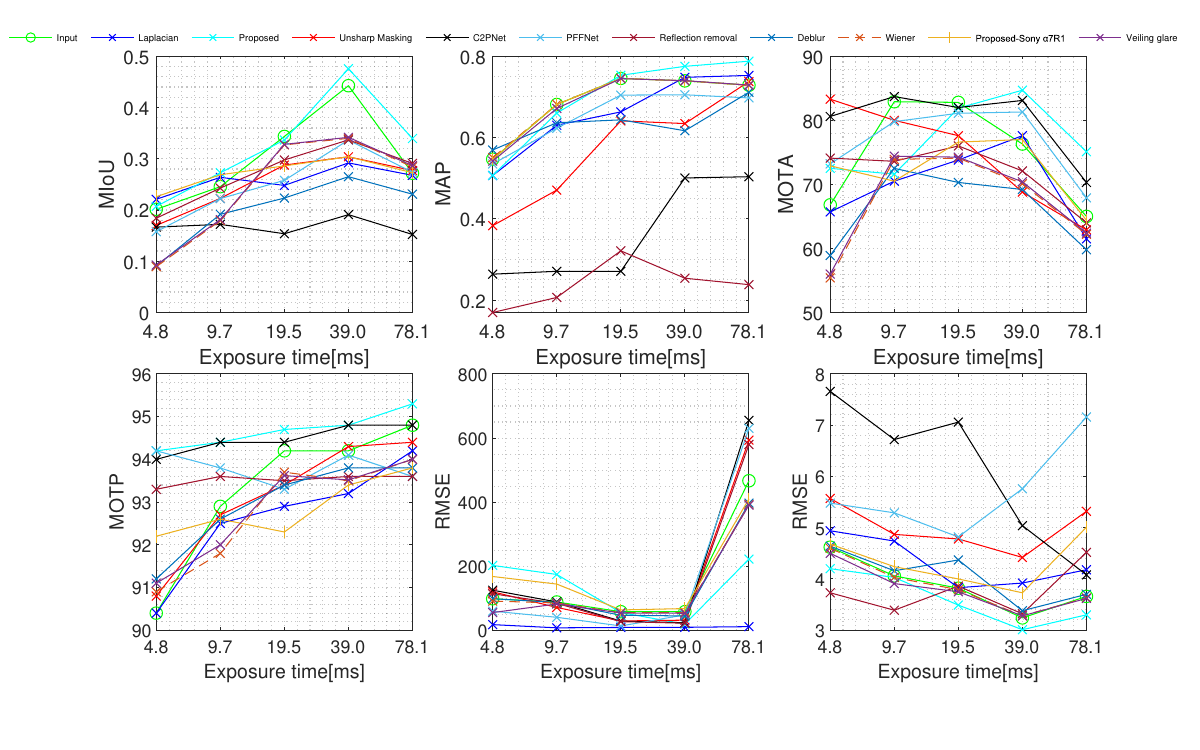}}
\caption{Comparison of glare reduction methods based on the performance of various CV applications. (Top: Left to Right) Object detection, Object recognition, Object tracking. (Bottom: Left to Right) Object tracking, Lane detection, Depth estimation. (Glare reduction methods) Input, Proposed, Laplacian Filter, Reflection removal \cite{Refl}, PFFNet \cite{mei2018pffn}, C2PNet \cite{zheng2023curricular}, Unsharp masking, Deblur \cite{Deblur_chen}, Wiener Filter, Veiling glare removal \cite{Talvala} }
\label{fig:Cv_comp}
\end{figure*}

\begin{figure*}[!h]
\centering
\begin{subfigure}{1.3in}
{\includegraphics[width=1.3in,height=1.3in,keepaspectratio]{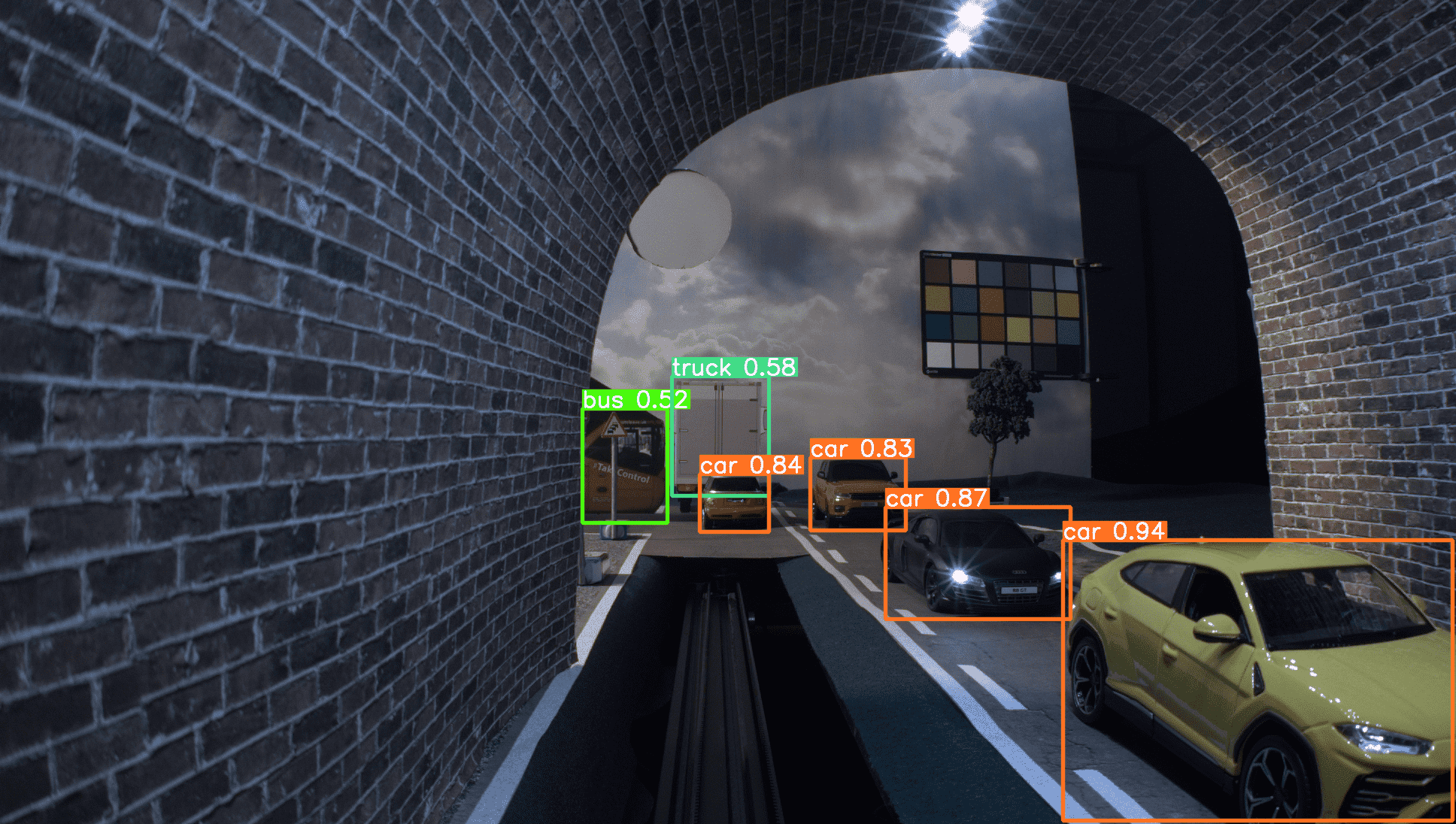}}
\end{subfigure}
\hfill
\begin{subfigure}{1.3in}
\subfloat{\includegraphics[width=1.3in,height=1.3in,keepaspectratio]{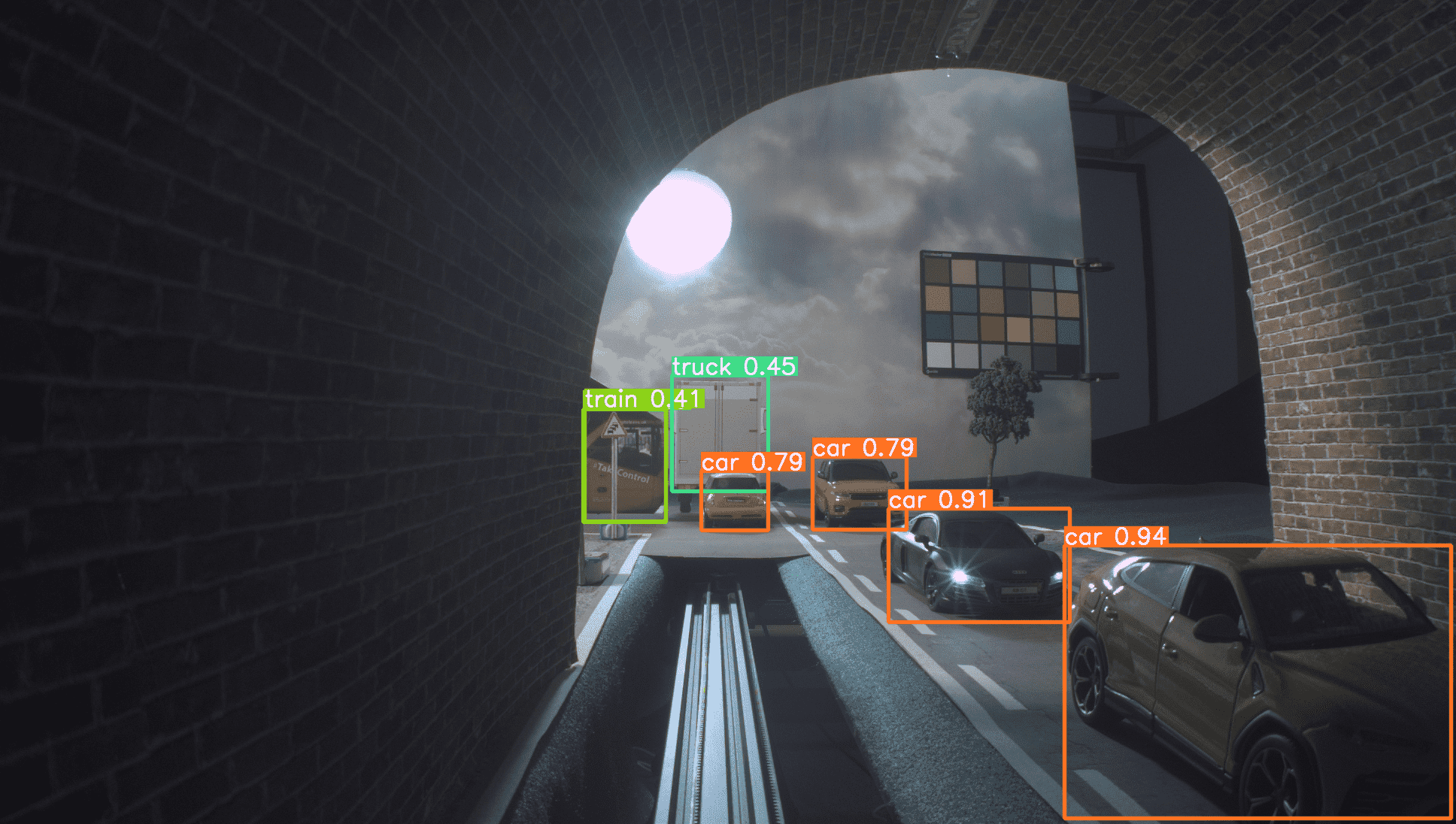}}
\end{subfigure}
\hfill
\begin{subfigure}{1.3in}
\footnotesize
\stackunder[5pt]{\includegraphics[width=1.3in,height=1.3in,keepaspectratio]{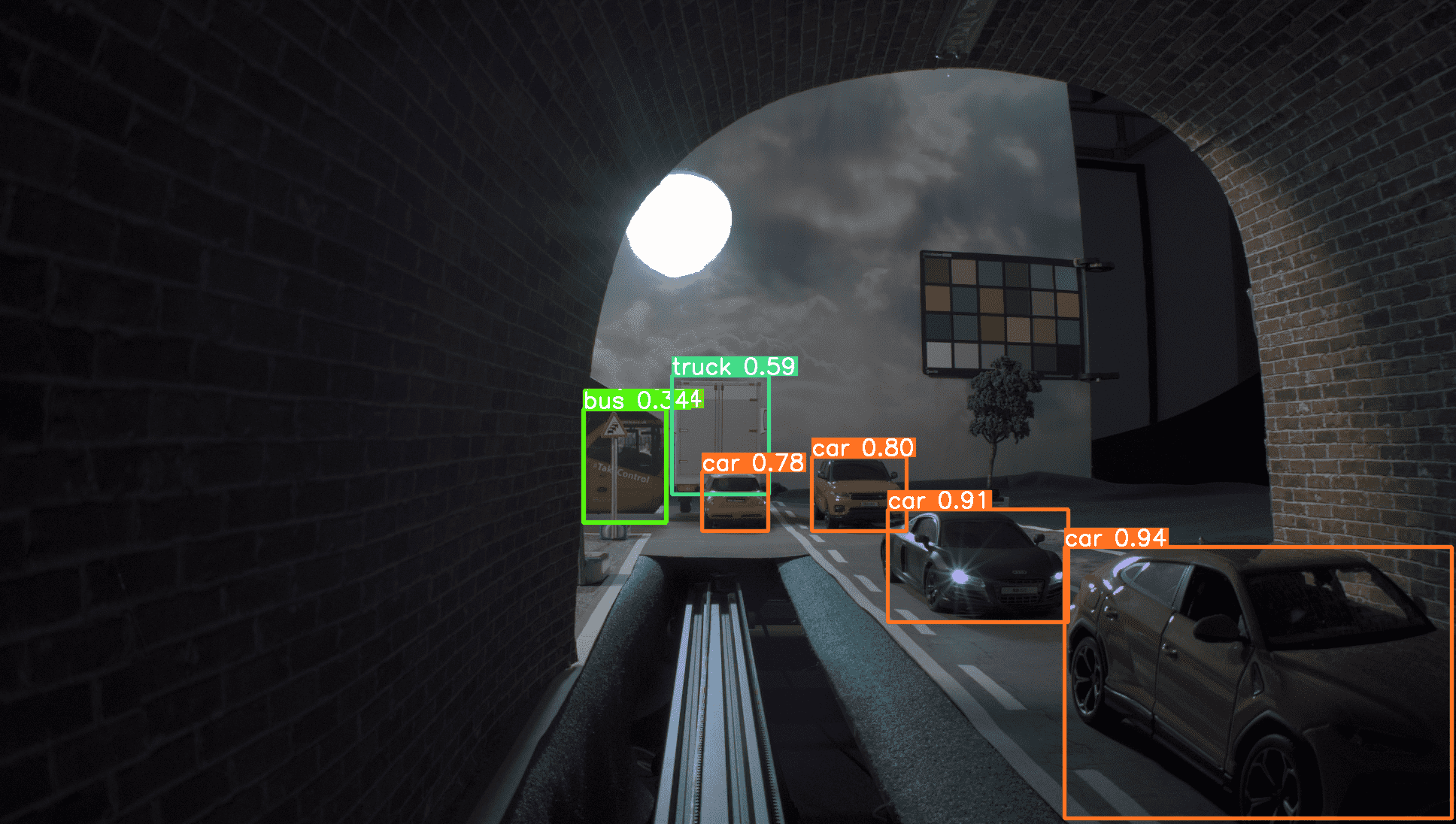}}{Top}
\end{subfigure}
\hfill
\begin{subfigure}{1.3in}
\subfloat{\includegraphics[width=1.3in,height=1.3in,keepaspectratio]{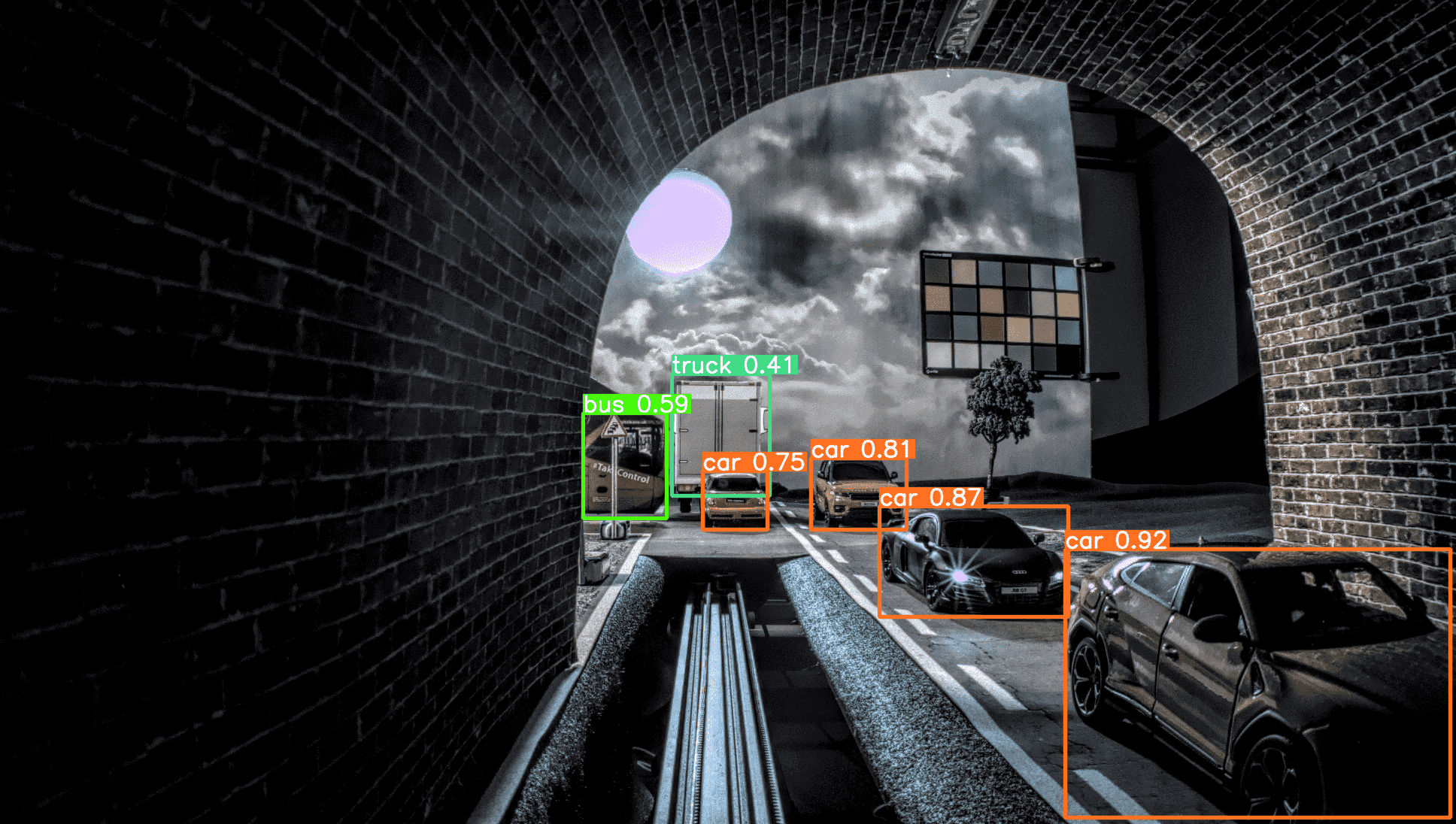}}
\end{subfigure}
\hfill
\begin{subfigure}{1.3in}
\subfloat{\includegraphics[width=1.3in,height=1.3in,keepaspectratio]{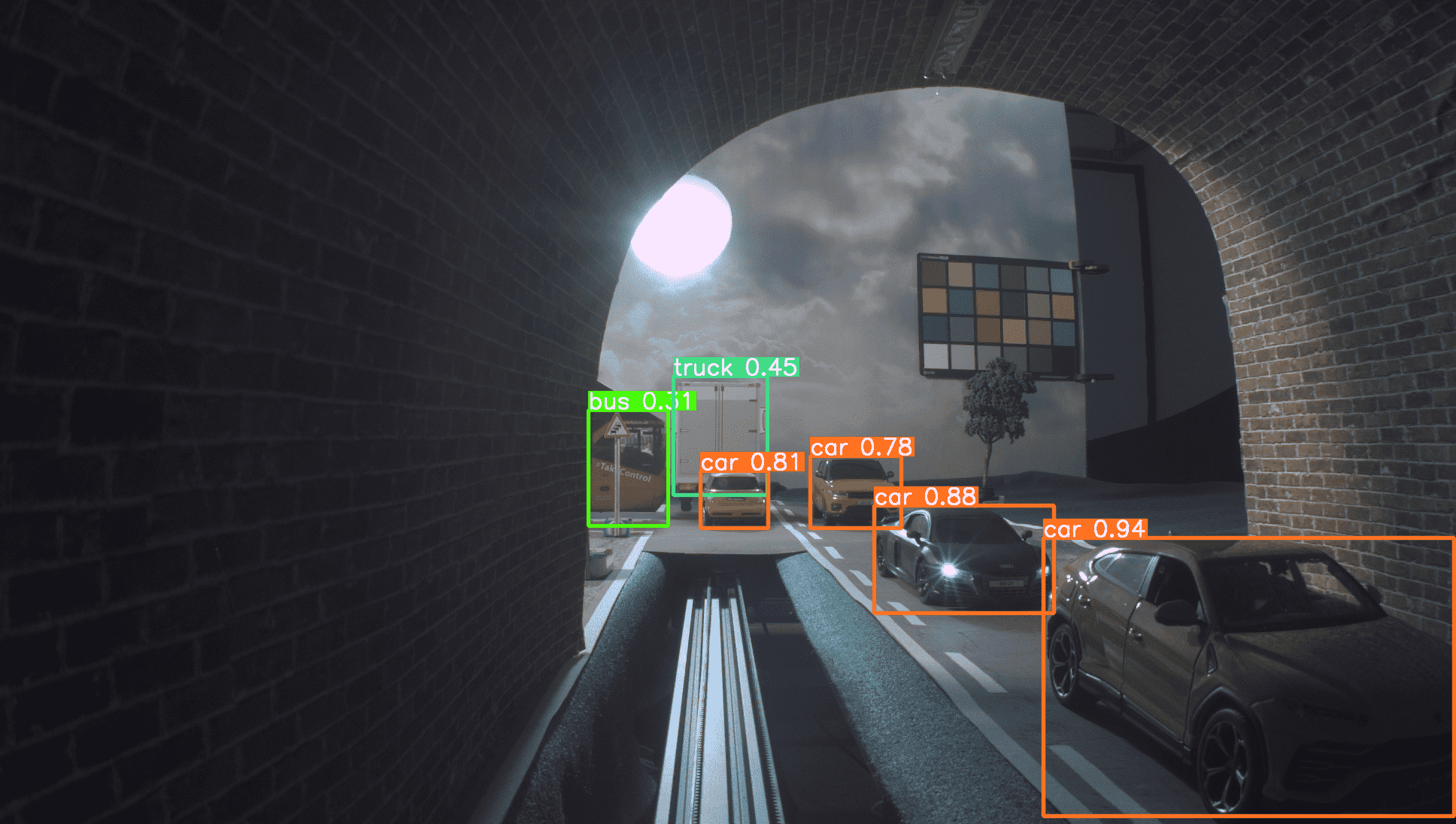}}
\end{subfigure}\hfill
\begin{subfigure}{1.3in}
\subfloat{\includegraphics[width=1.3in,height=1.3in,keepaspectratio]{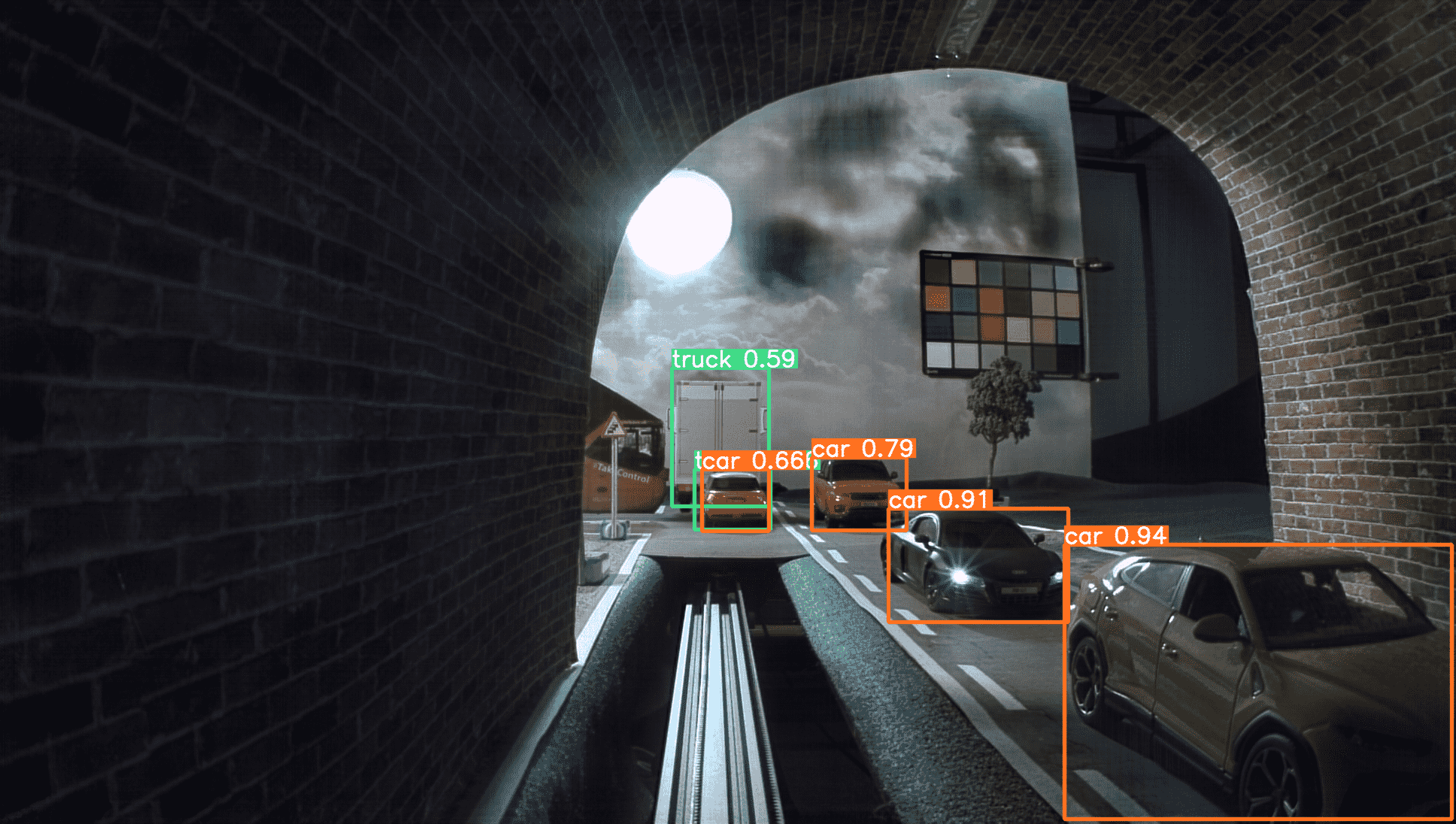}}
\end{subfigure}
\hfill
\begin{subfigure}{1.3in}
\subfloat{\includegraphics[width=1.3in,height=1.3in,keepaspectratio]{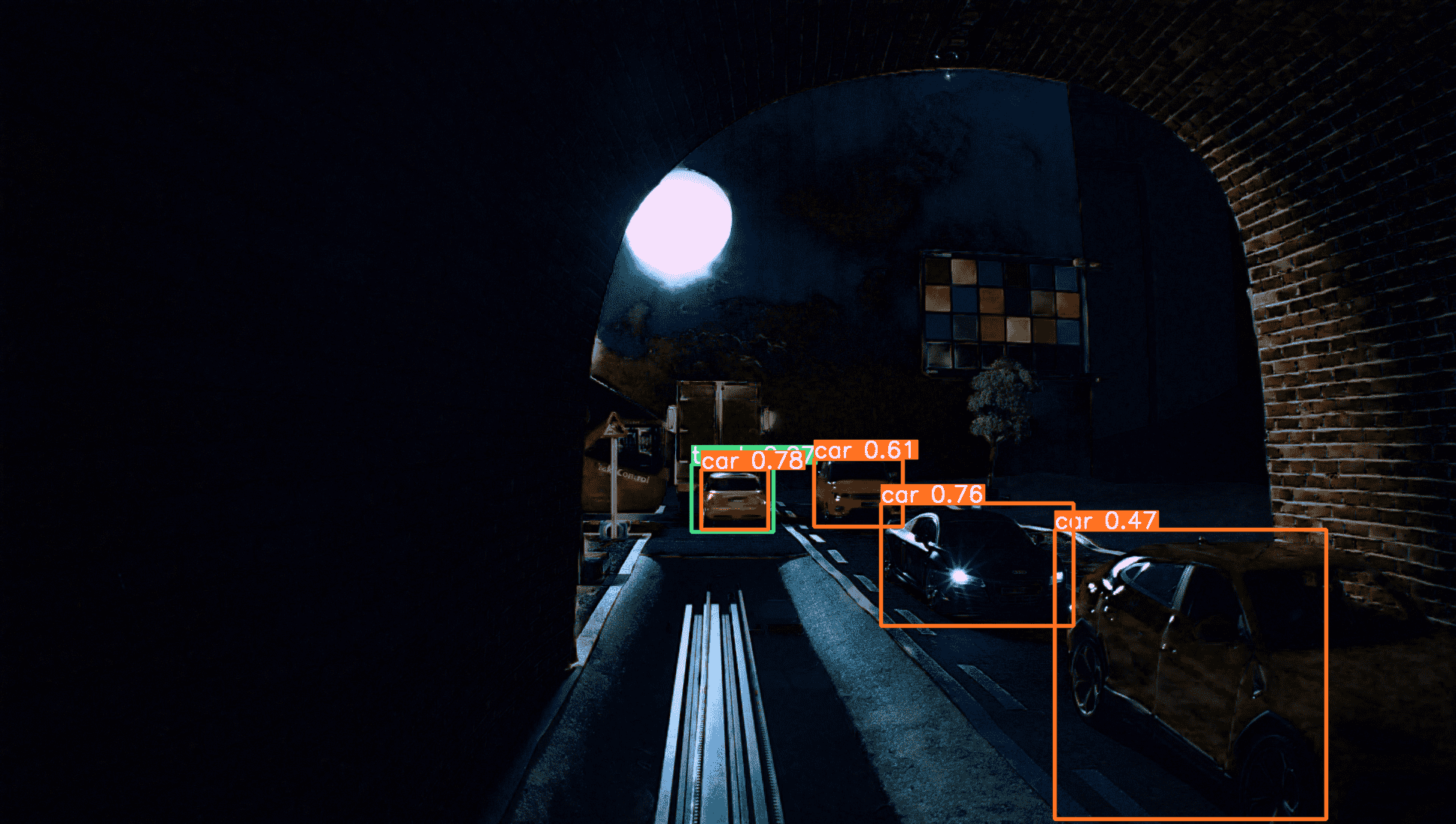}}
\end{subfigure}
\hfill
\begin{subfigure}{1.3in}
\footnotesize
\stackunder[5pt]{\includegraphics[width=1.3in,height=1.3in,keepaspectratio]{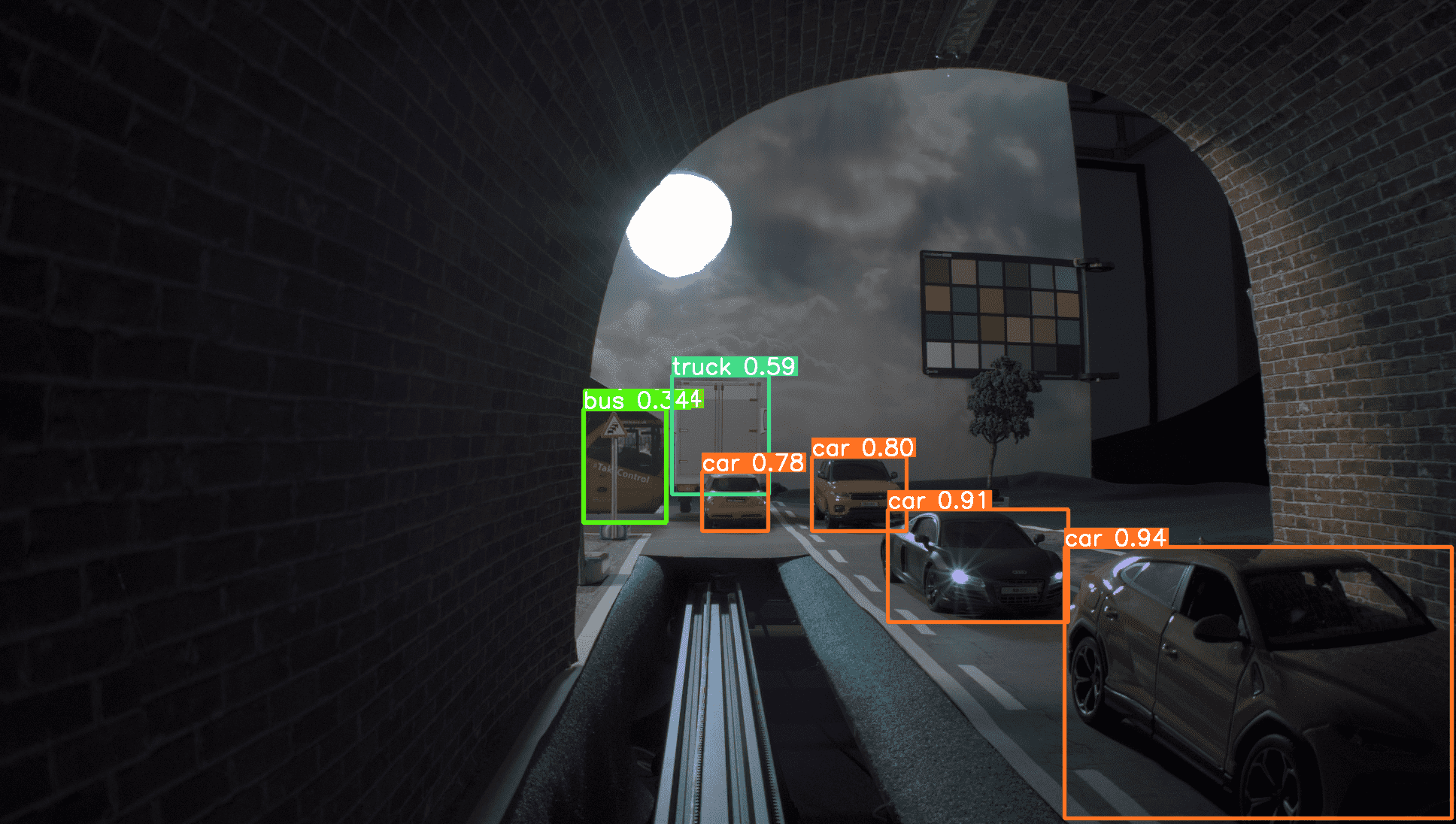}}{Bottom}
\end{subfigure}
\hfill
\begin{subfigure}{1.3in}
\subfloat{\includegraphics[width=1.3in,height=1.3in,keepaspectratio]{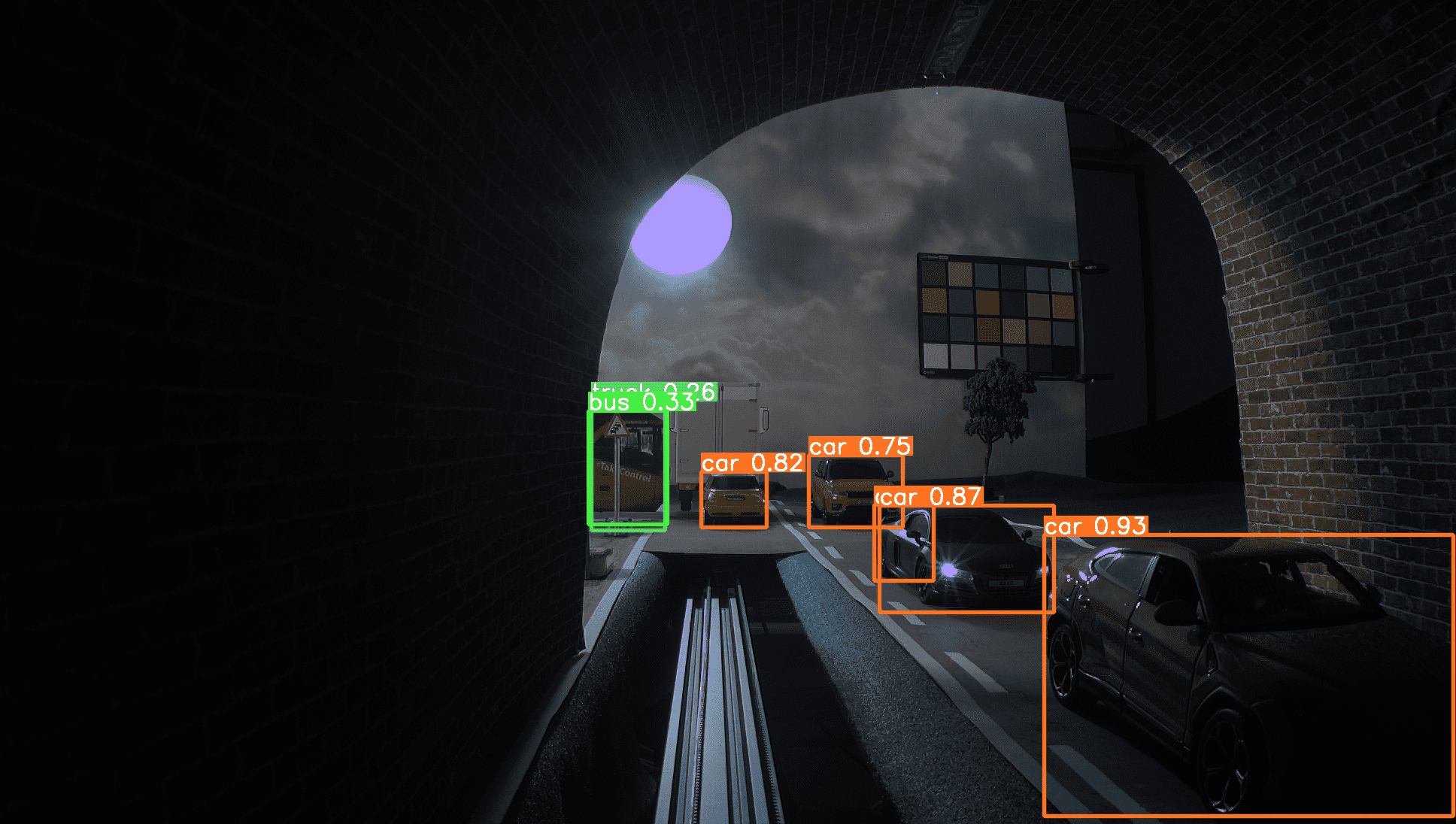}}
\end{subfigure}
\hfill
\begin{subfigure}{1.3in}
\subfloat{\includegraphics[width=1.3in,height=1.3in,keepaspectratio]{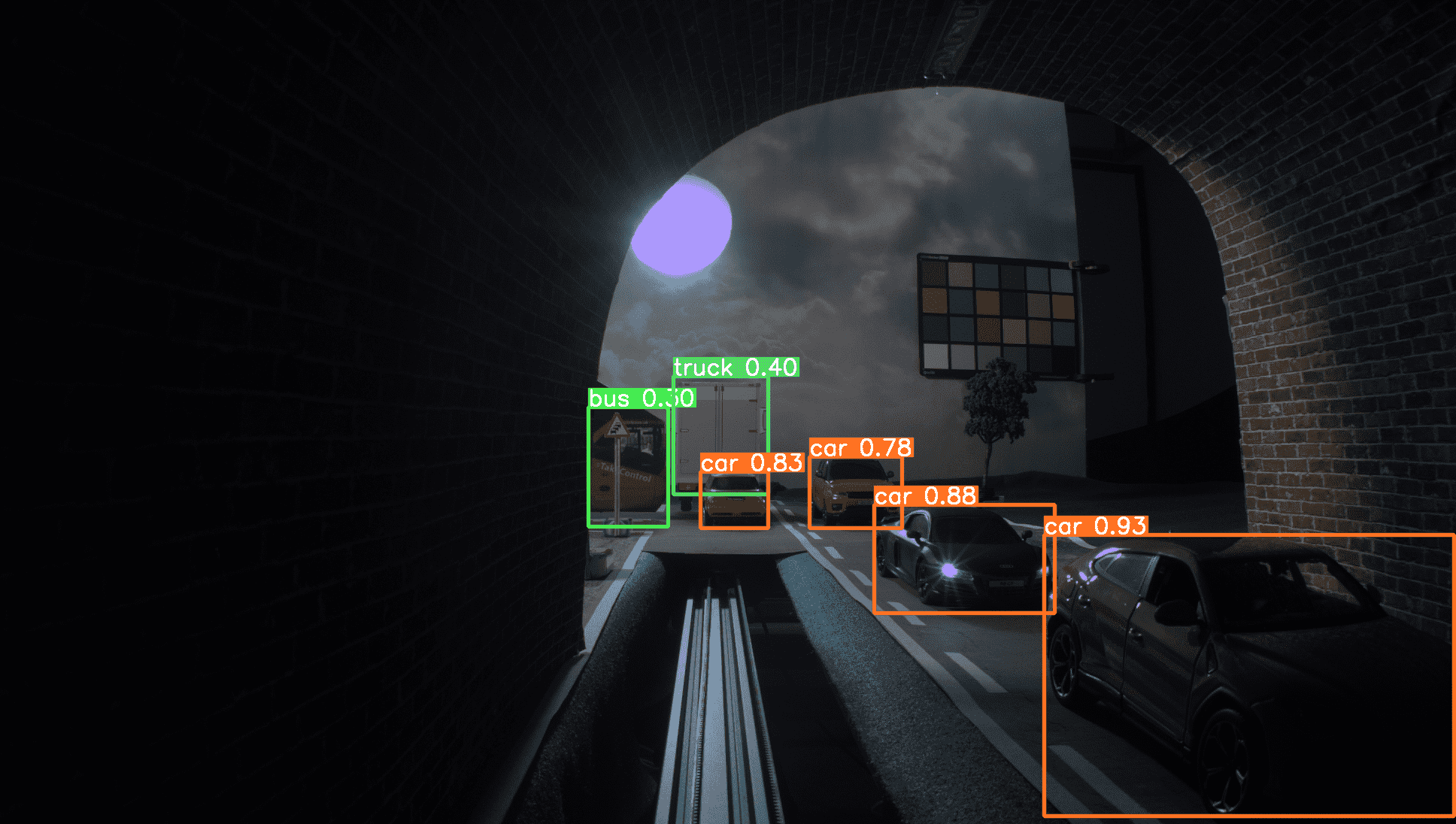}}
\end{subfigure}\hfill
\begin{subfigure}{1.3in}
\subfloat{\includegraphics[trim={75 70 75 75},clip,width=1.3in,height=1.3in,keepaspectratio]{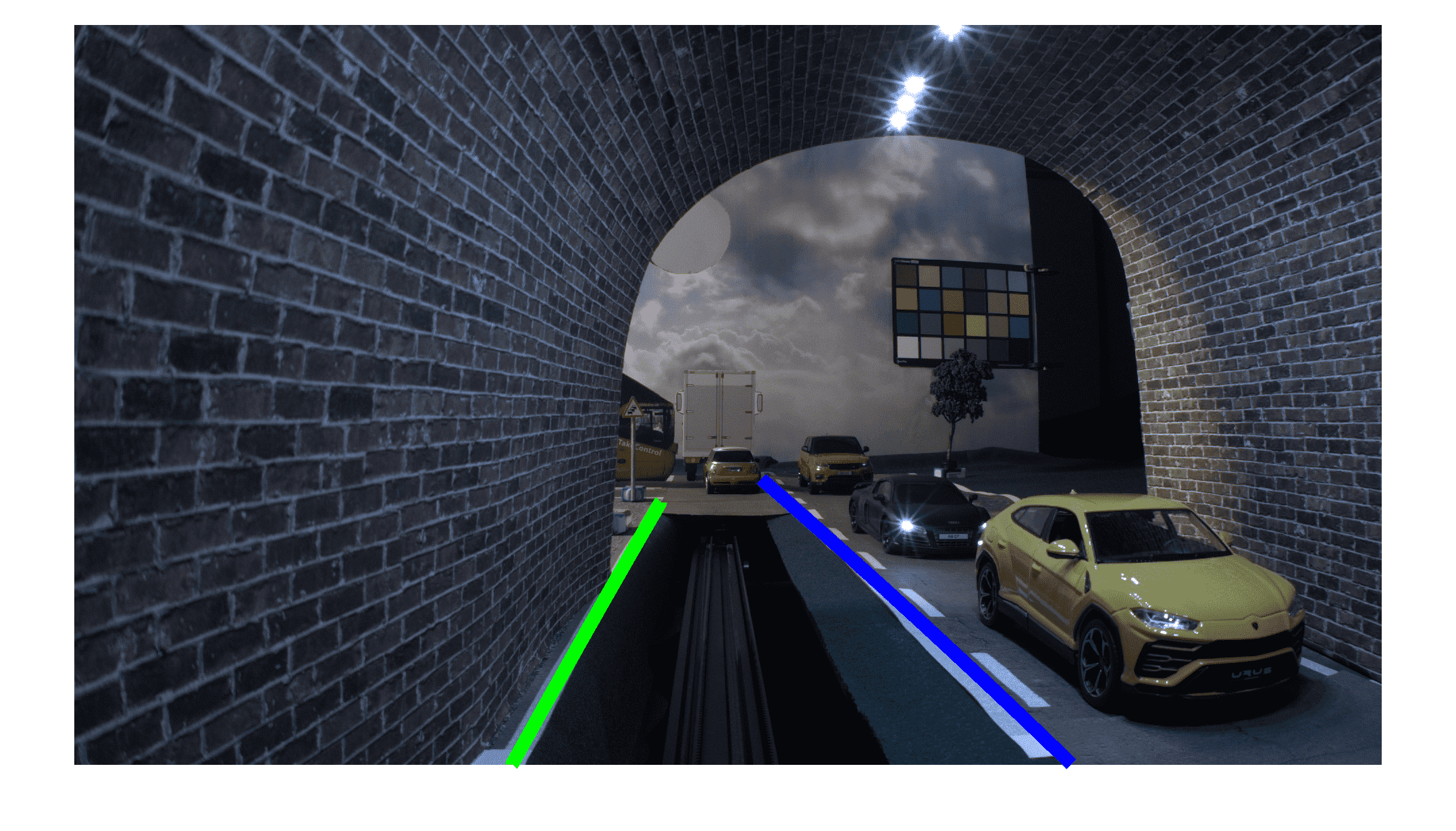}}
\end{subfigure}\hfill
\begin{subfigure}{1.3in}
\subfloat{\includegraphics[trim={75 70 75 75},clip,width=1.3in,height=1.3in,keepaspectratio]{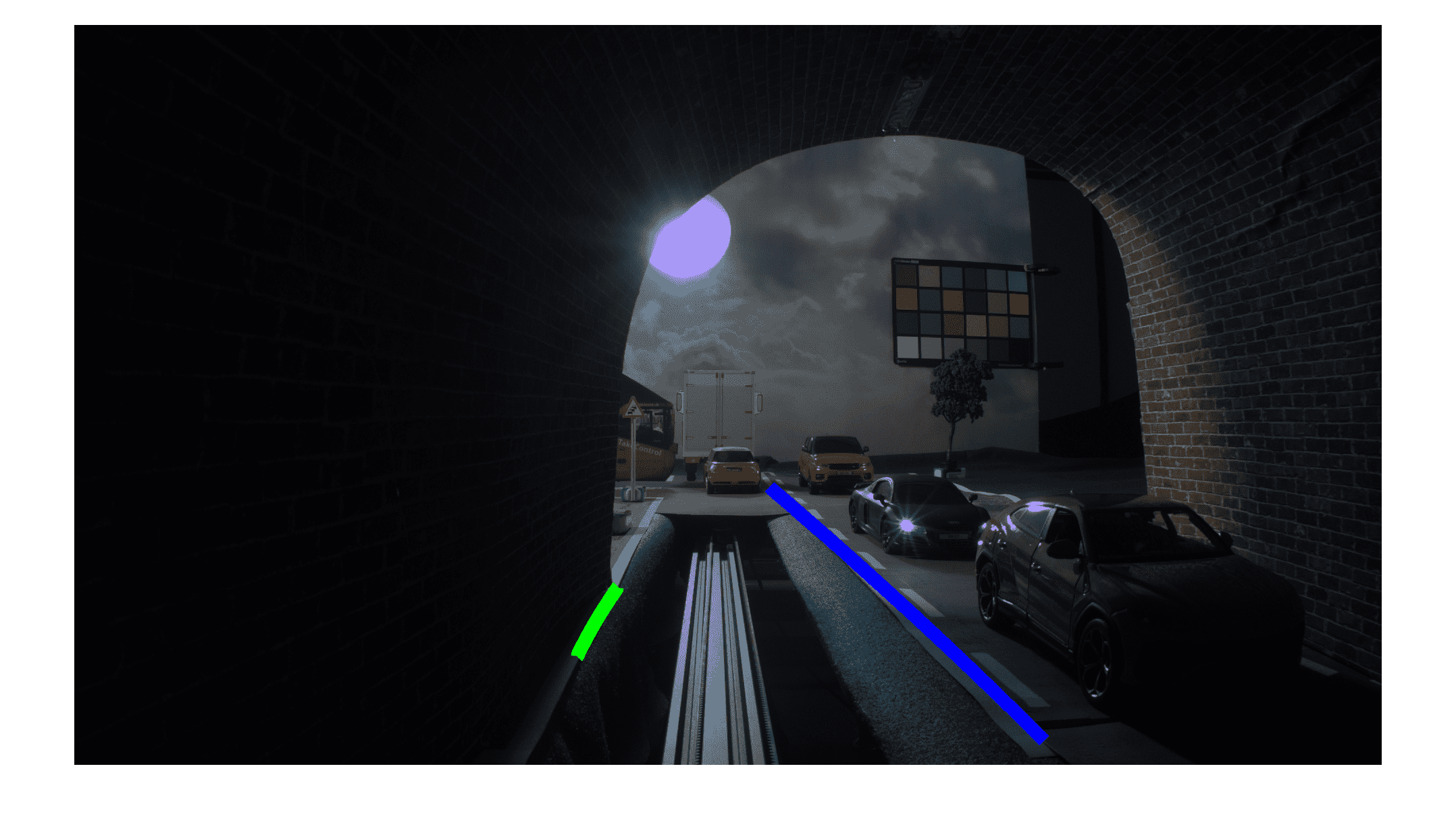}}
\end{subfigure}\hfill
\begin{subfigure}{1.3in}
\footnotesize
\stackunder[5pt]{\includegraphics[trim={75 70 75 75 },clip,width=1.3in,height=1.3in,keepaspectratio]{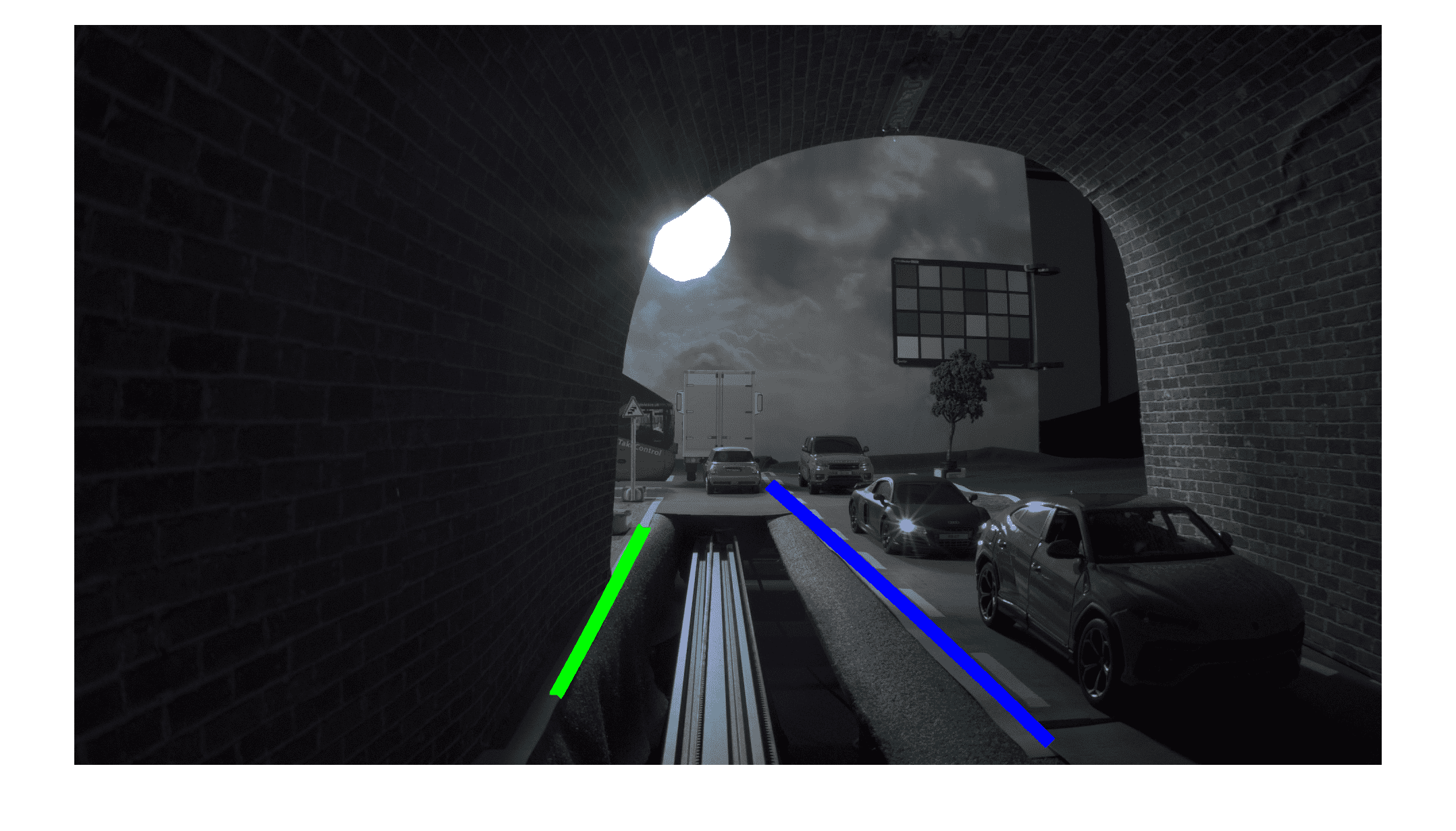}}{Top}
\end{subfigure}\hfill
\begin{subfigure}{1.3in}
\subfloat{\includegraphics[trim={75 70 75 75 },clip,width=1.3in,height=1.3in,keepaspectratio]{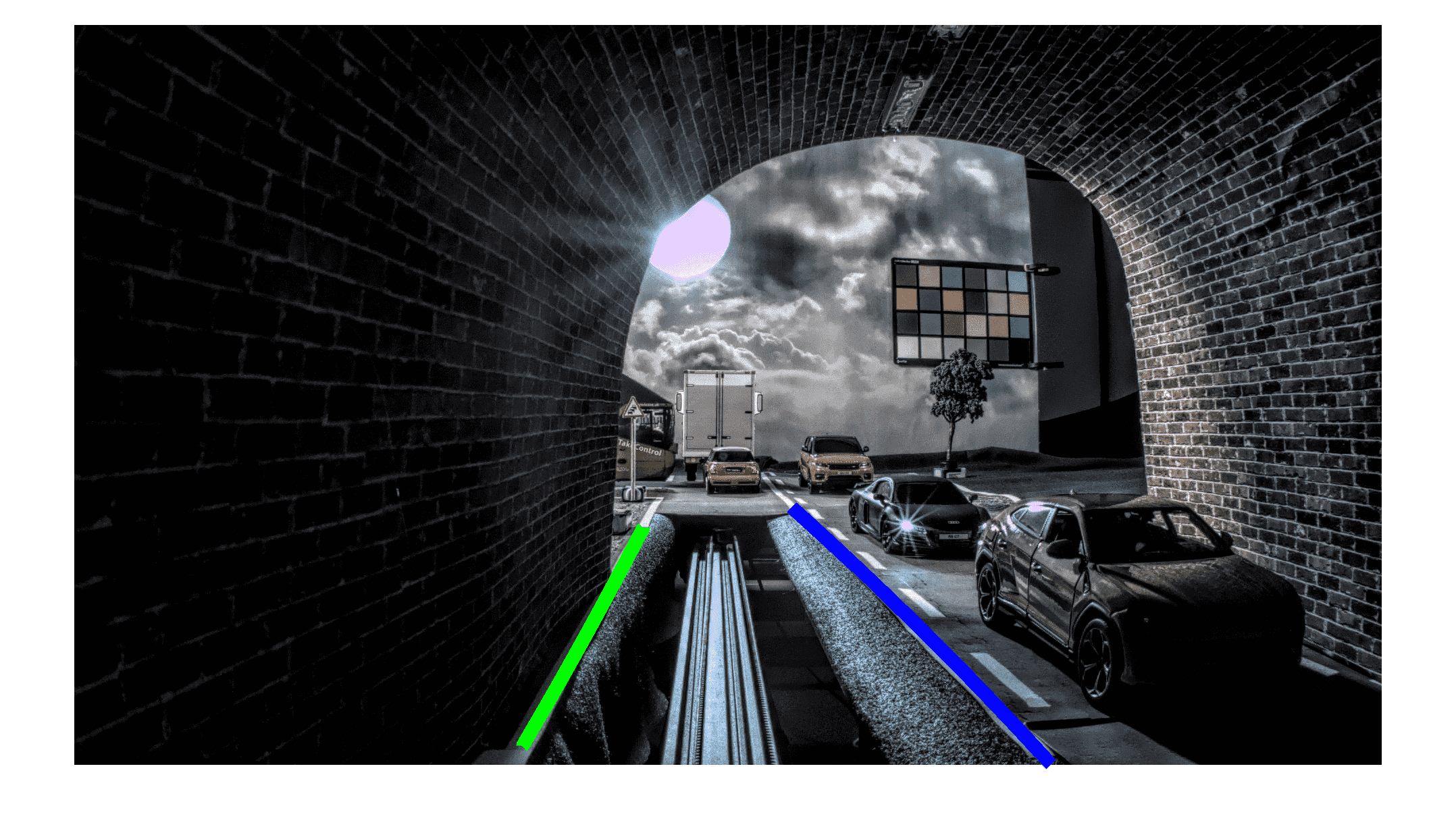}}
\end{subfigure}\hfill
\begin{subfigure}{1.3in}
\subfloat{\includegraphics[trim={75 70 75 75},clip,width=1.3in,height=1.3in,keepaspectratio]{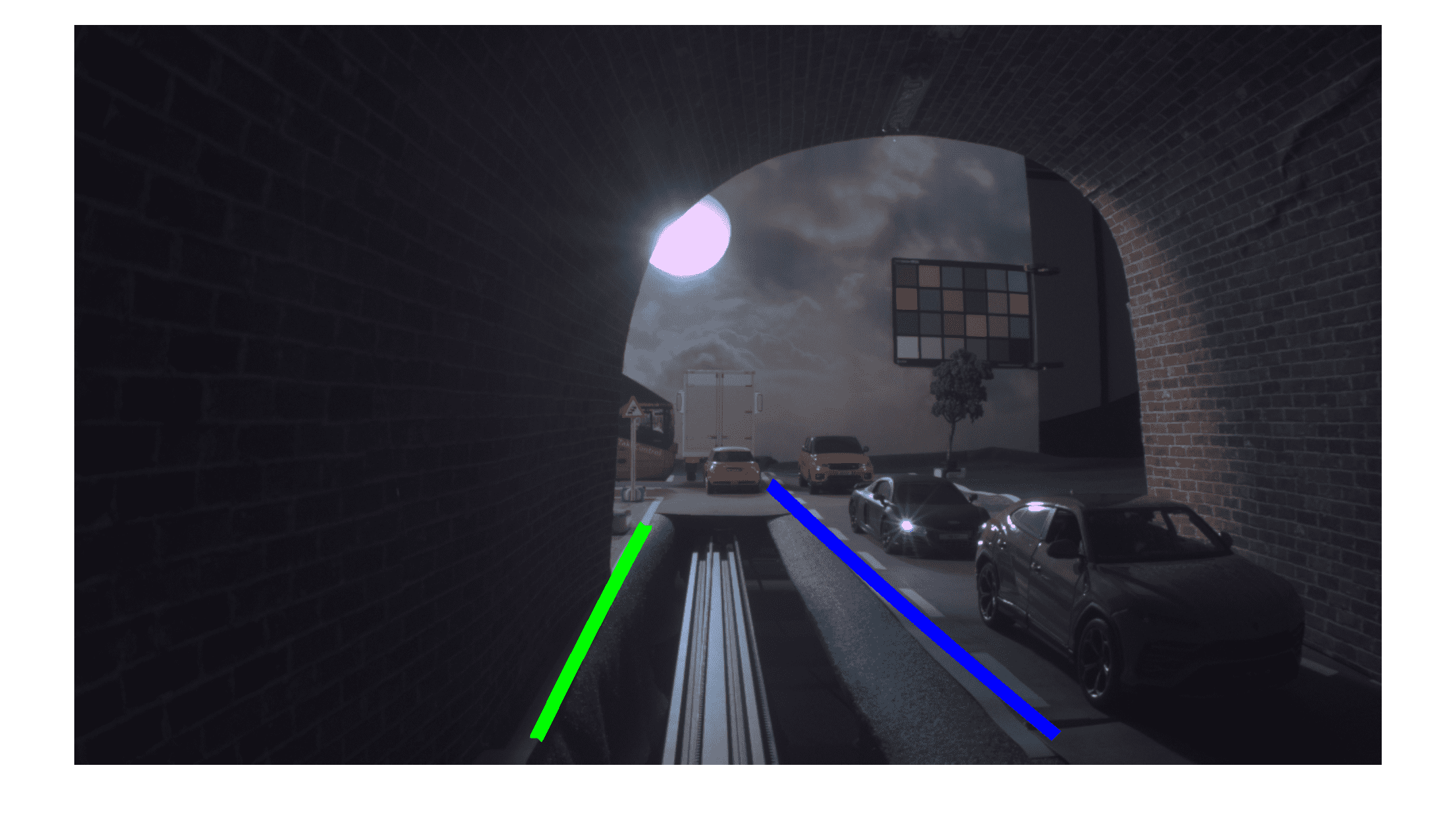}}
\end{subfigure}\hfill
\begin{subfigure}{1.3in}
\subfloat{\includegraphics[trim={75 70 75 75},clip,width=1.3in,height=1.3in,keepaspectratio]{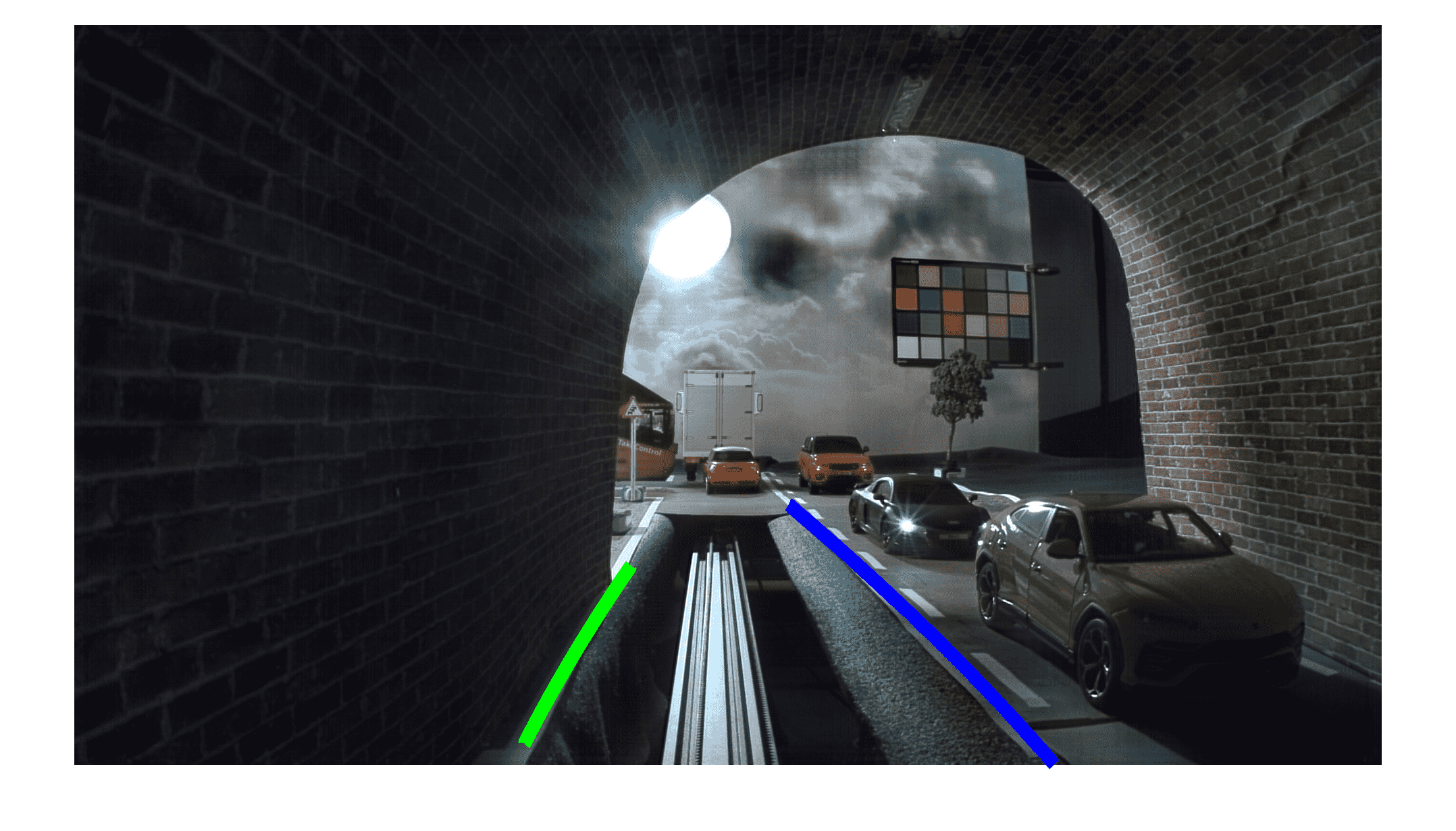}}
\end{subfigure}\hfill
\begin{subfigure}{1.3in}
\subfloat{\includegraphics[trim={75 70 75 75},clip,width=1.3in,height=1.3in,keepaspectratio]{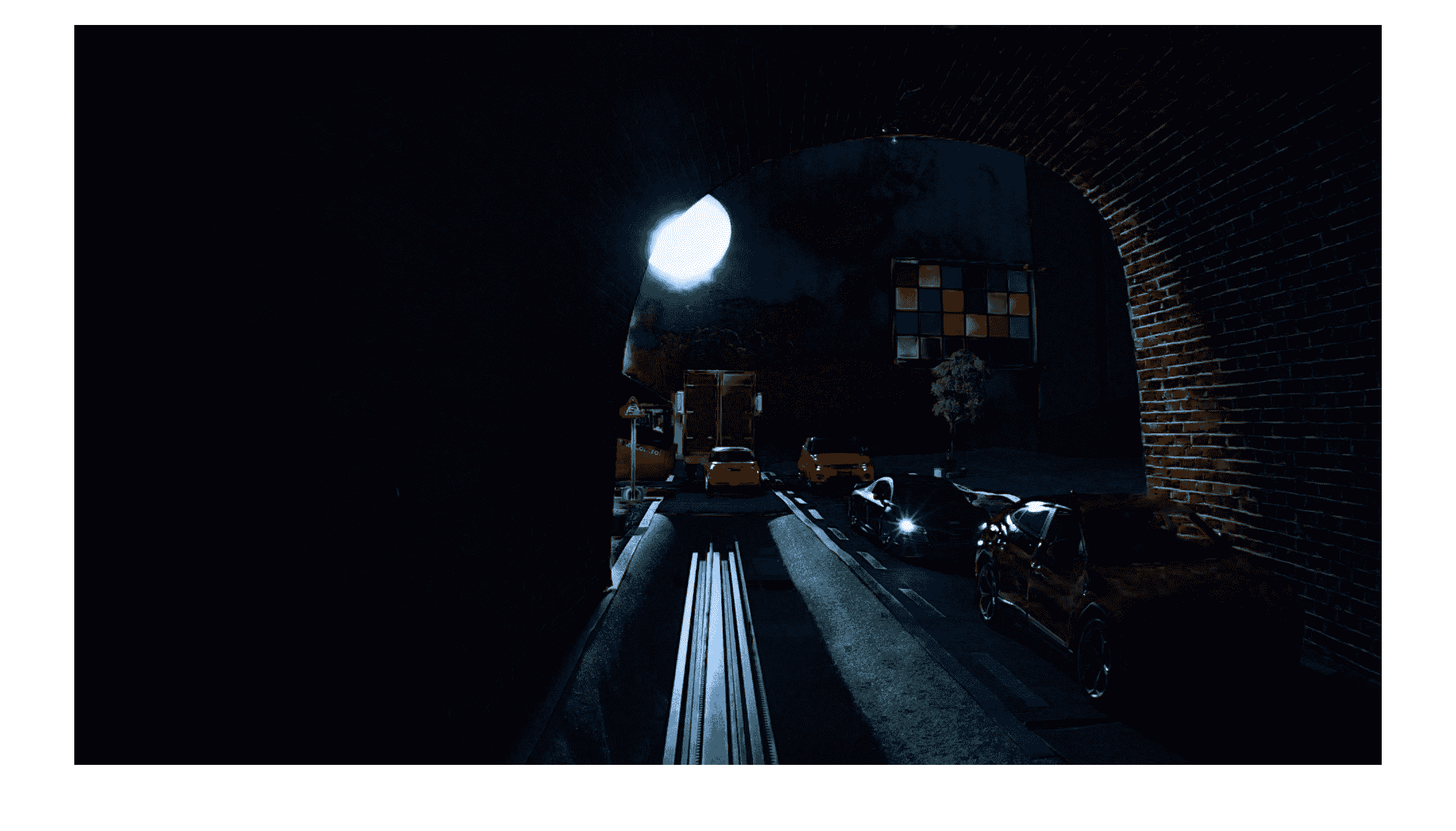}}
\end{subfigure}\hfill
\begin{subfigure}{1.3in}
\footnotesize
\stackunder[5pt]{\includegraphics[trim={75 70 75 75},clip,width=1.3in,height=1.3in,keepaspectratio]{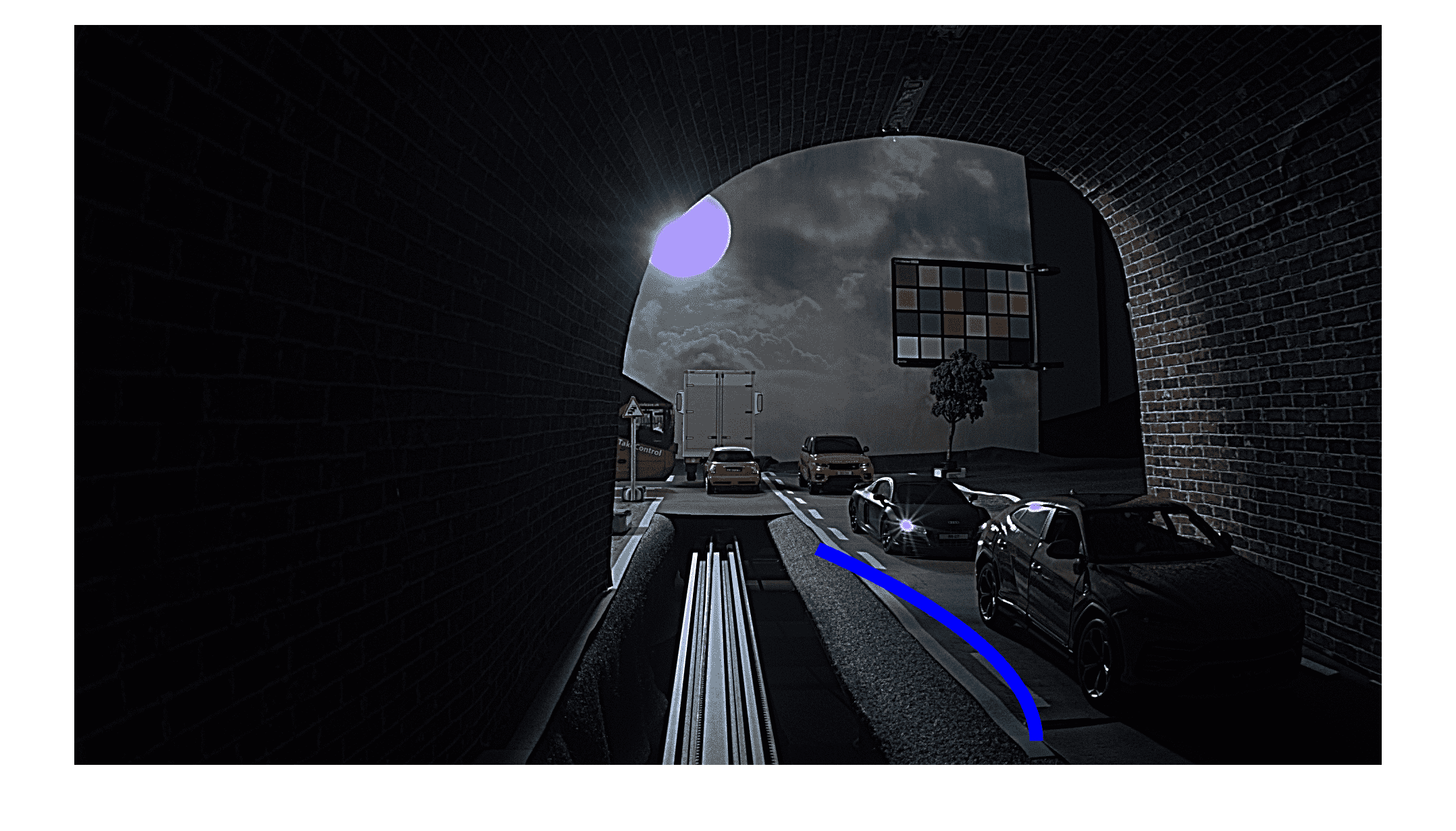}}{Bottom}
\end{subfigure}\hfill
\begin{subfigure}{1.3in}
\subfloat{\includegraphics[trim={75 70 75 75},clip,width=1.3in,height=1.3in,keepaspectratio]{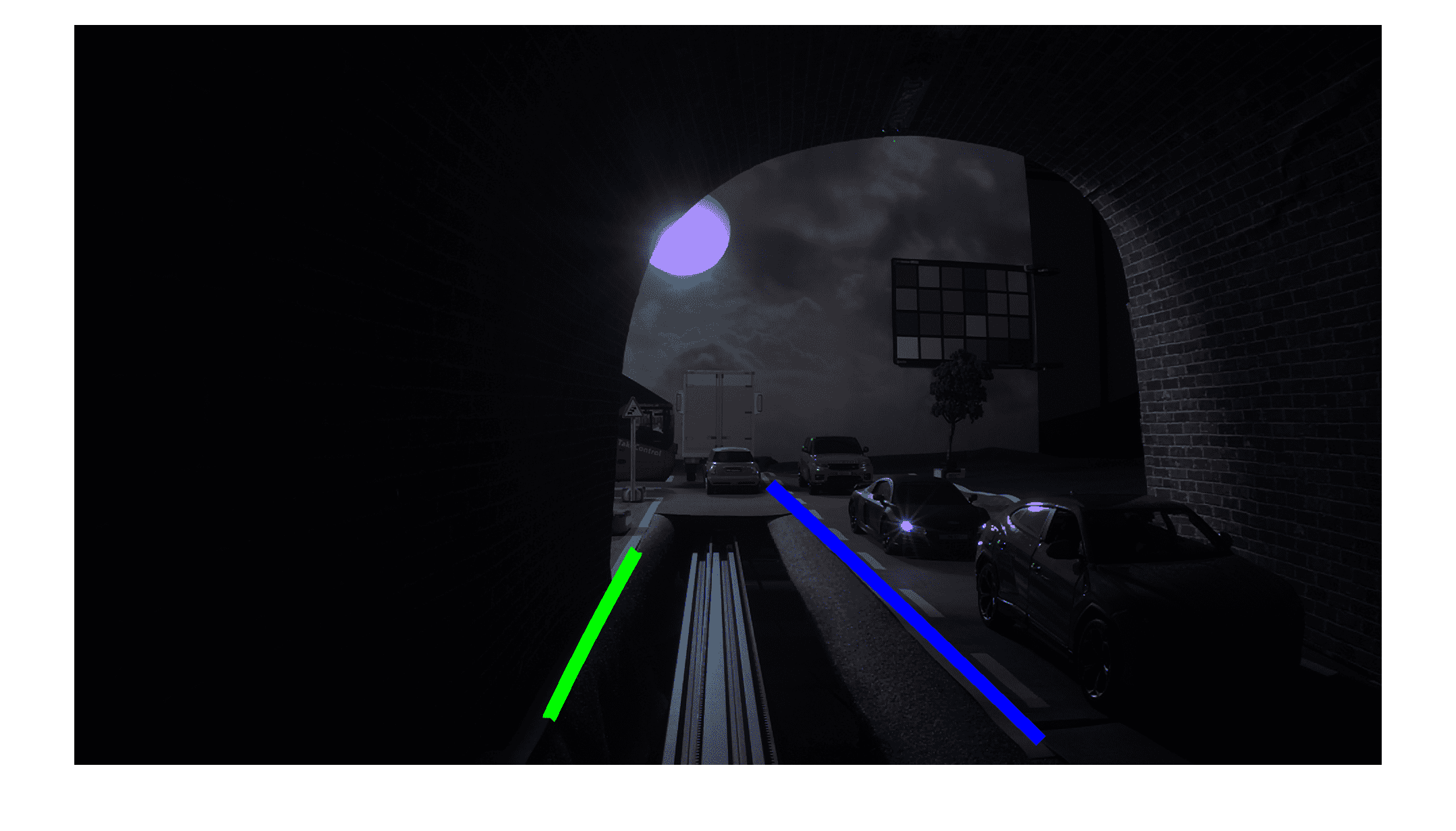}}
\end{subfigure}\hfill
\begin{subfigure}{1.3in}
\subfloat{\includegraphics[trim={75 70 75 75},clip,width=1.3in,height=1.3in,keepaspectratio]{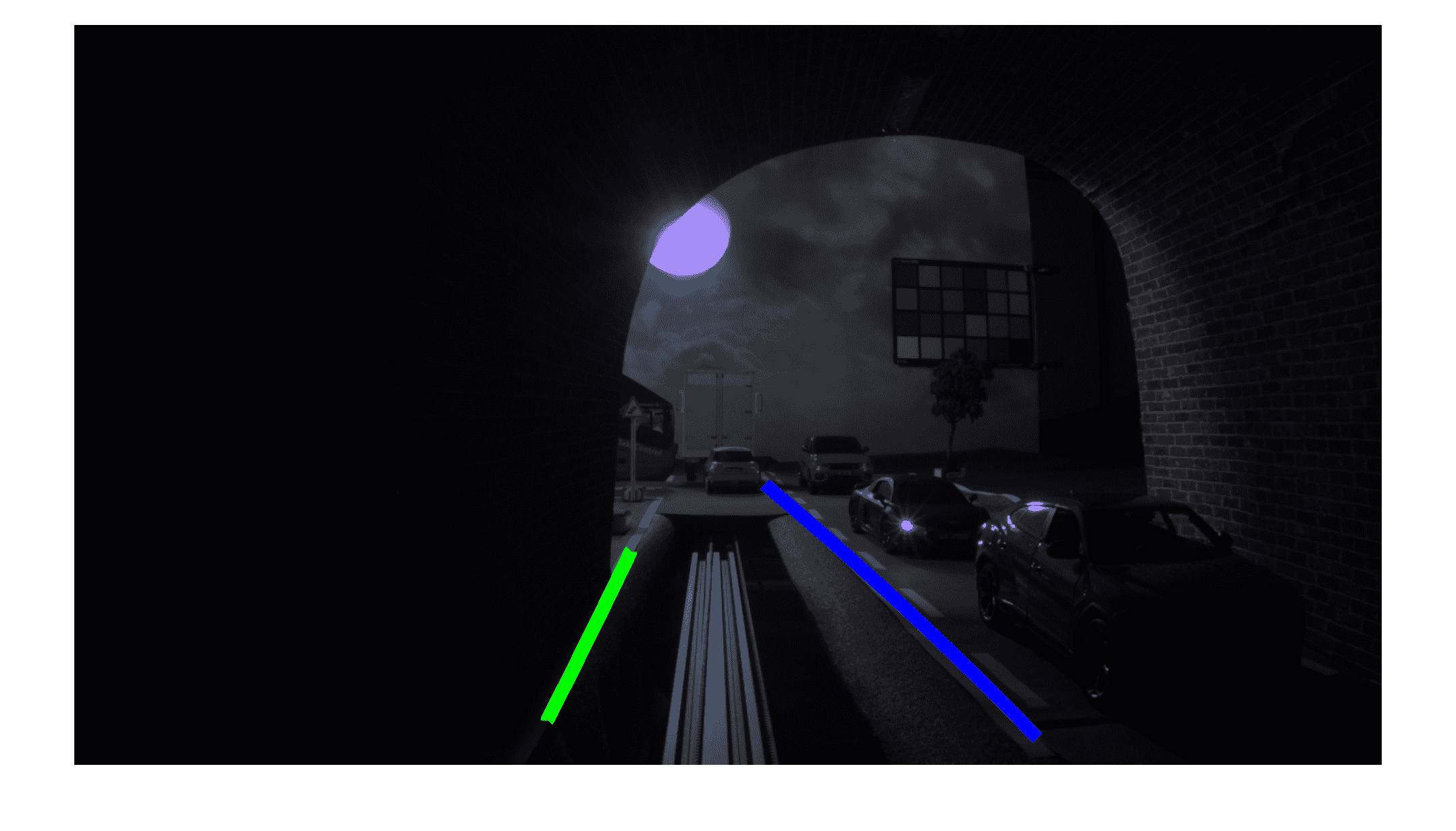}}
\end{subfigure}\hfill
\begin{subfigure}{1.3in}
\subfloat{\includegraphics[width=1.3in,height=1.3in,keepaspectratio]{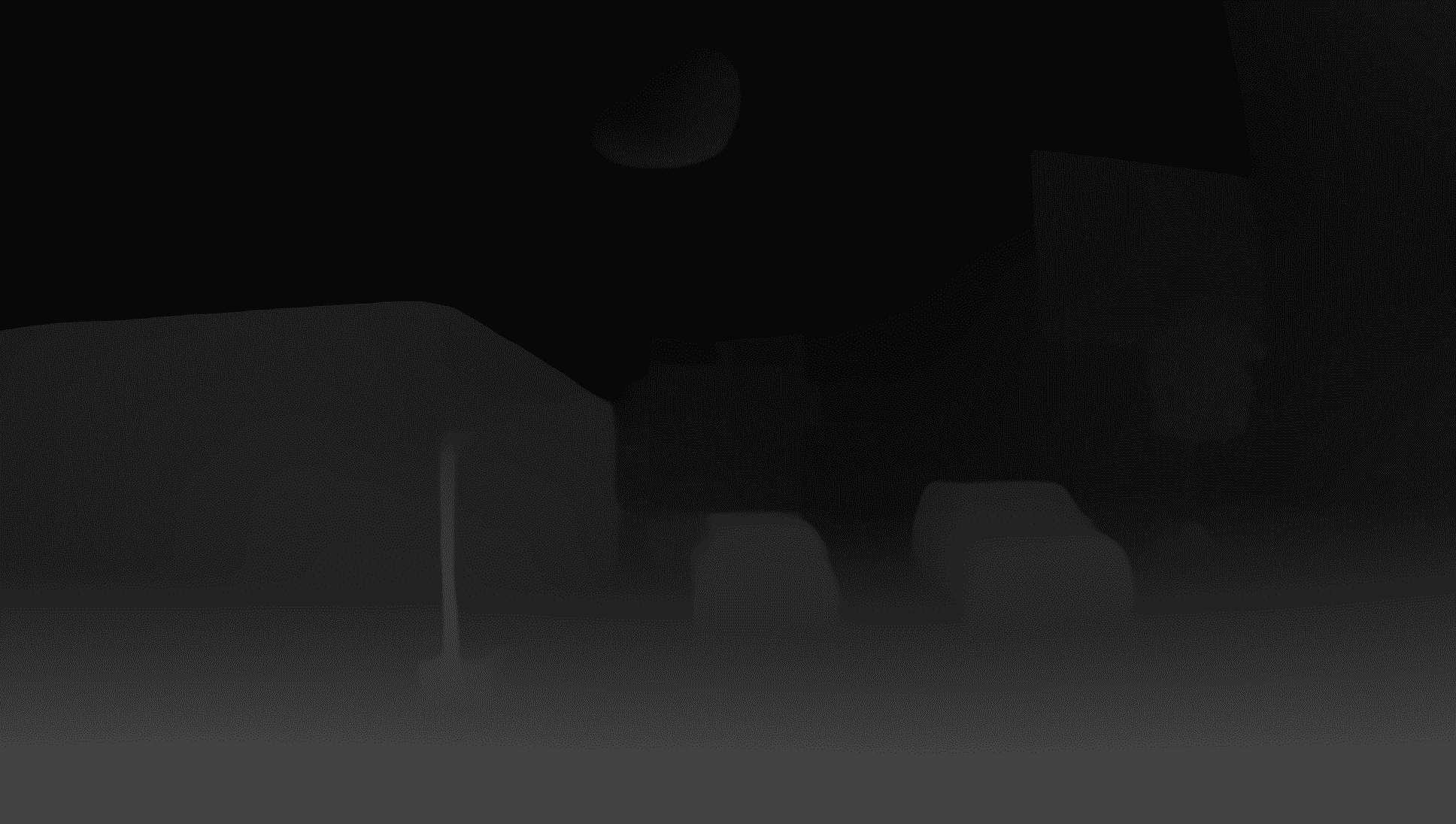}}
\end{subfigure}\hfill
\begin{subfigure}{1.3in}
\subfloat{\includegraphics[width=1.3in,height=1.3in,keepaspectratio]{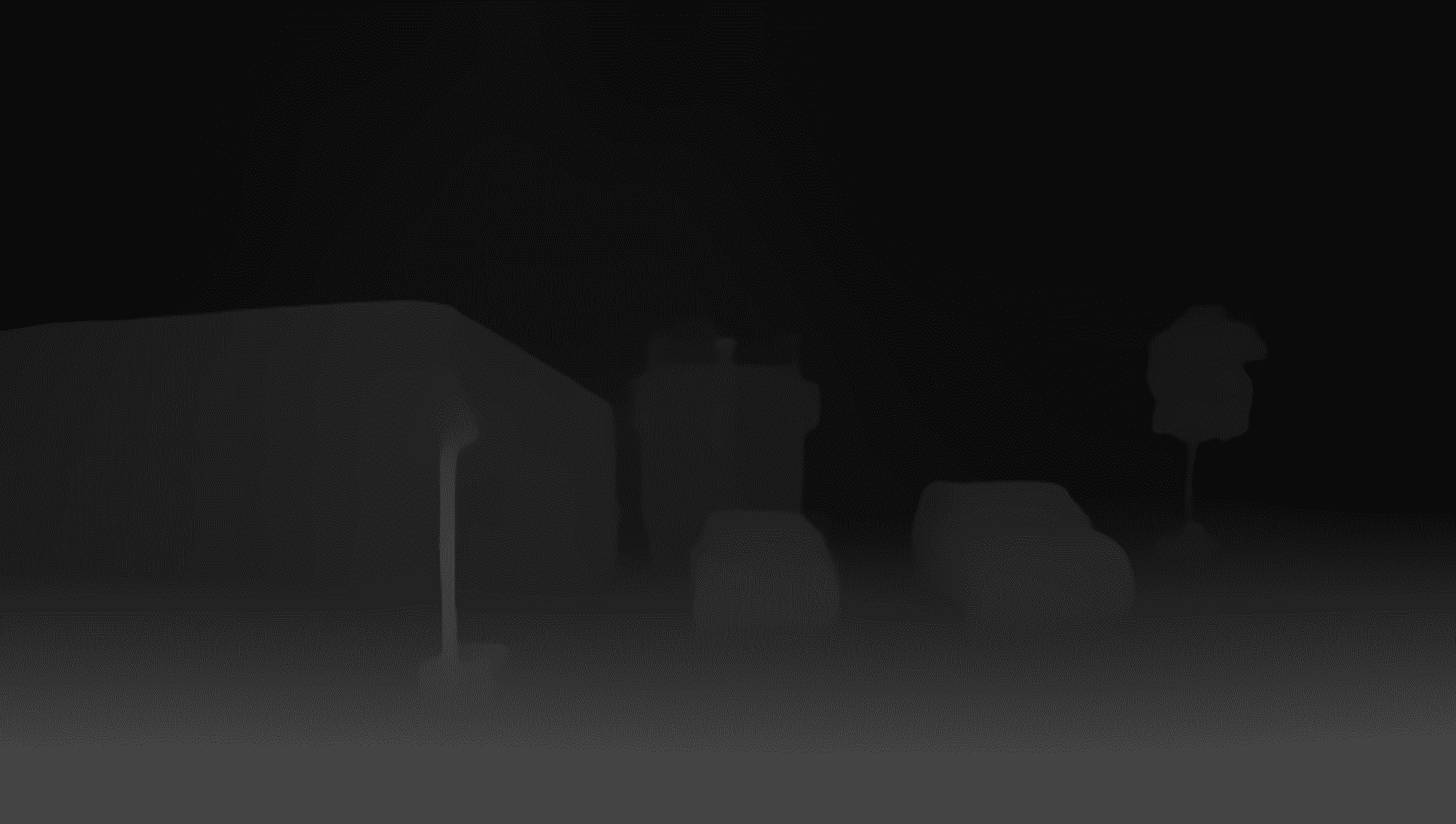}}
\end{subfigure}\hfill
\begin{subfigure}{1.3in}
\footnotesize
\stackunder[5pt]{\includegraphics[width=1.3in,height=1.3in,keepaspectratio]{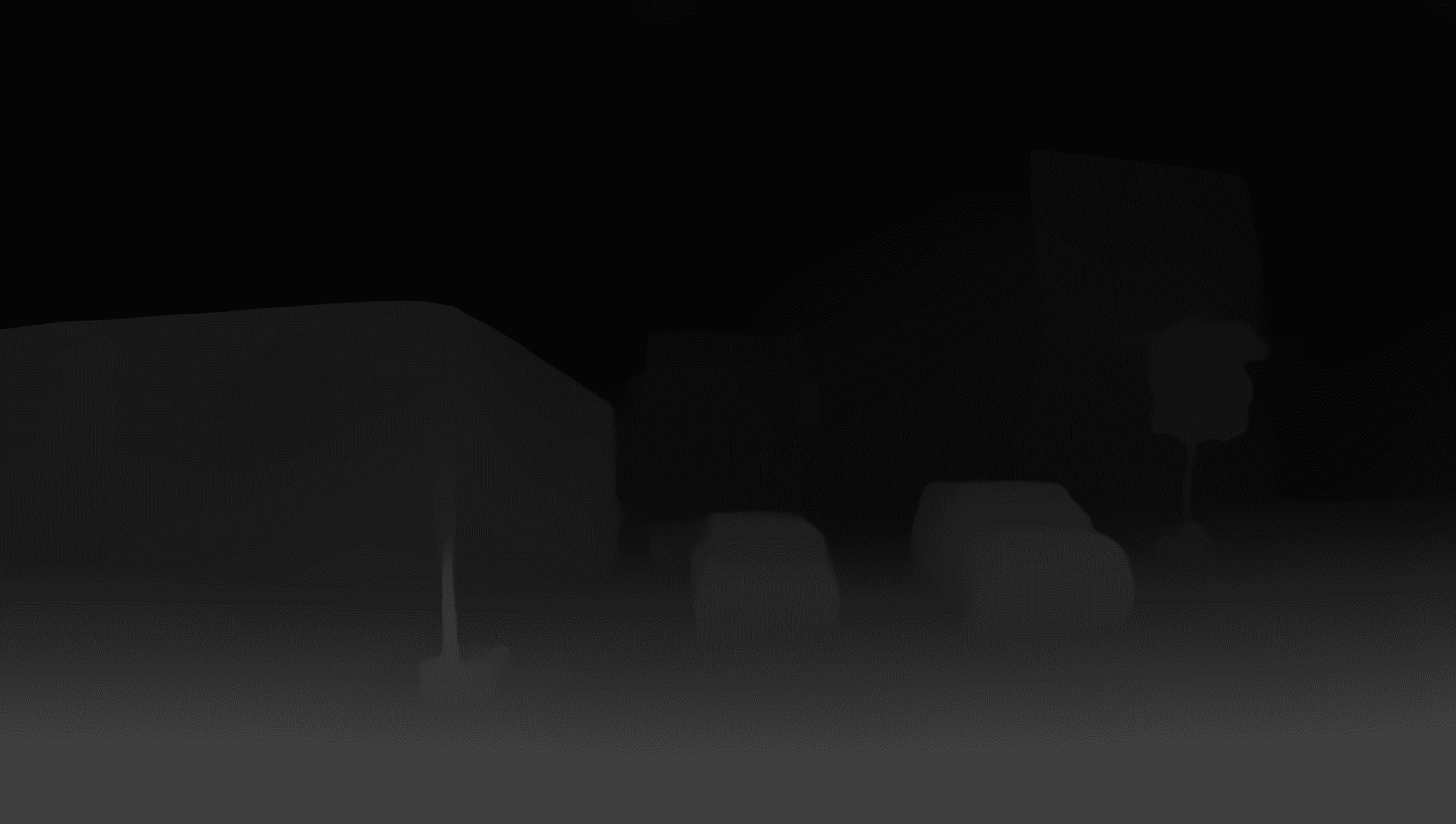}}{Top}
\end{subfigure}\hfill
\begin{subfigure}{1.3in}
\subfloat{\includegraphics[width=1.3in,height=1.3in,keepaspectratio]{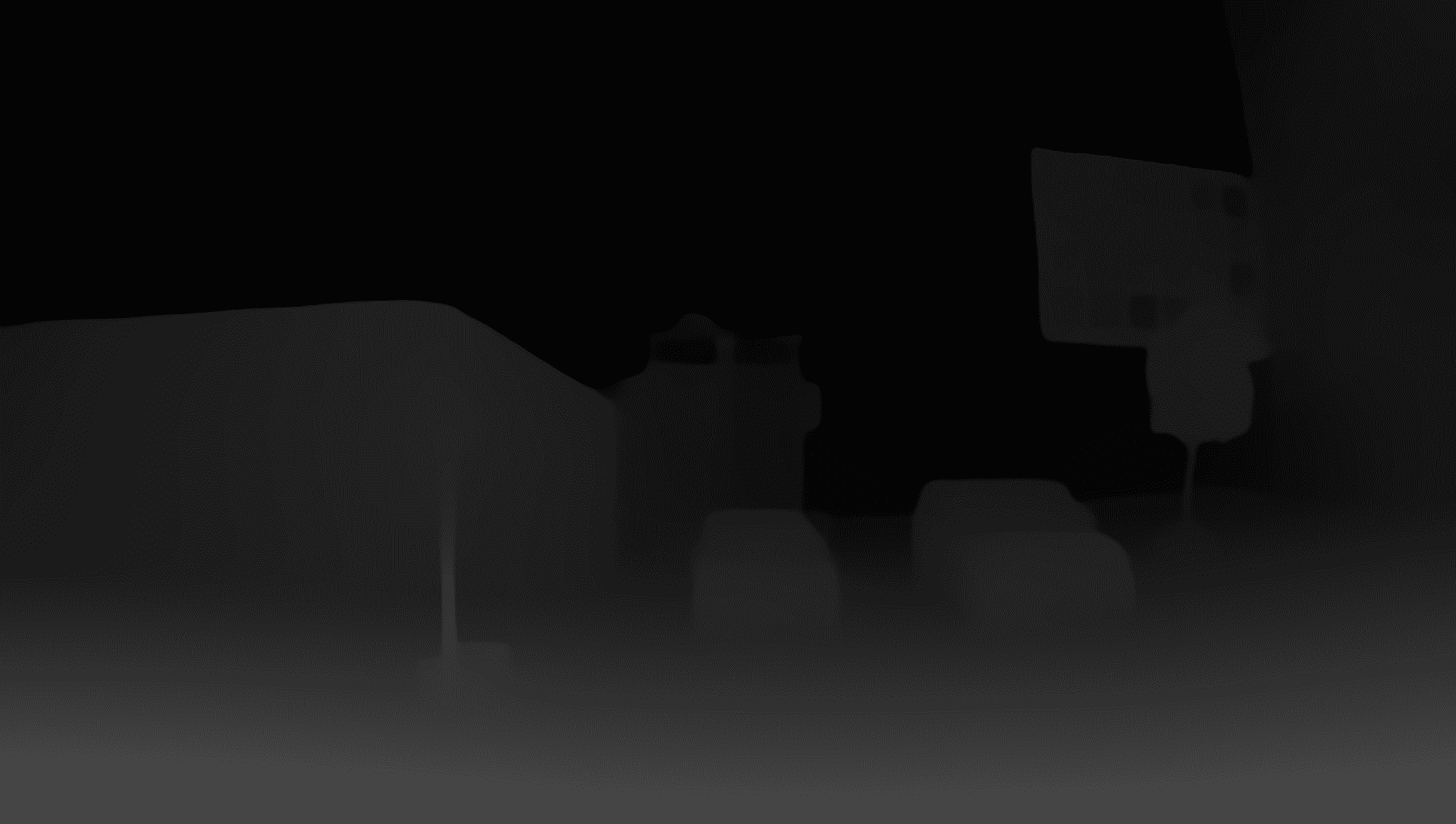}}
\end{subfigure}\hfill
\begin{subfigure}{1.3in}
\subfloat{\includegraphics[width=1.3in,height=1.3in,keepaspectratio]{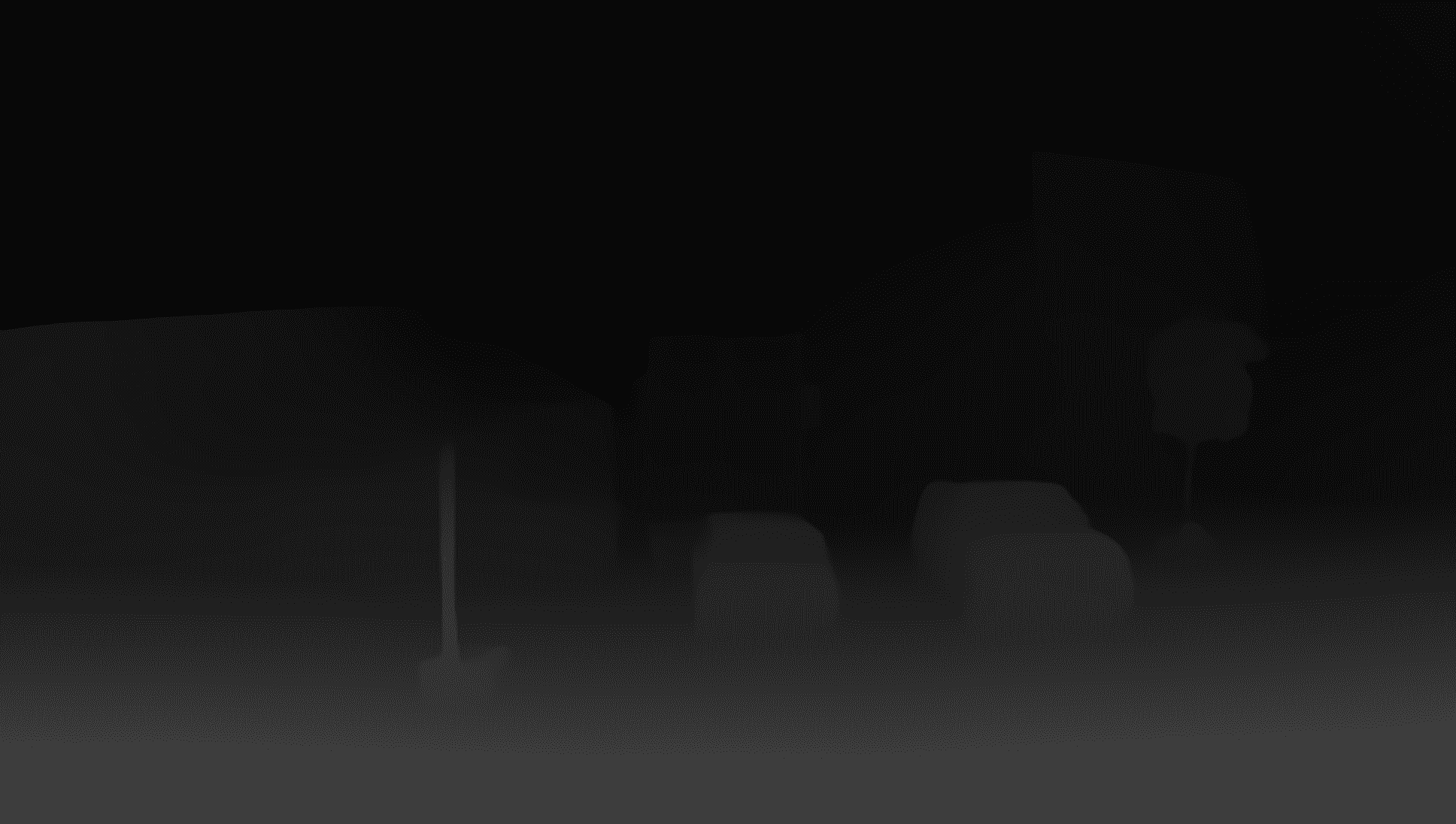}}
\end{subfigure}\hfill
\begin{subfigure}{1.3in}
\subfloat{\includegraphics[width=1.3in,height=1.3in,keepaspectratio]{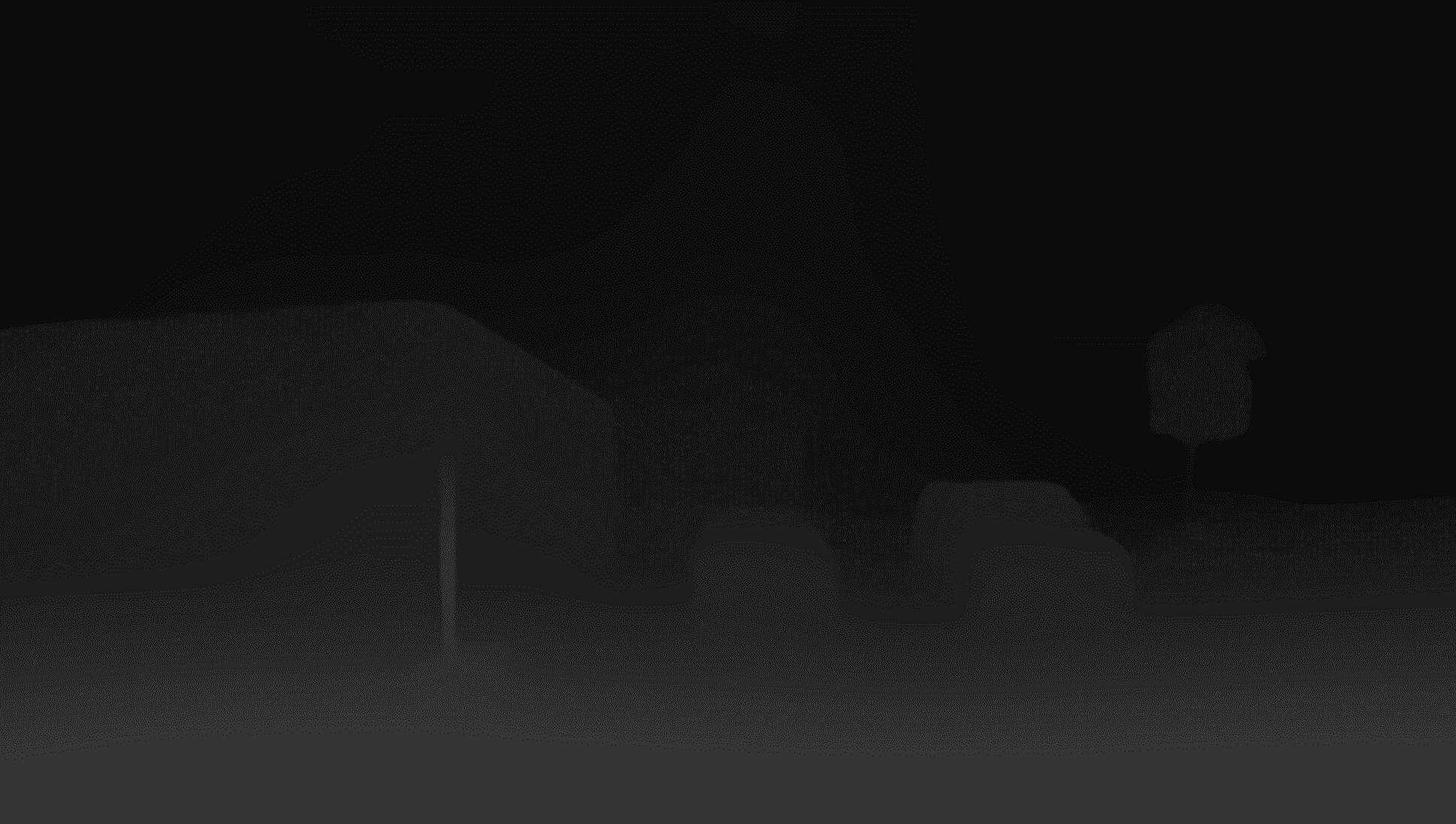}}
\end{subfigure}\hfill
\begin{subfigure}{1.3in}
\subfloat{\includegraphics[width=1.3in,height=1.3in,keepaspectratio]{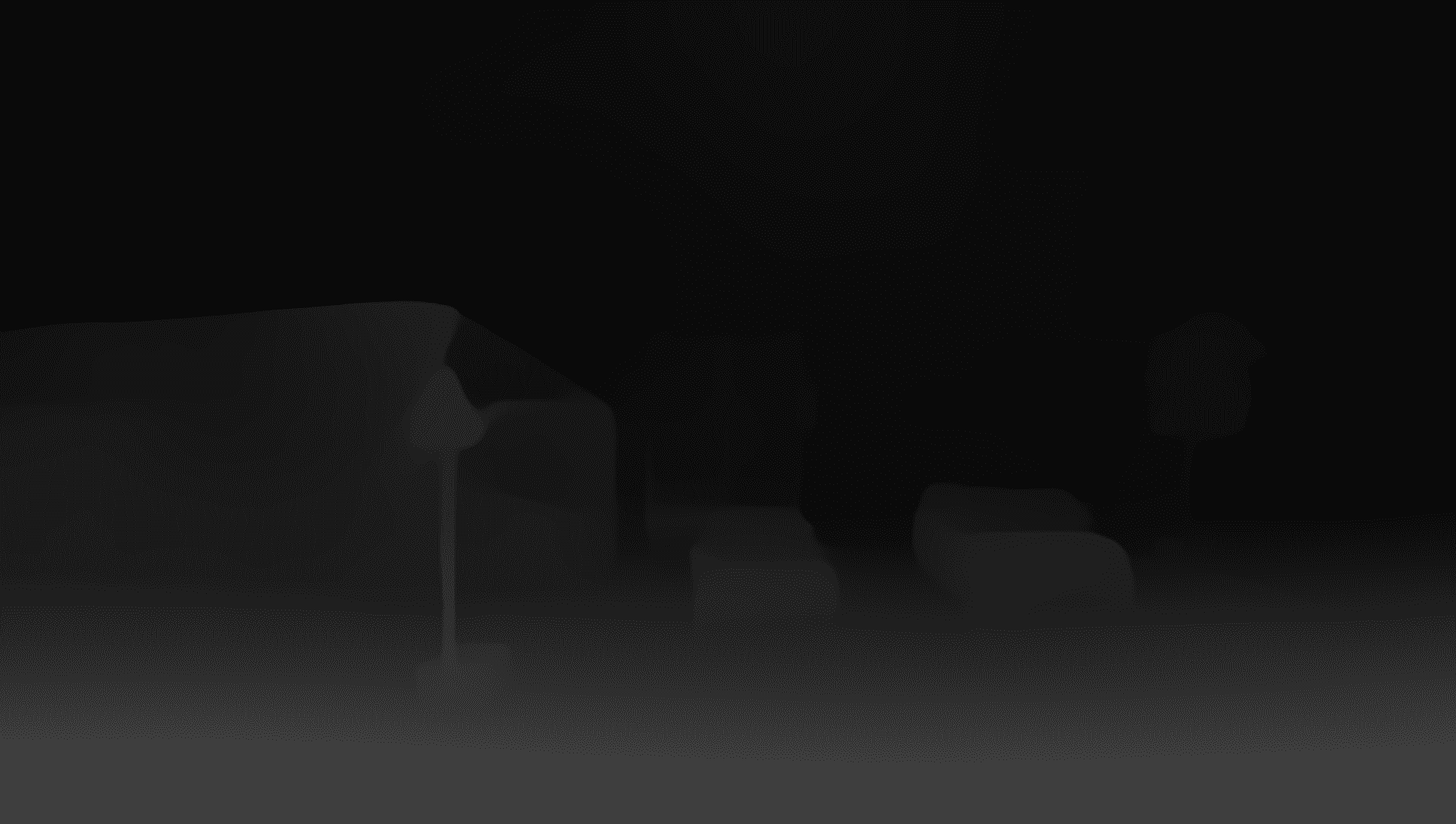}}
\end{subfigure}\hfill
\begin{subfigure}{1.3in}
\footnotesize
\stackunder[5pt]{\includegraphics[width=1.3in,height=1.3in,keepaspectratio]{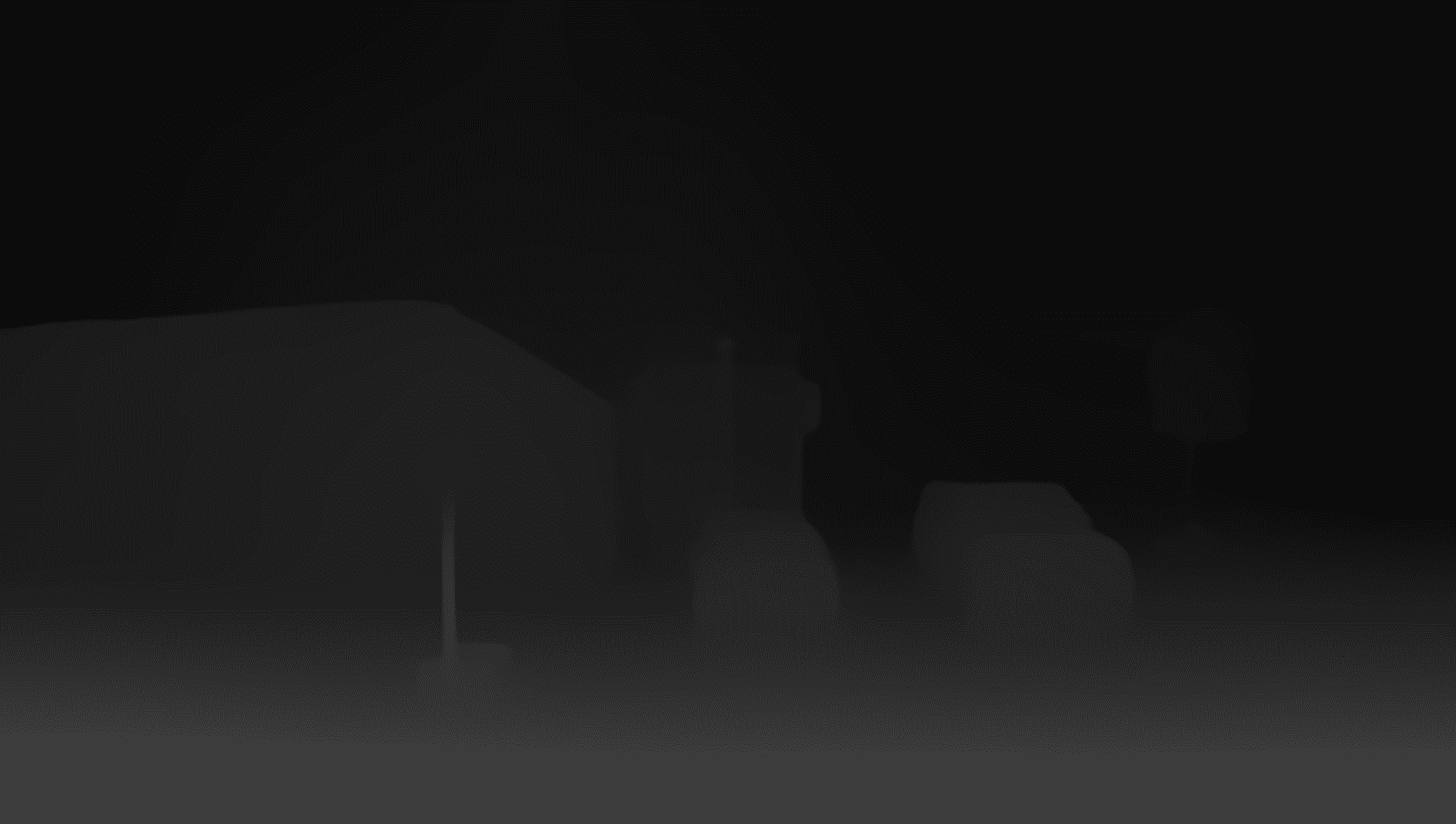}}{Bottom}
\end{subfigure}\hfill
\begin{subfigure}{1.3in}
\subfloat{\includegraphics[width=1.3in,height=1.3in,keepaspectratio]{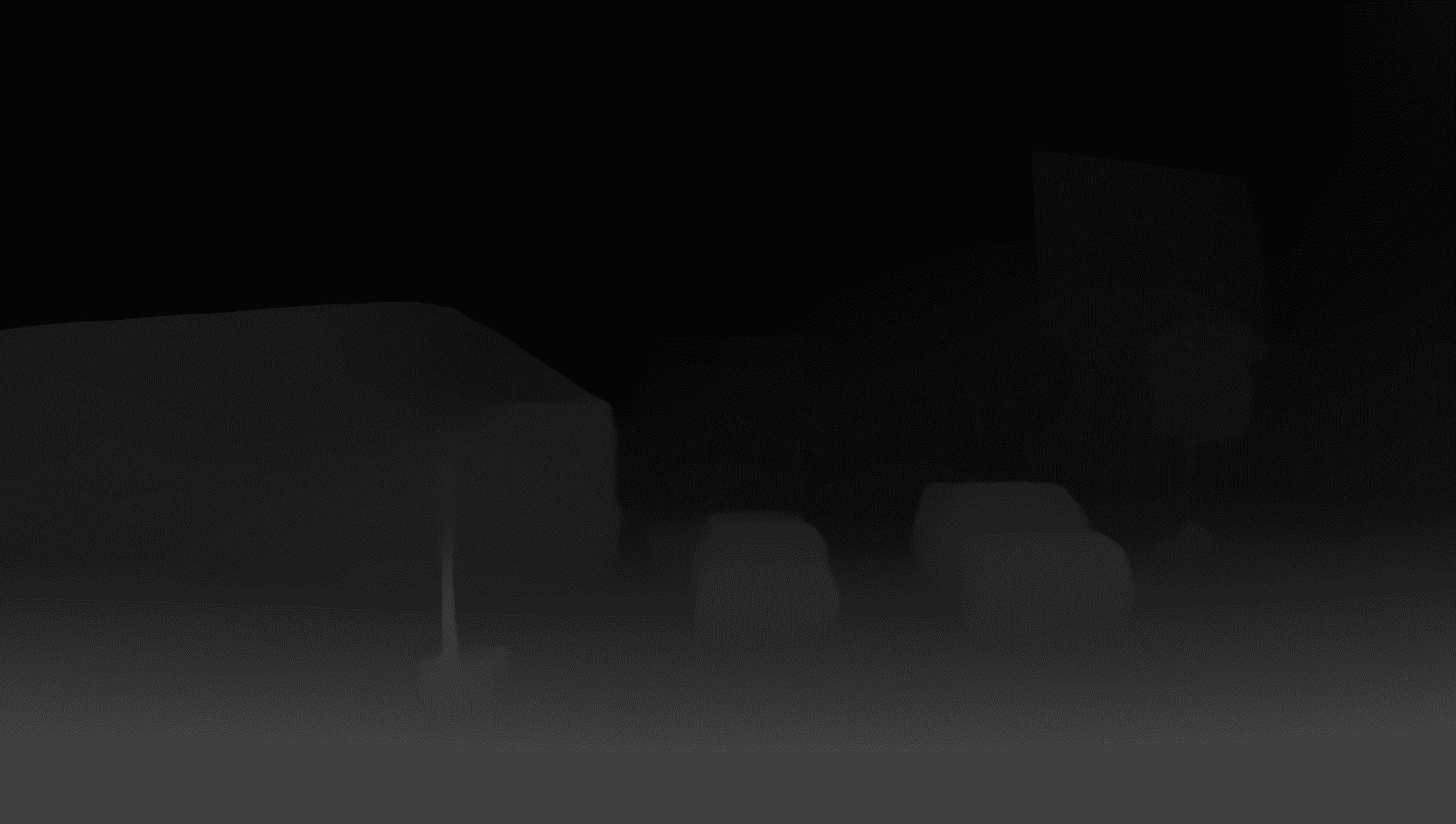}}
\end{subfigure}\hfill
\begin{subfigure}{1.3in}
\subfloat{\includegraphics[width=1.3in,height=1.3in,keepaspectratio]{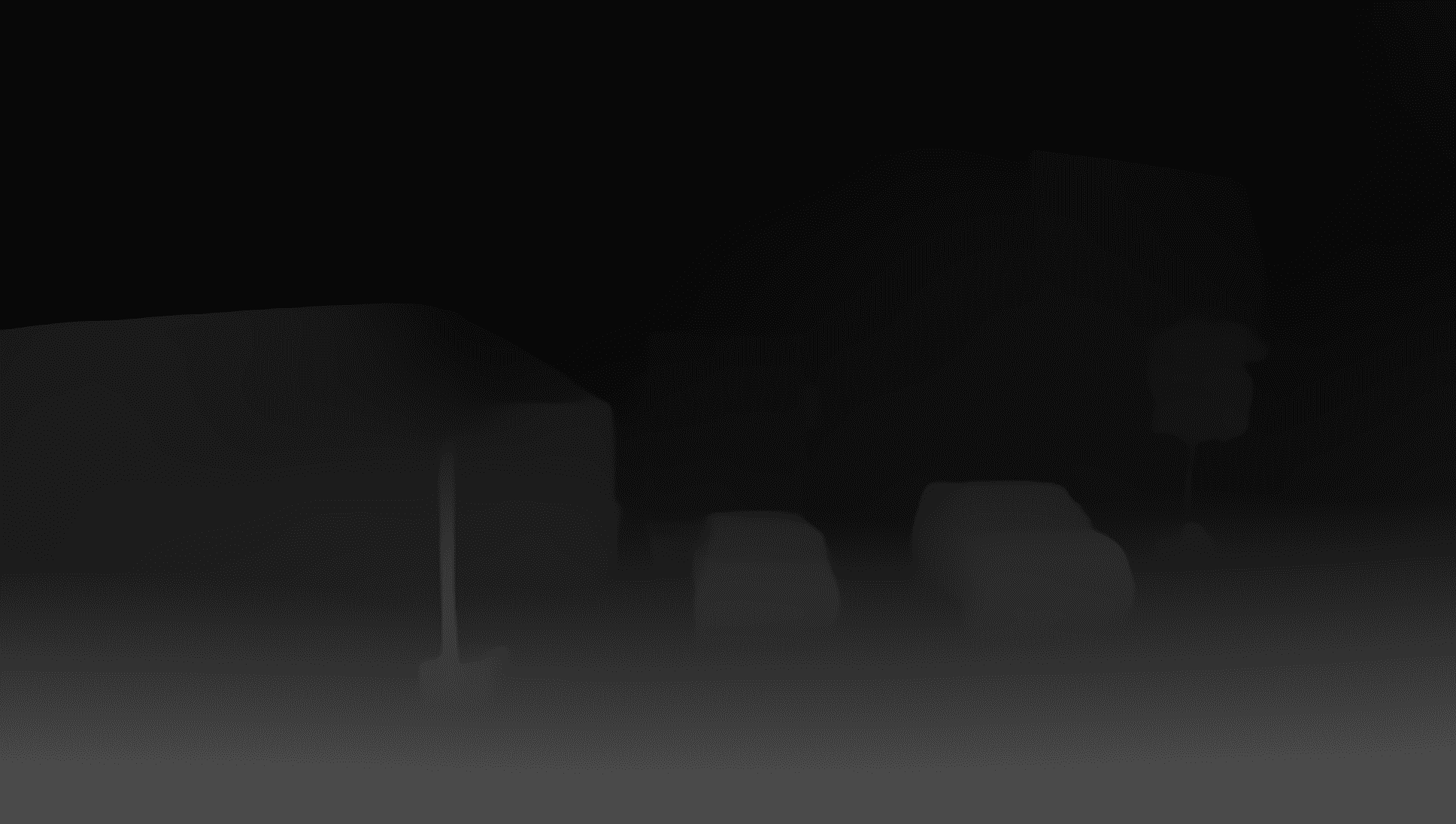}}
\end{subfigure}\hfill
\caption{Visual comparison of the output of various CV algorithms used to produce the results in Fig. \ref{fig:Cv_comp}. (Top: Left to Right) Ground truth, Input, Proposed, Laplacian Filter, Reflection removal \cite{Refl}. (Bottom: Left to Right) PFFNet \cite{mei2018pffn}, C2PNet \cite{zheng2023curricular}, Unsharp masking, Deblur \cite{Deblur_chen}, Wiener Filter. (Row 1 \& 2) Object detection\ recognition. (Row 3 \& 4) Lane detection. (Row 5 \& 6) Depth estimation. }
\label{fig:Cv_compviz}
\end{figure*}

\begin{figure*}[!h]
\centering
\raisebox{.9\height}{ \rotatebox[origin=]{90}{Top}}
\begin{subfigure}{1.3in}
{\includegraphics[width=1.3in,height=1.3in,keepaspectratio]{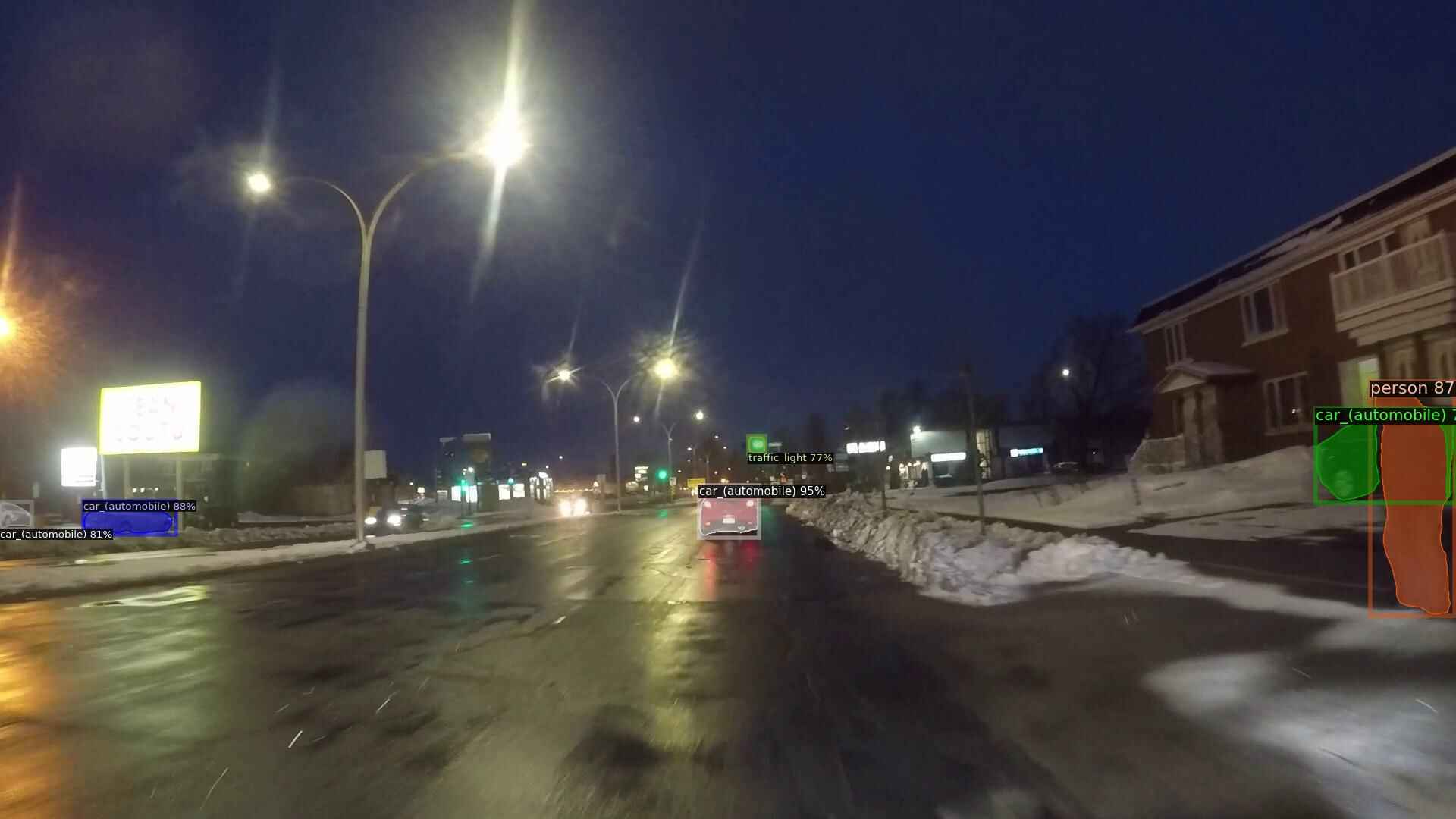}}
\end{subfigure}
\begin{subfigure}{1.3in}
{\includegraphics[width=1.3in,height=1.3in,keepaspectratio]{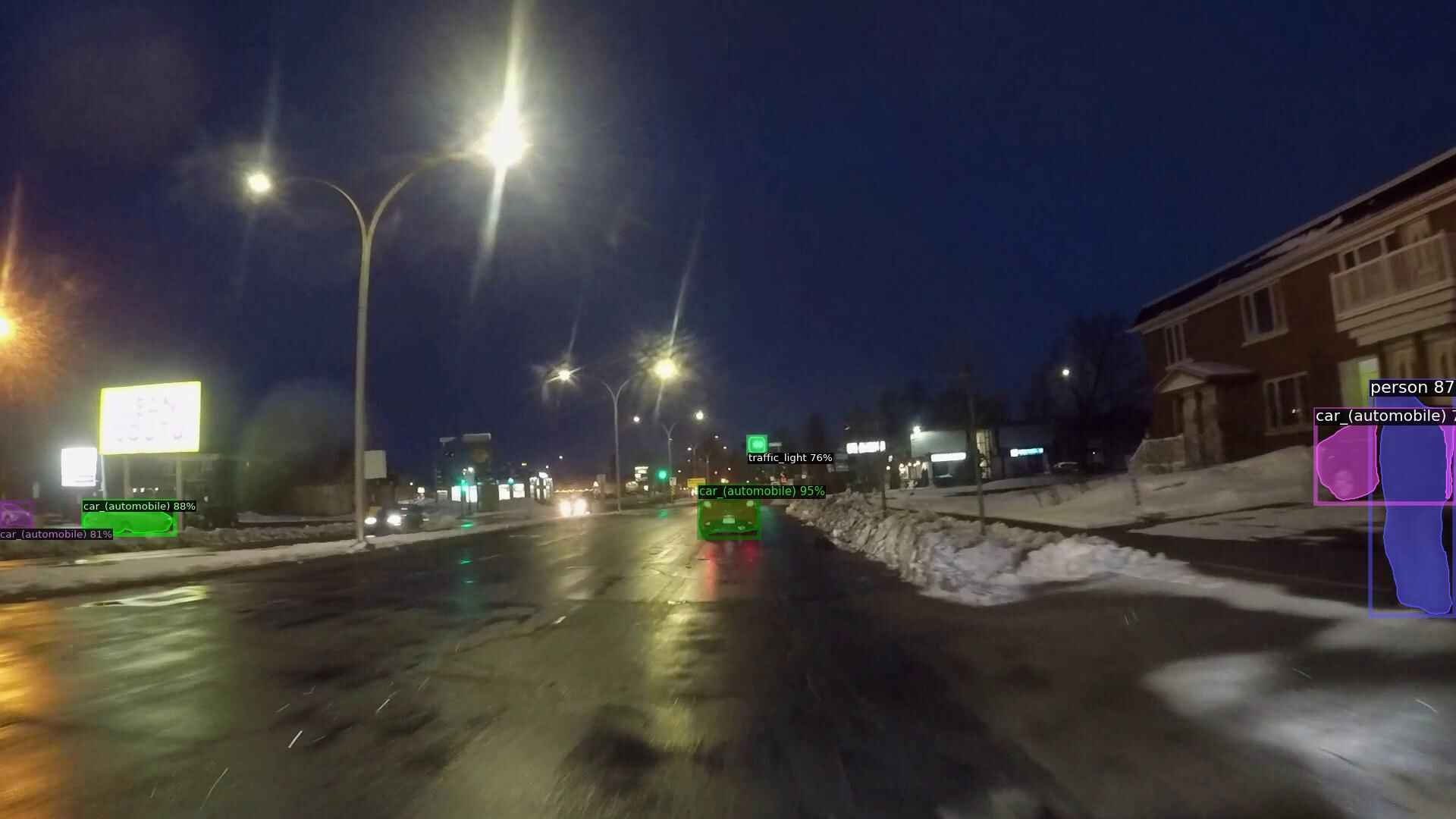}}
\end{subfigure}
\begin{subfigure}{1.3in}
{\includegraphics[width=1.3in,height=1.3in,keepaspectratio]{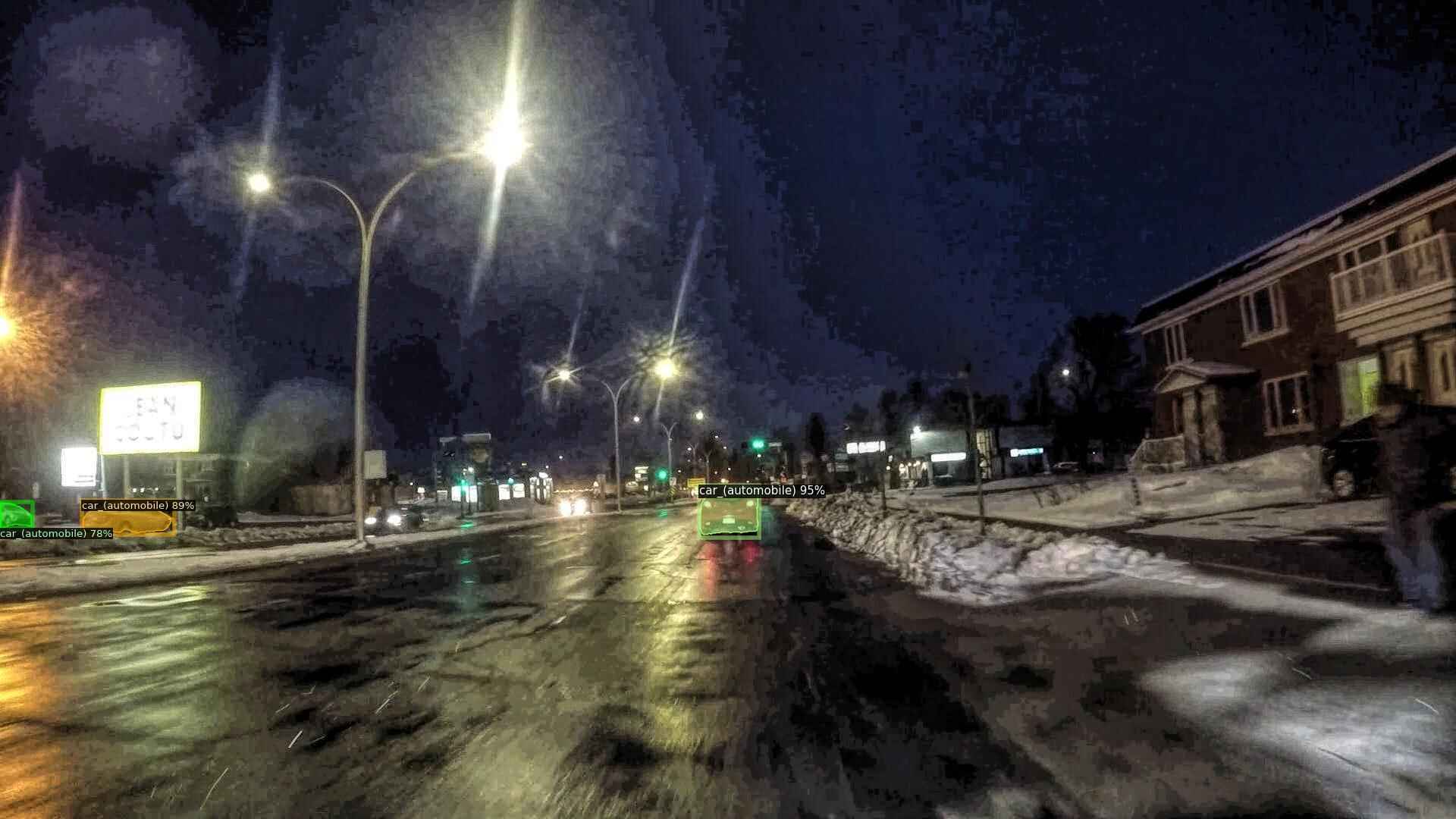}}
\end{subfigure}
\begin{subfigure}{1.3in}
{\includegraphics[width=1.3in,height=1.3in,keepaspectratio]{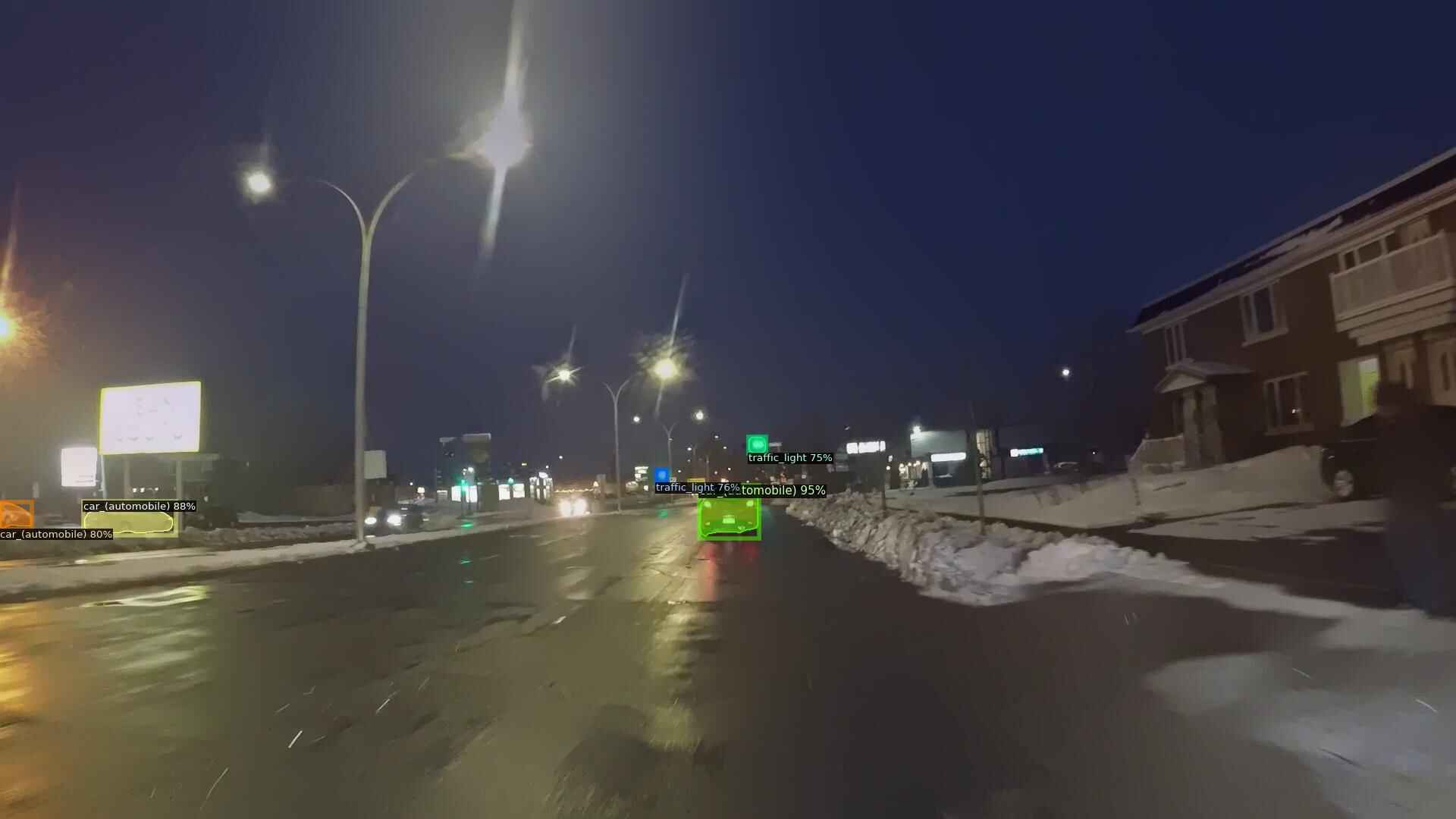}}
\end{subfigure}
\begin{subfigure}{1.3in}
{\includegraphics[width=1.3in,height=1.3in,keepaspectratio]{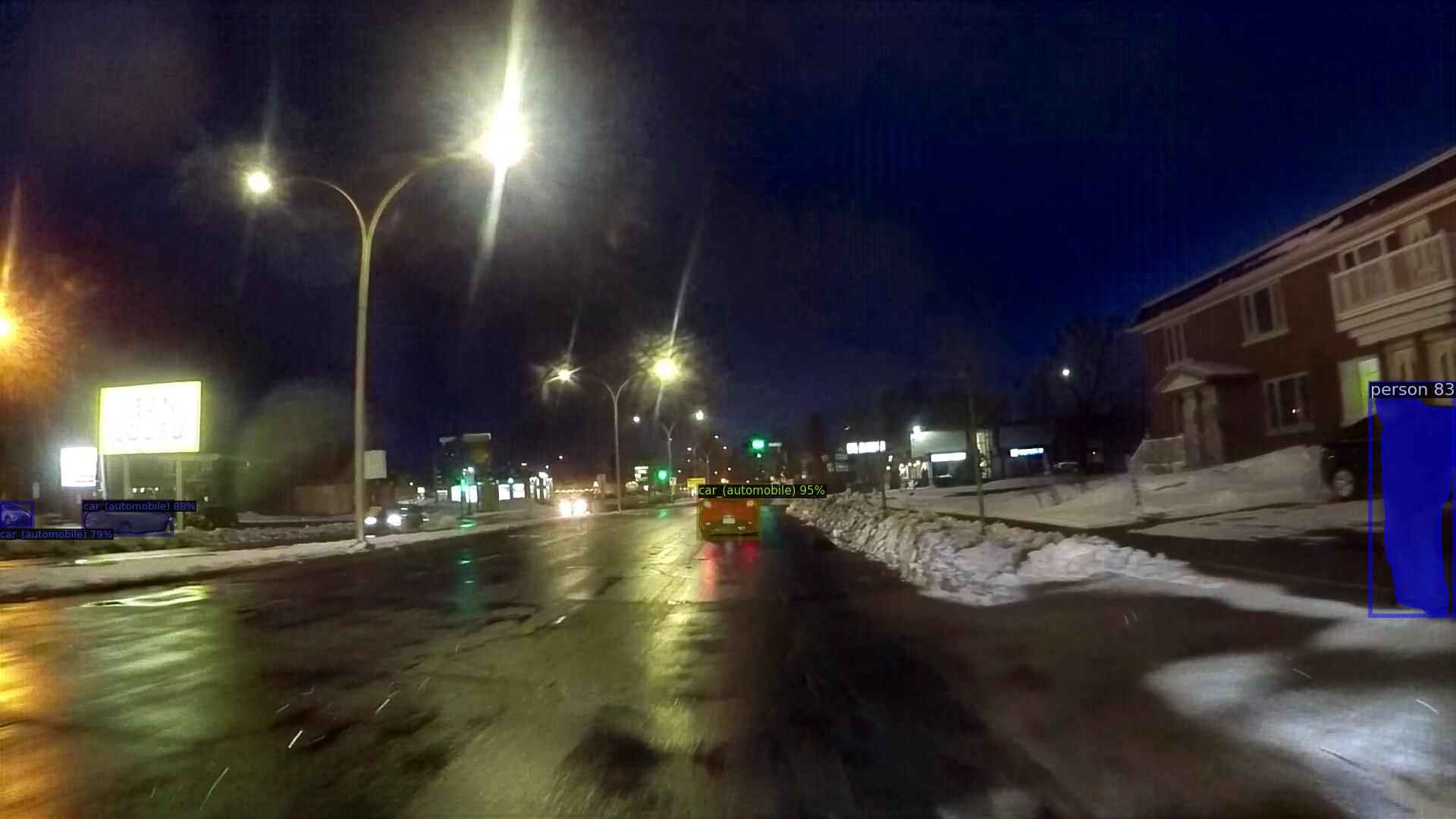}}
\end{subfigure}\\
\hspace{-0.04cm}
\raisebox{.9\height}{ \rotatebox[origin=]{90}{Bottom}}
\begin{subfigure}{1.3in}
{\includegraphics[width=1.3in,height=1.3in,keepaspectratio]{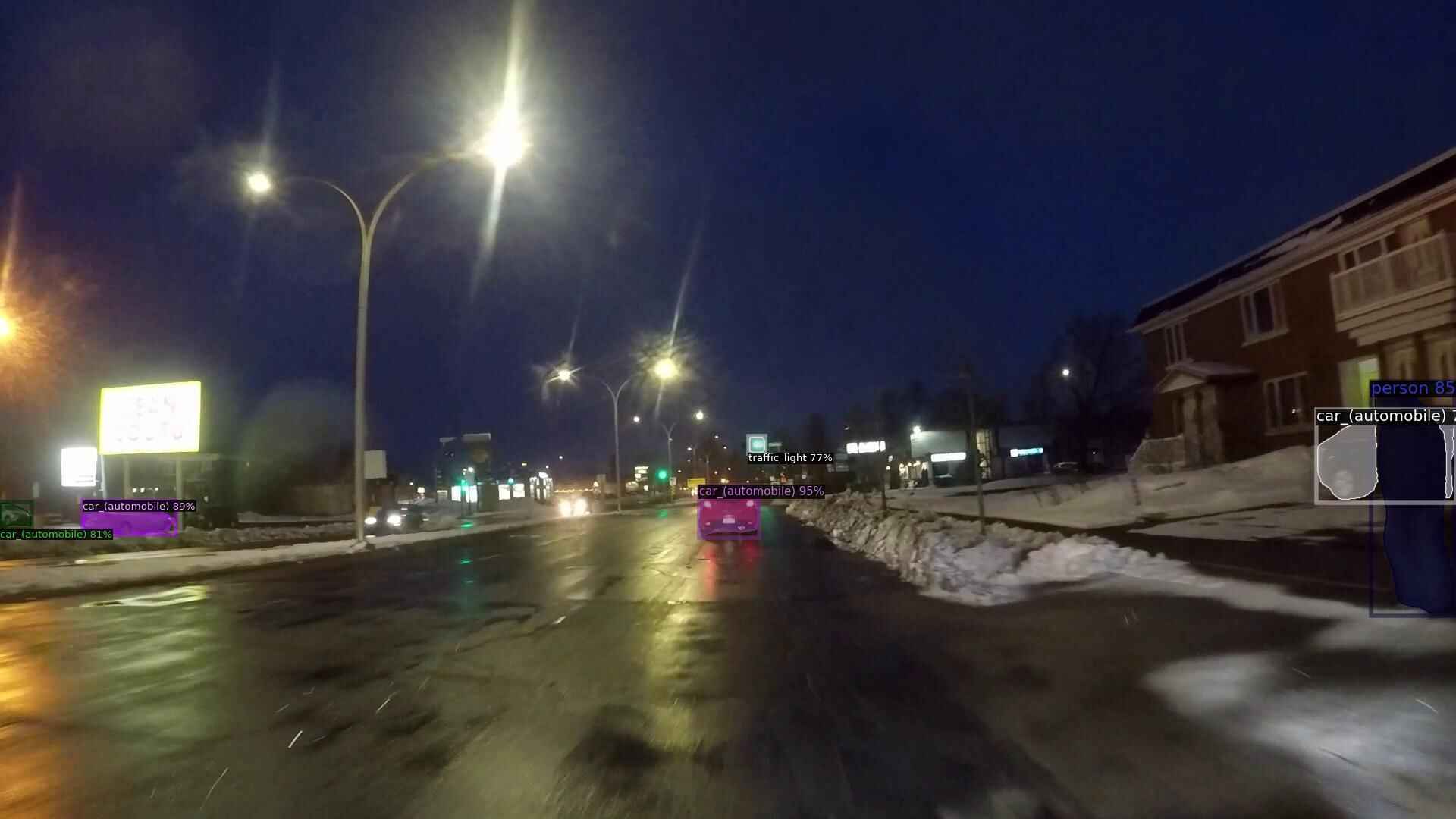}}
\end{subfigure}
\begin{subfigure}{1.3in}
{\includegraphics[width=1.3in,height=1.3in,keepaspectratio]{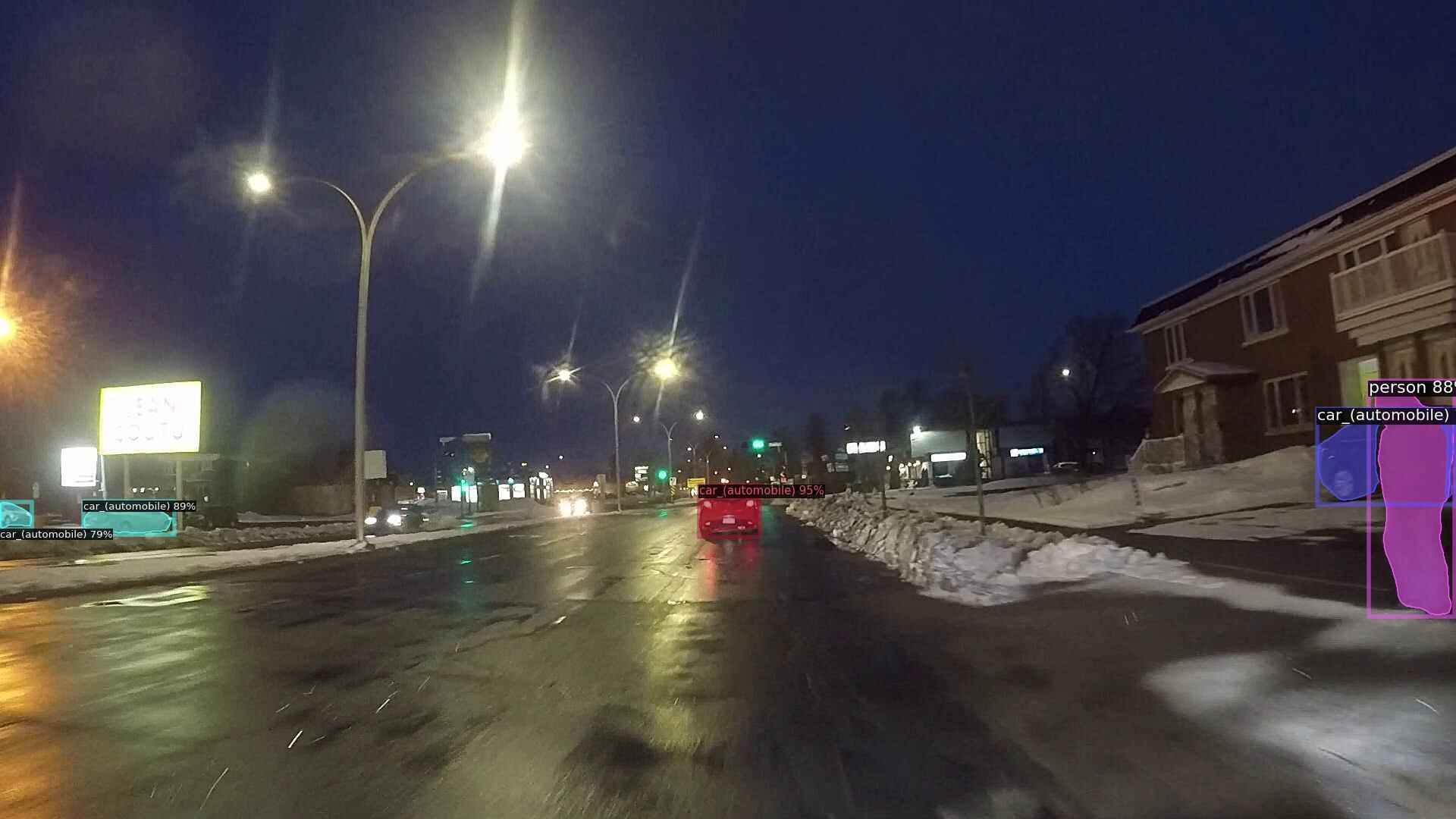}}
\end{subfigure}
\begin{subfigure}{1.3in}
{\includegraphics[width=1.3in,height=1.3in,keepaspectratio]{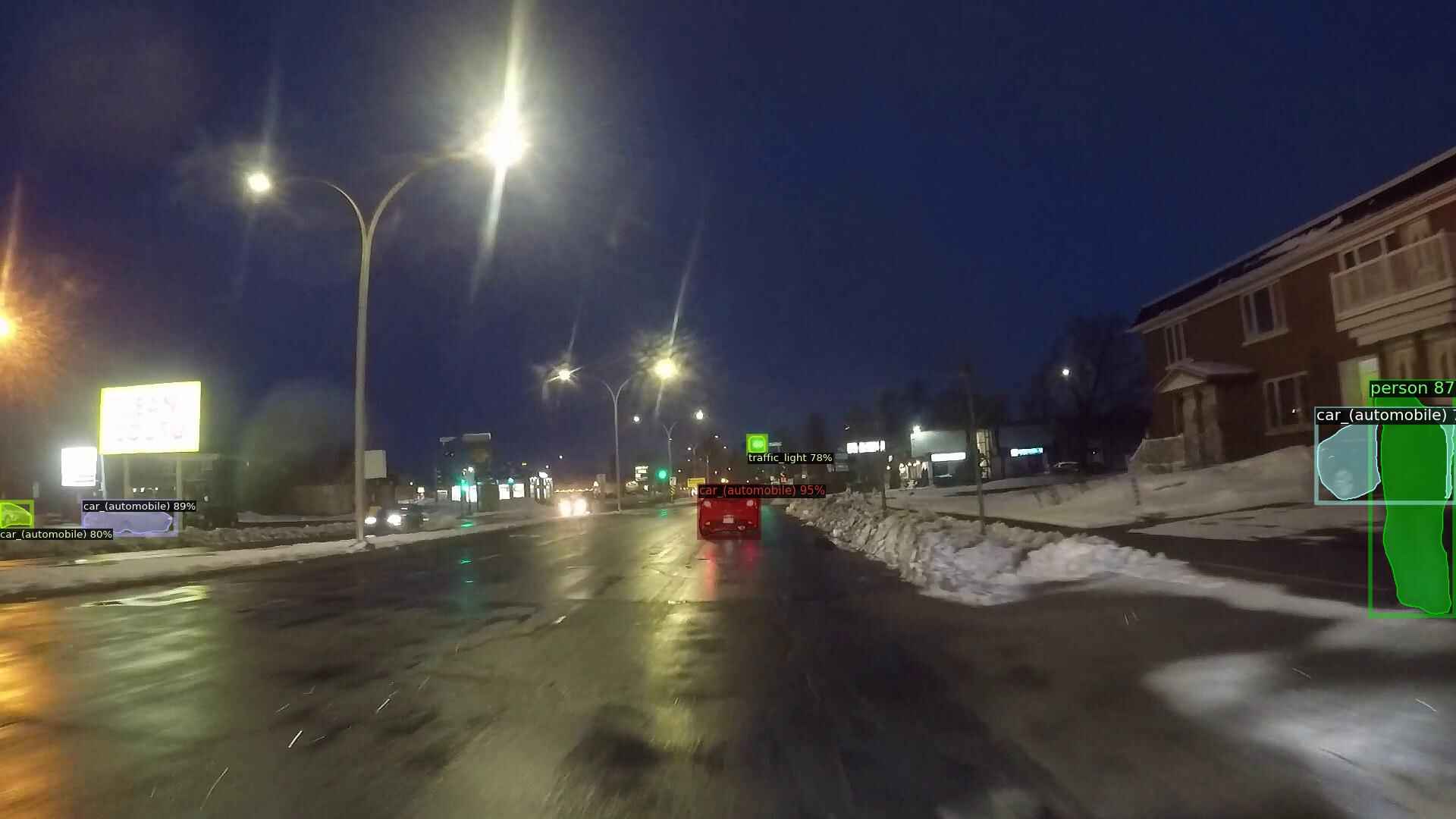}}
\end{subfigure}
\begin{subfigure}{1.3in}
{\includegraphics[width=1.3in,height=1.3in,keepaspectratio]{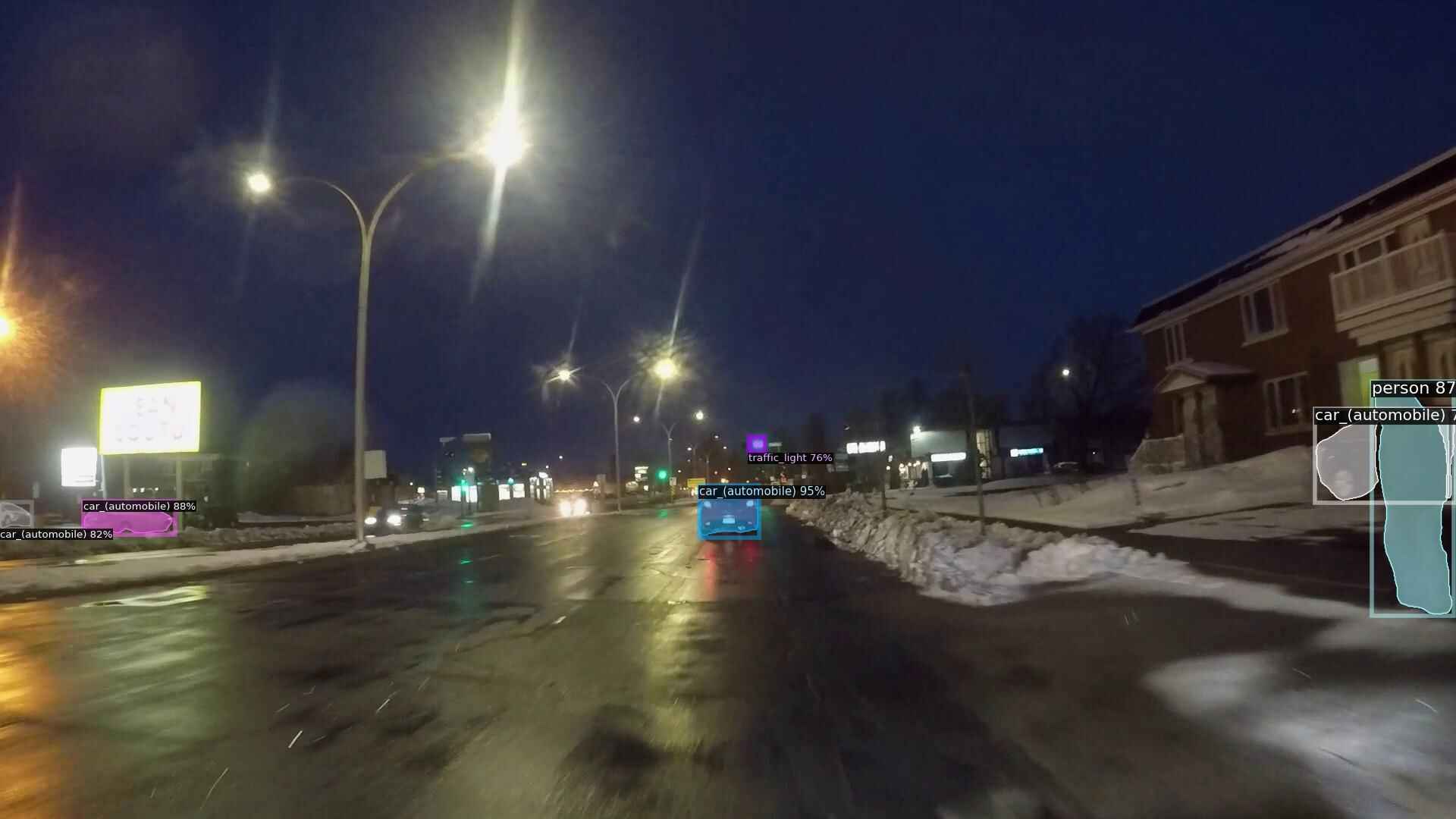}}
\end{subfigure}\\
\raisebox{.9\height}{ \rotatebox[origin=]{90}{TOP}}
\begin{subfigure}{1.3in}
{\includegraphics[trim={75 70 75 75},clip,width=1.3in,height=1.3in,keepaspectratio]{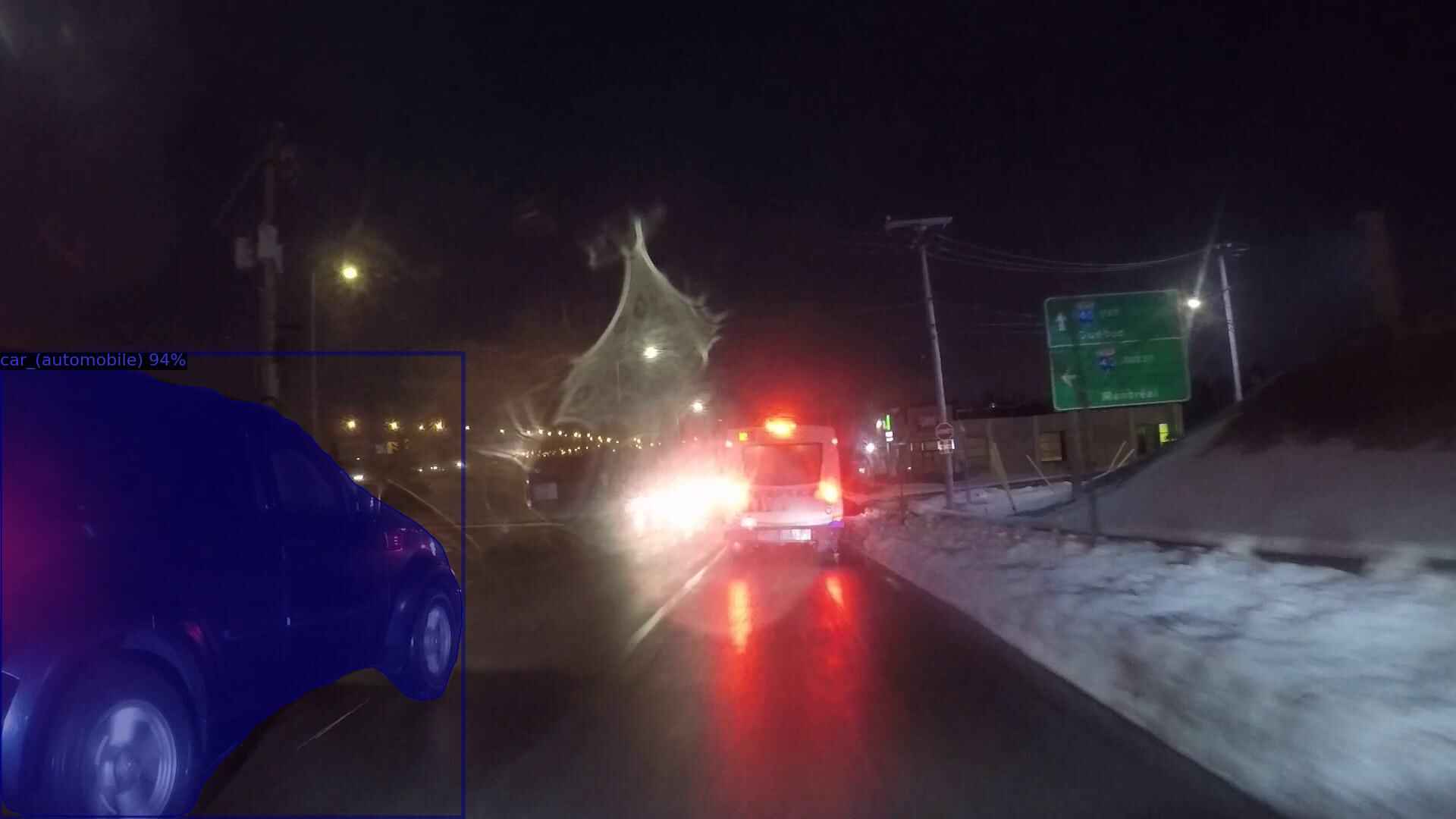}}
\end{subfigure}
\begin{subfigure}{1.3in}
{\includegraphics[trim={75 70 75 75},clip,width=1.3in,height=1.3in,keepaspectratio]{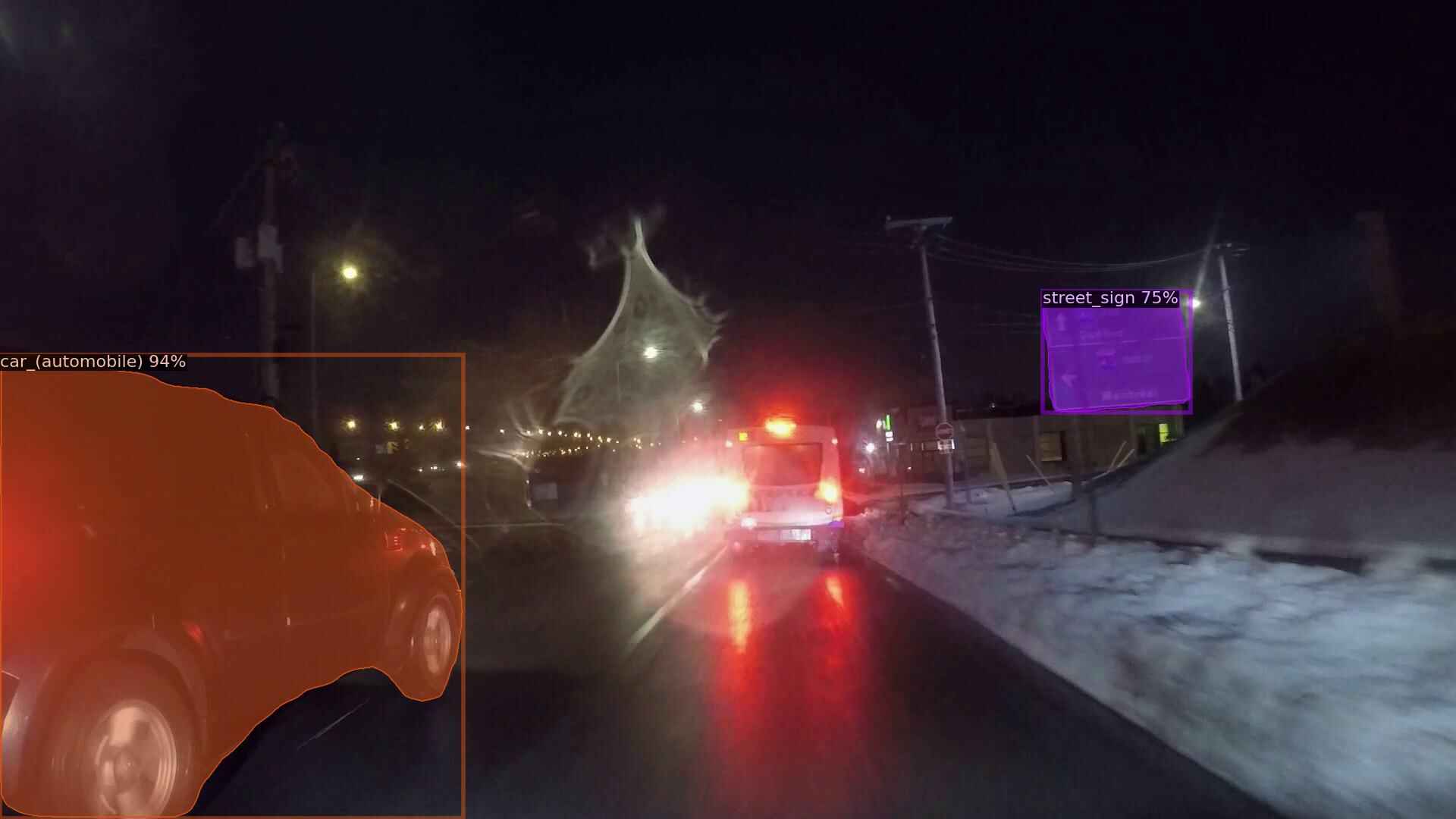}}
\end{subfigure}
\begin{subfigure}{1.3in}
{\includegraphics[trim={75 70 75 75 },clip,width=1.3in,height=1.3in,keepaspectratio]{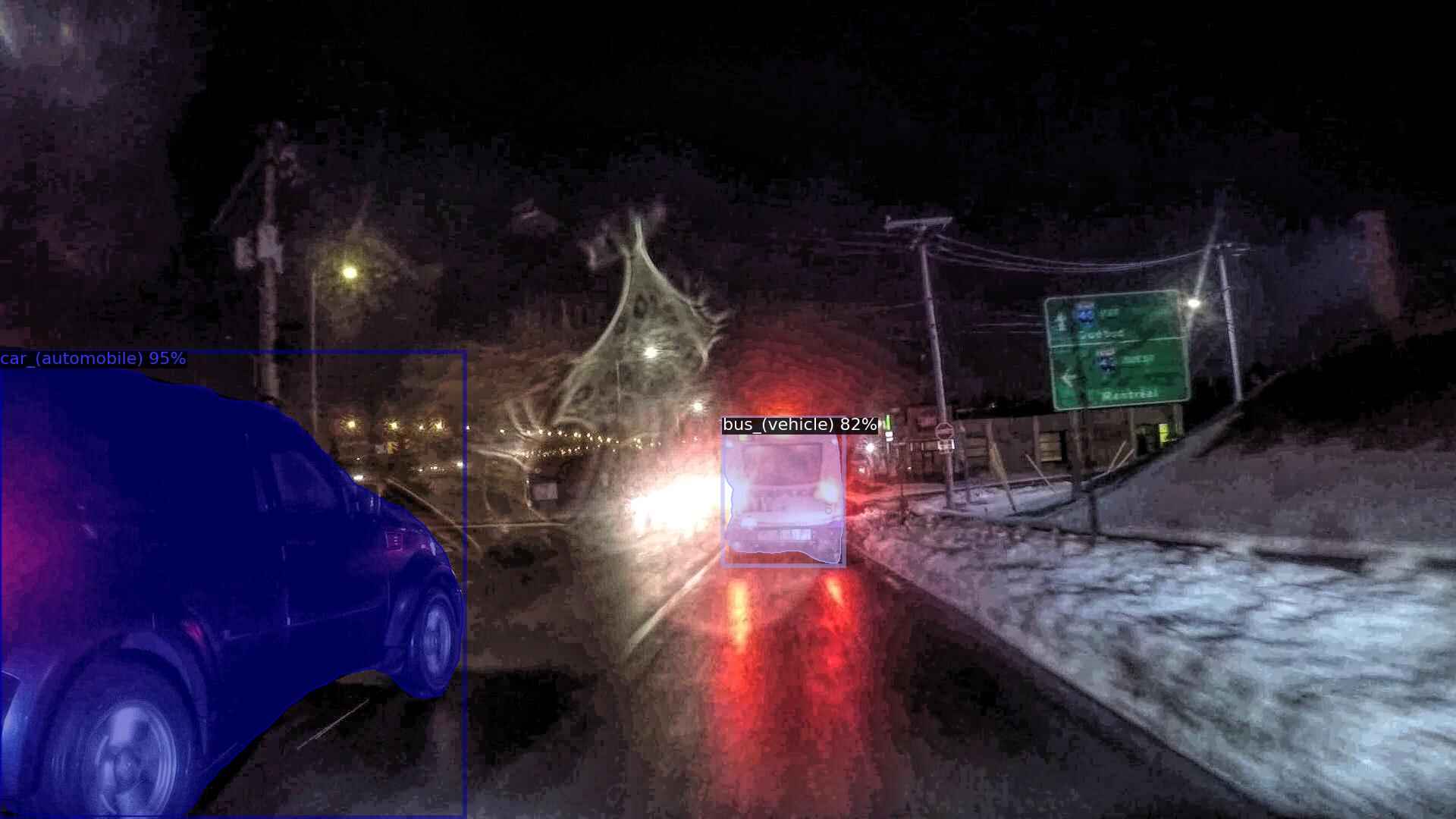}}
\end{subfigure}
\begin{subfigure}{1.3in}
{\includegraphics[trim={75 70 75 75 },clip,width=1.3in,height=1.3in,keepaspectratio]{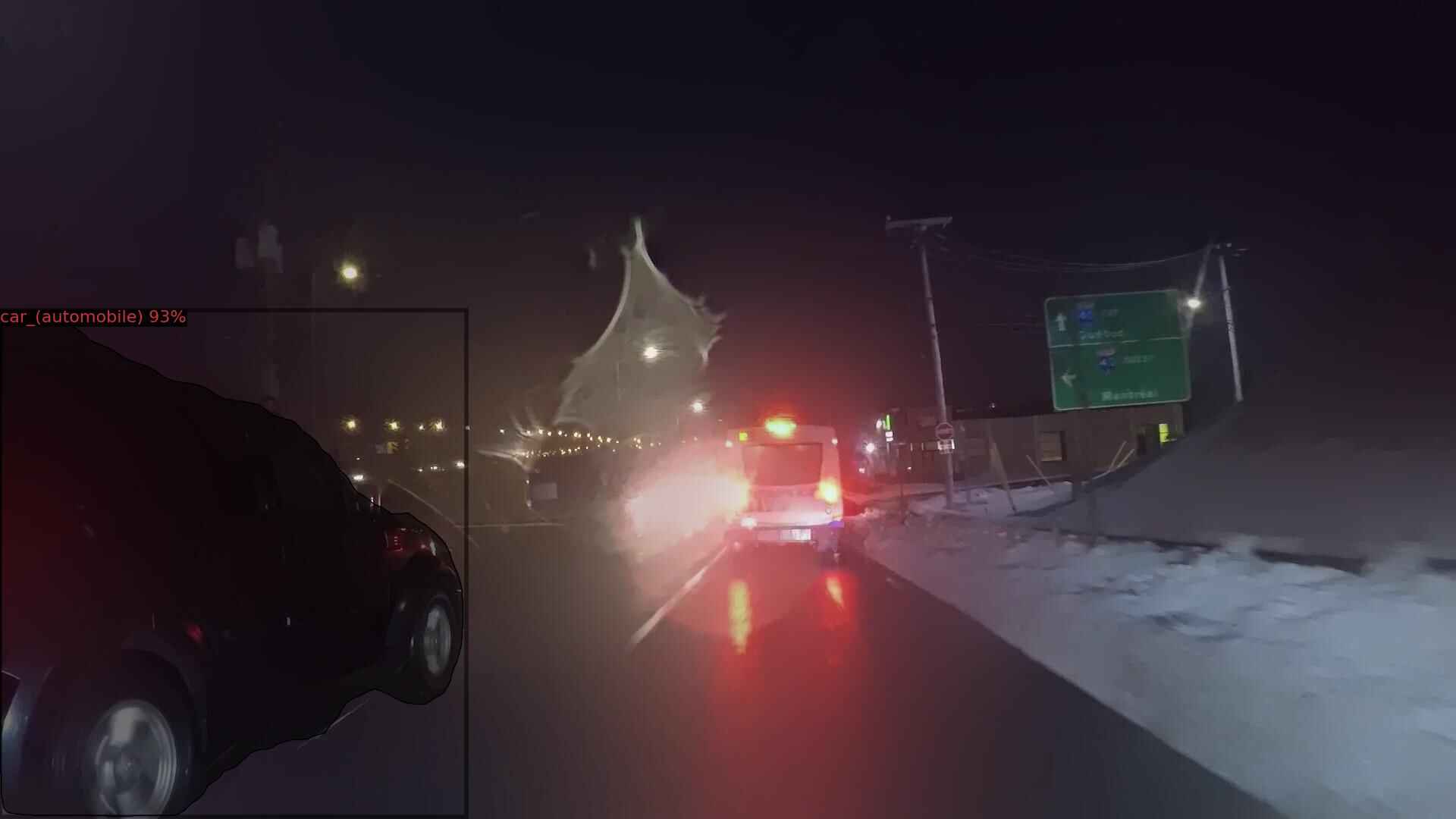}}
\end{subfigure}
\begin{subfigure}{1.3in}
{\includegraphics[trim={75 70 75 75},clip,width=1.3in,height=1.3in,keepaspectratio]{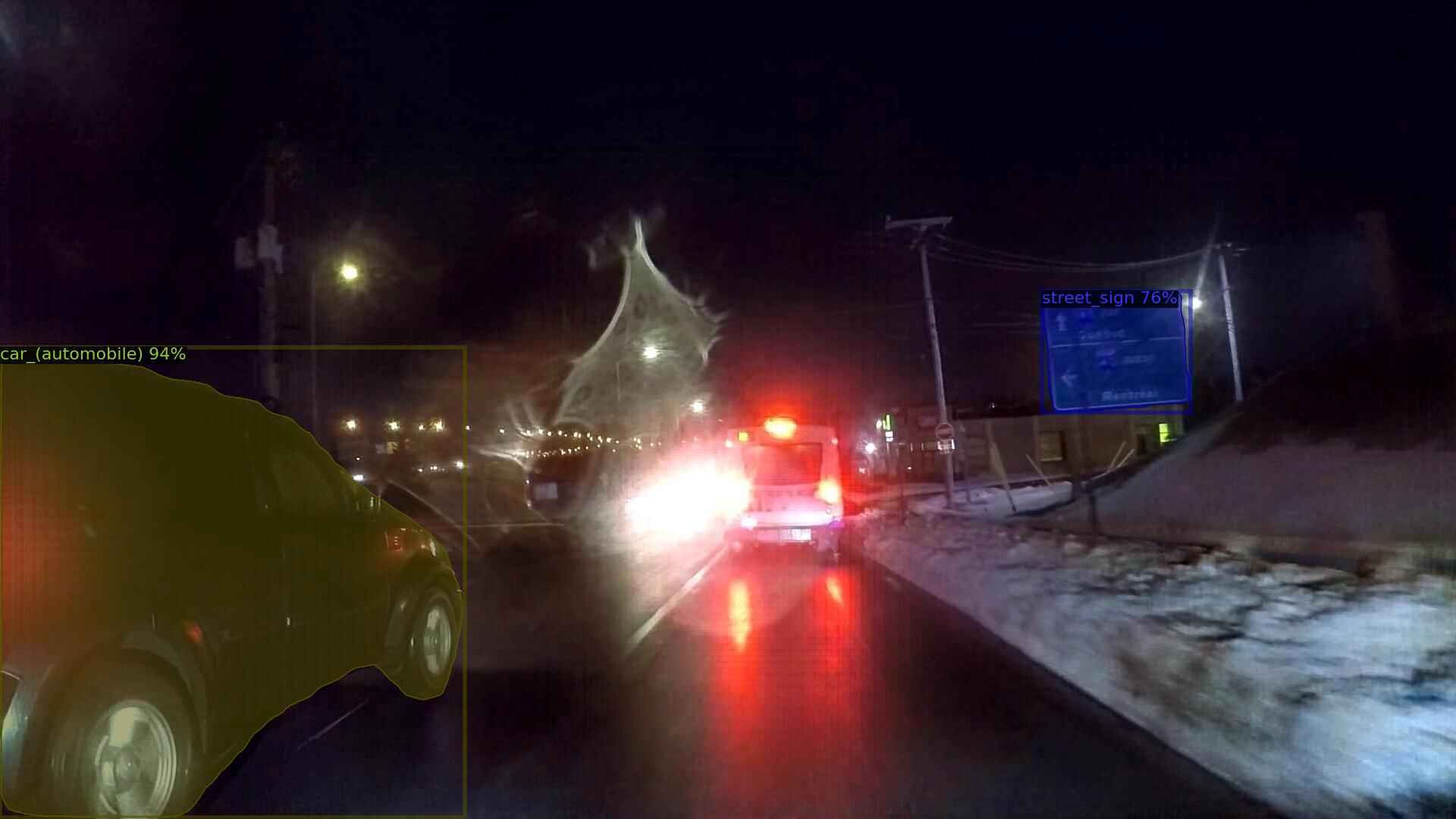}}
\end{subfigure}\\
\hspace{-0.08cm}
\raisebox{.9\height}{ \rotatebox[origin=]{90}{Bottom}}
\begin{subfigure}{1.3in}
{\includegraphics[trim={75 70 75 75},clip,width=1.3in,height=1.3in,keepaspectratio]{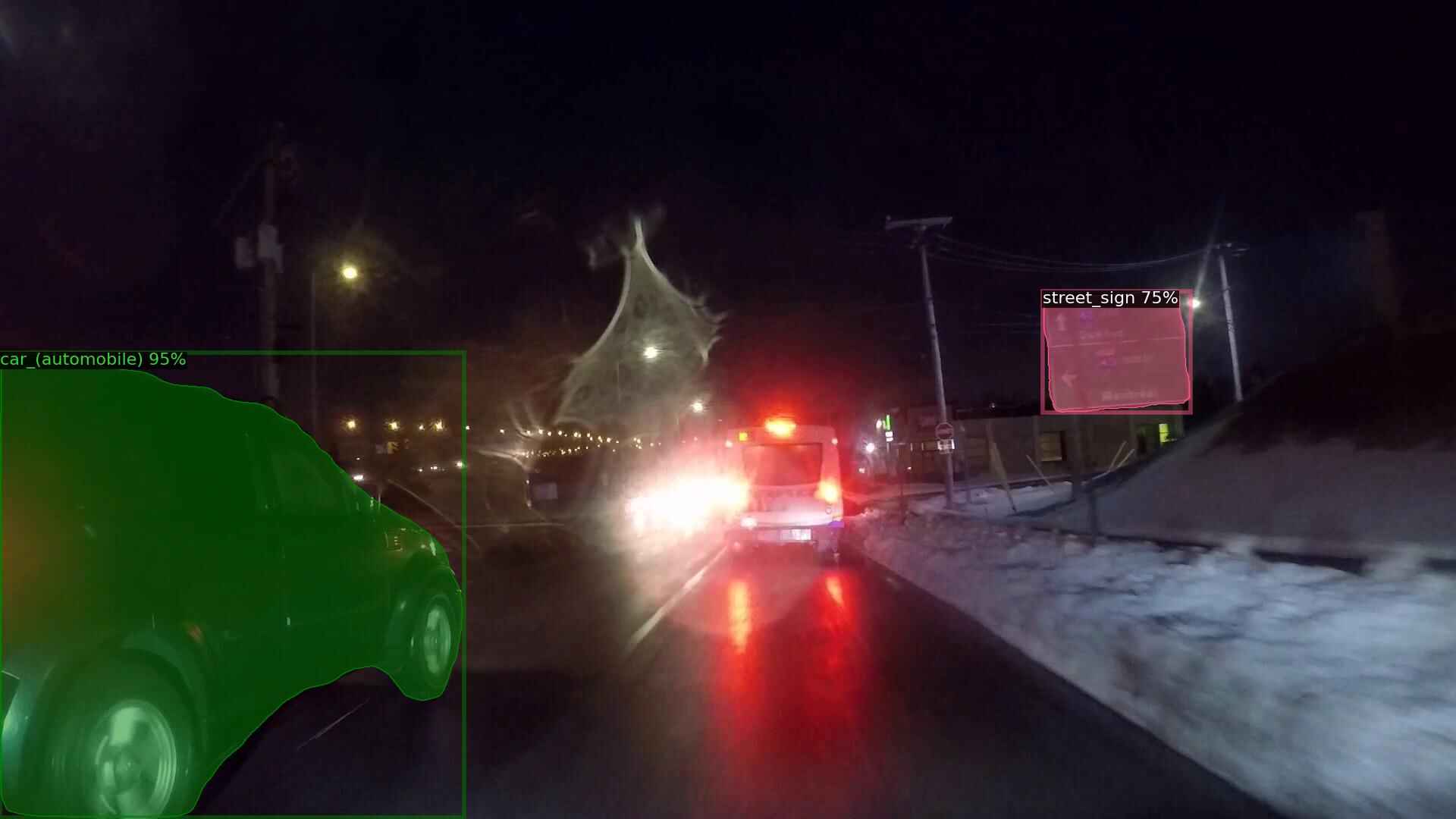}}
\end{subfigure}
\begin{subfigure}{1.3in}
{\includegraphics[trim={75 70 75 75},clip,width=1.3in,height=1.3in,keepaspectratio]{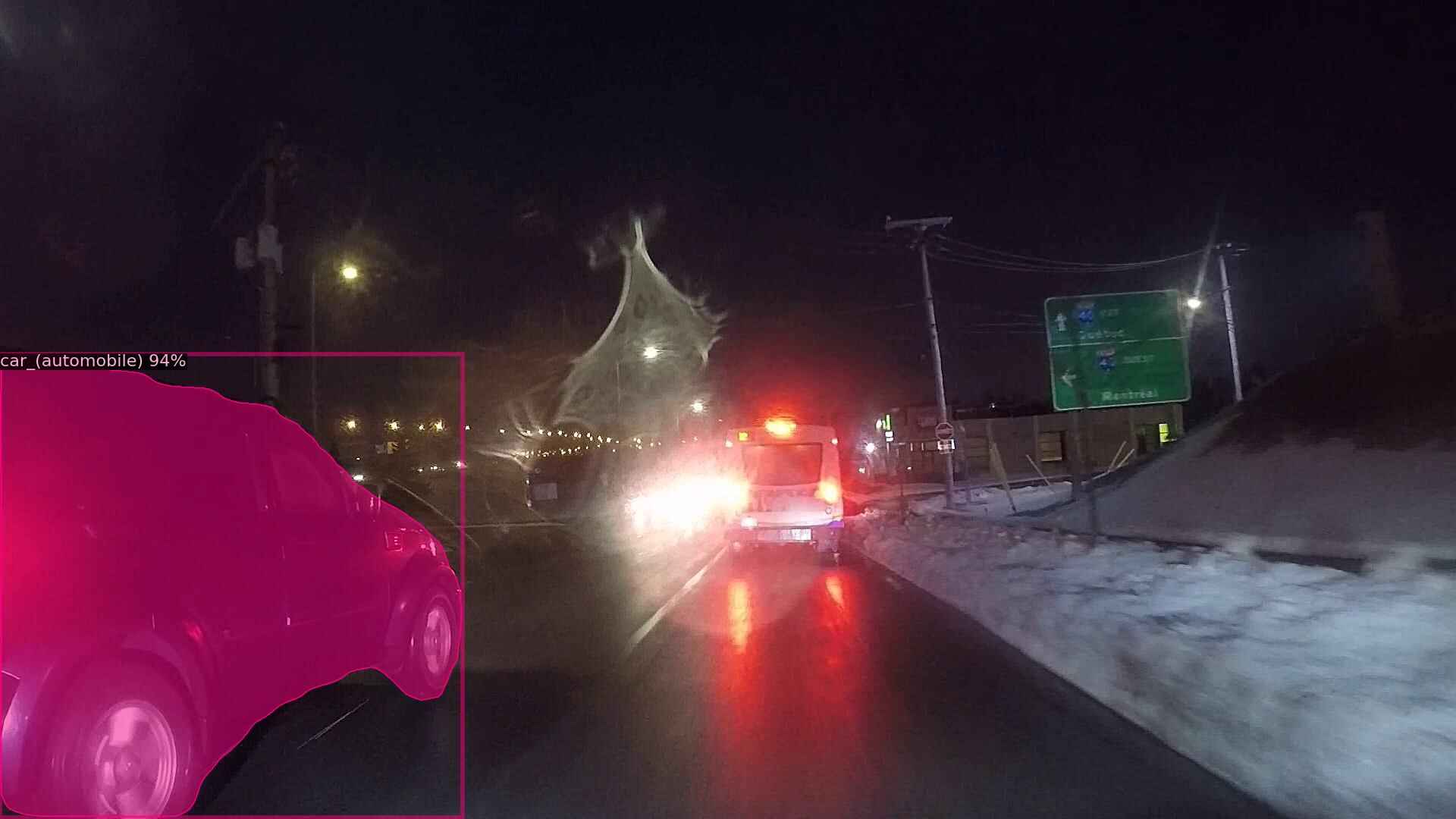}}
\end{subfigure}
\begin{subfigure}{1.3in}
{\includegraphics[trim={75 70 75 75},clip,width=1.3in,height=1.3in,keepaspectratio]{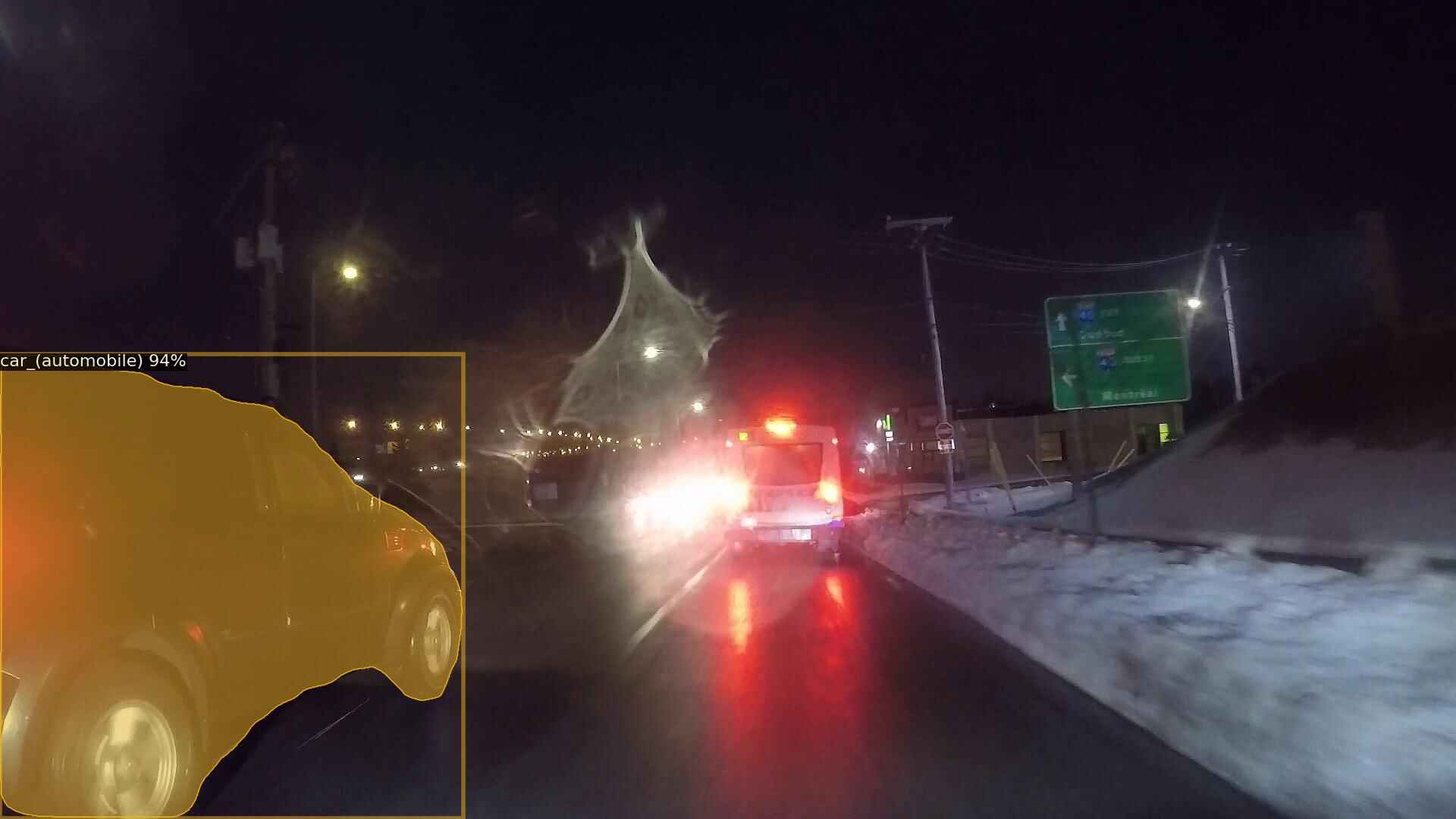}}
\end{subfigure}
\begin{subfigure}{1.3in}
{\includegraphics[trim={75 70 75 75},clip,width=1.3in,height=1.3in,keepaspectratio]{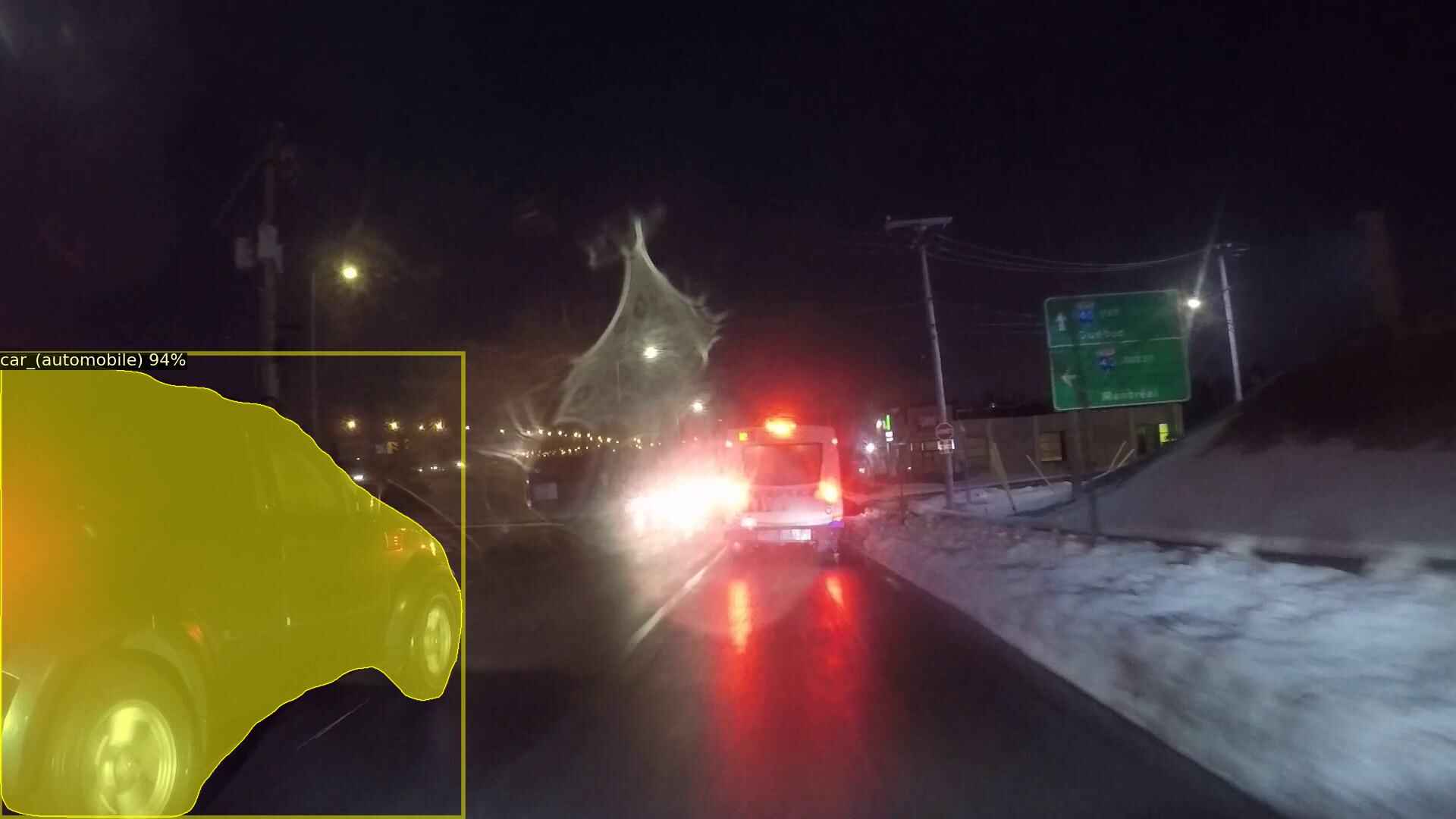}}
\end{subfigure}\\
\raisebox{.7\height}{ \rotatebox[origin=]{90}{Top}}
\begin{subfigure}{1.3in}
{\includegraphics[width=1.3in,height=1.3in,keepaspectratio]{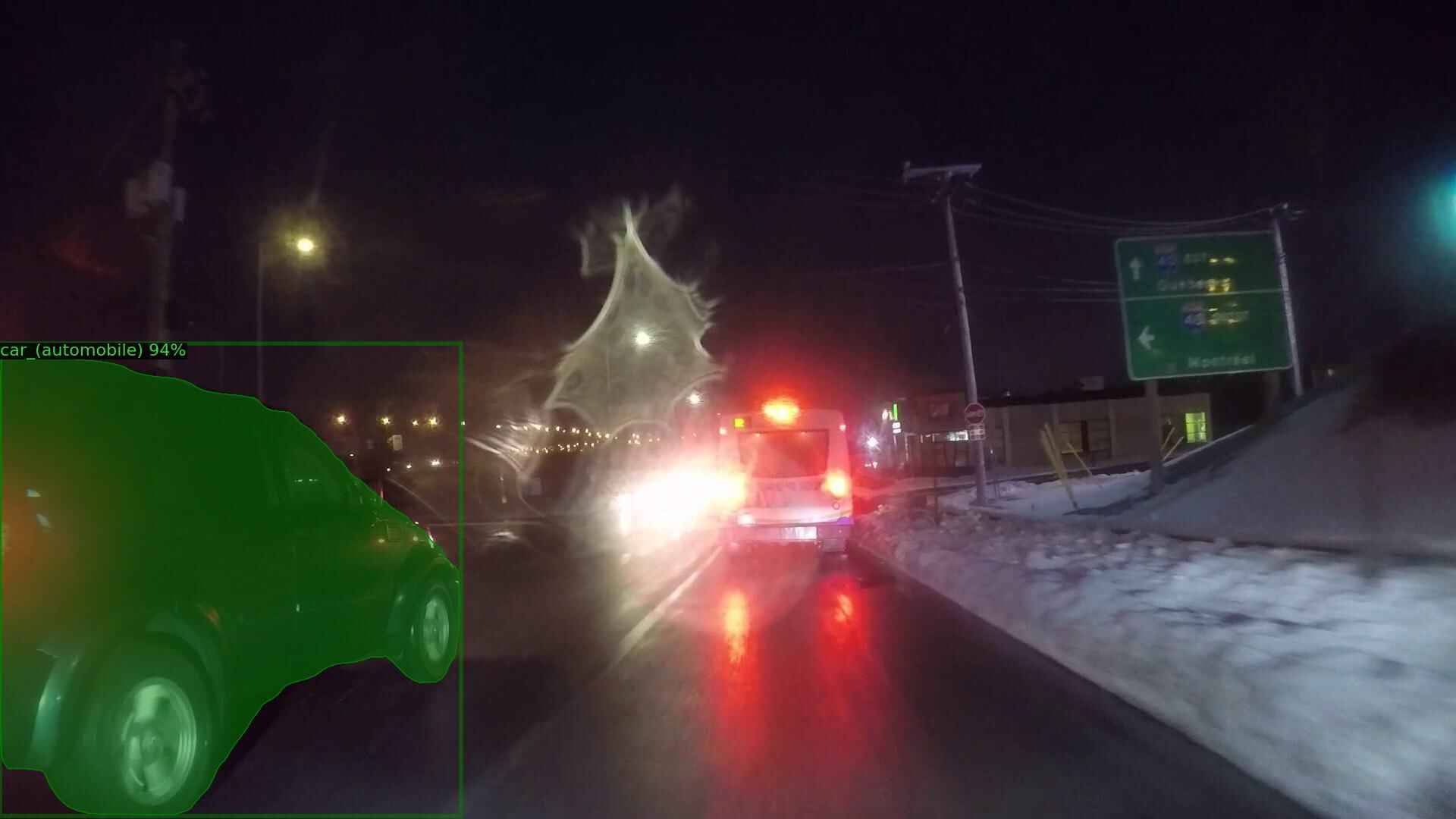}}
\end{subfigure}
\begin{subfigure}{1.3in}
{\includegraphics[width=1.3in,height=1.3in,keepaspectratio]{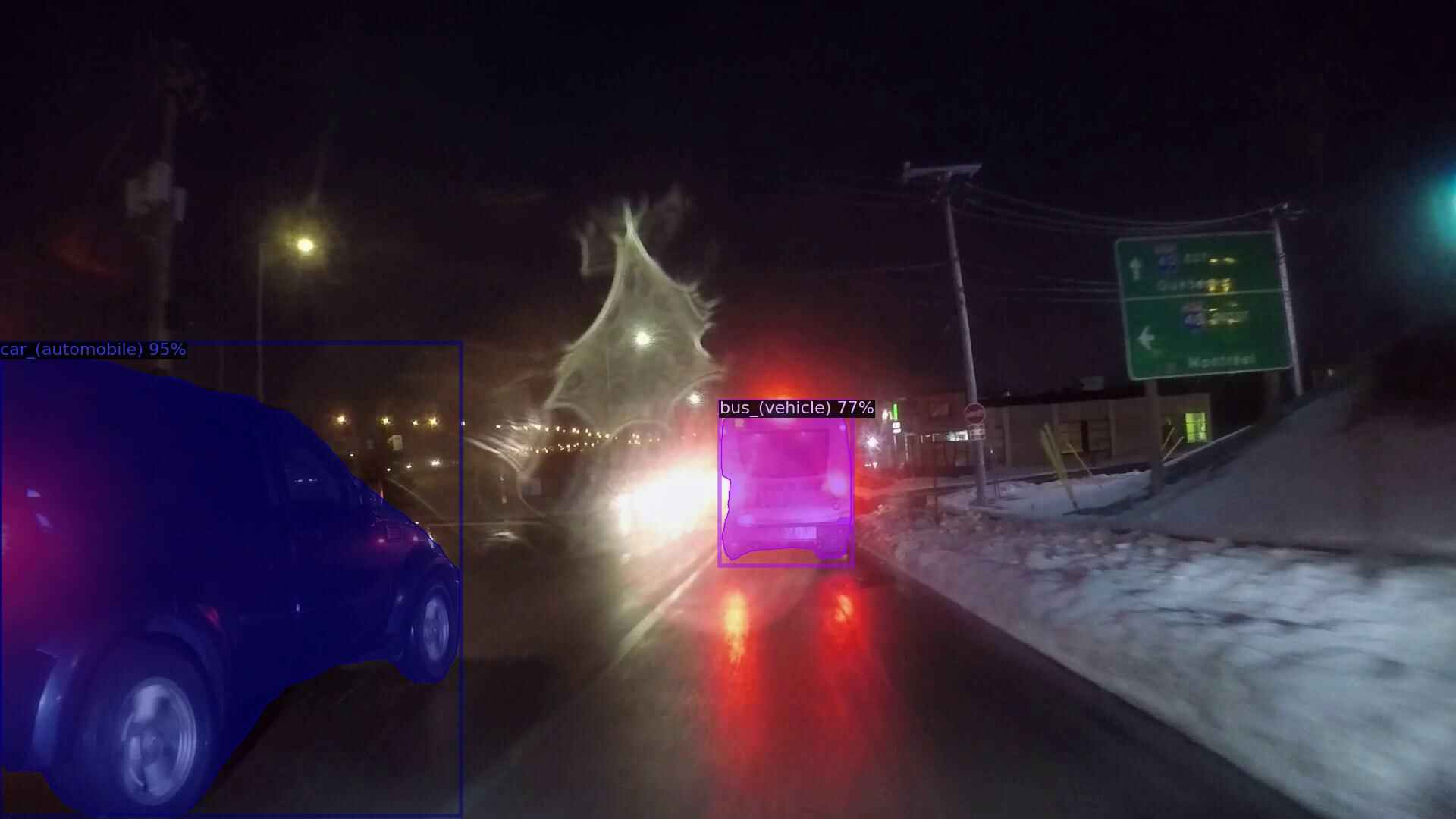}}
\end{subfigure}
\begin{subfigure}{1.3in}
{\includegraphics[width=1.3in,height=1.3in,keepaspectratio]{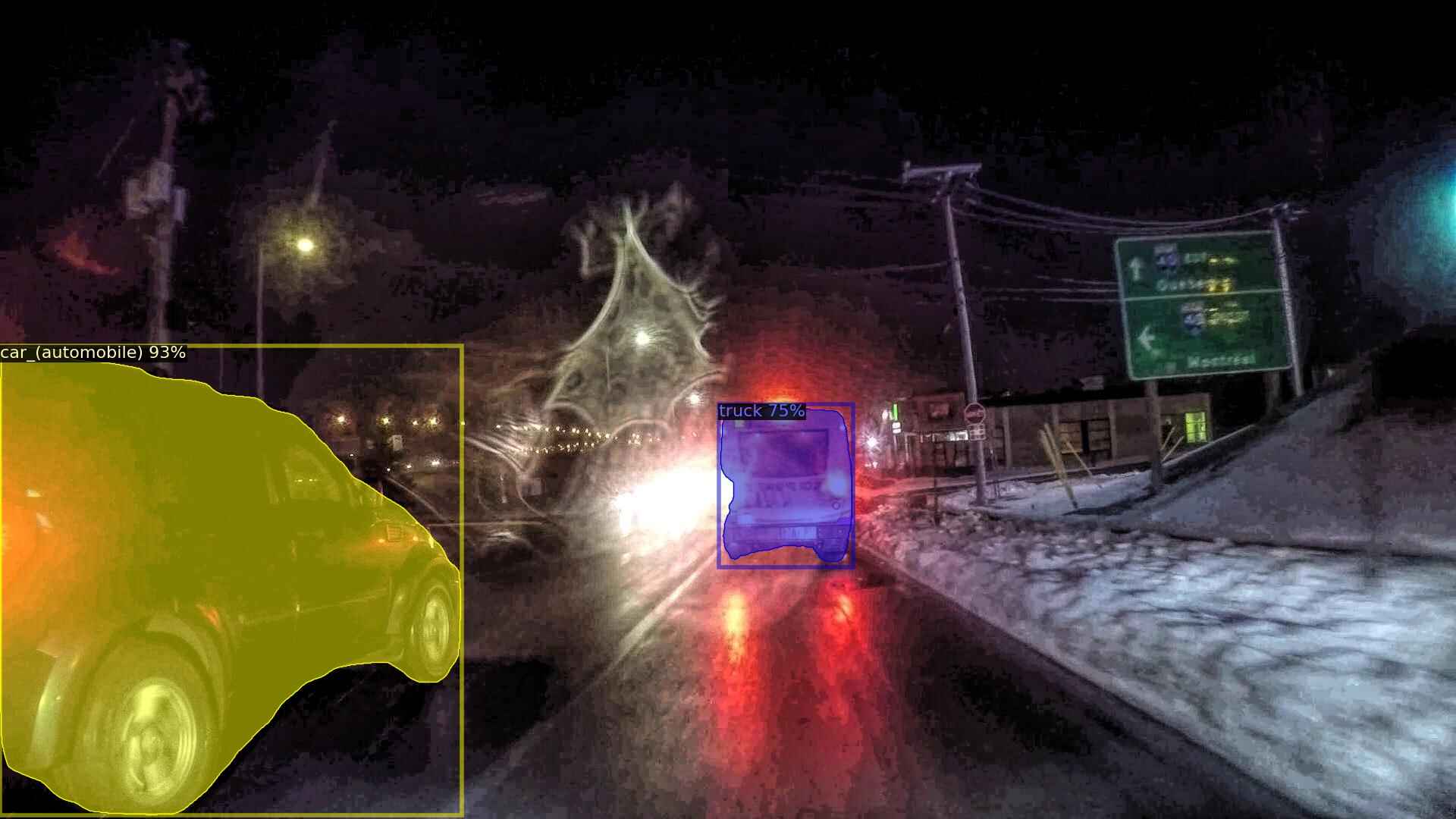}}
\end{subfigure}
\begin{subfigure}{1.3in}
{\includegraphics[width=1.3in,height=1.3in,keepaspectratio]{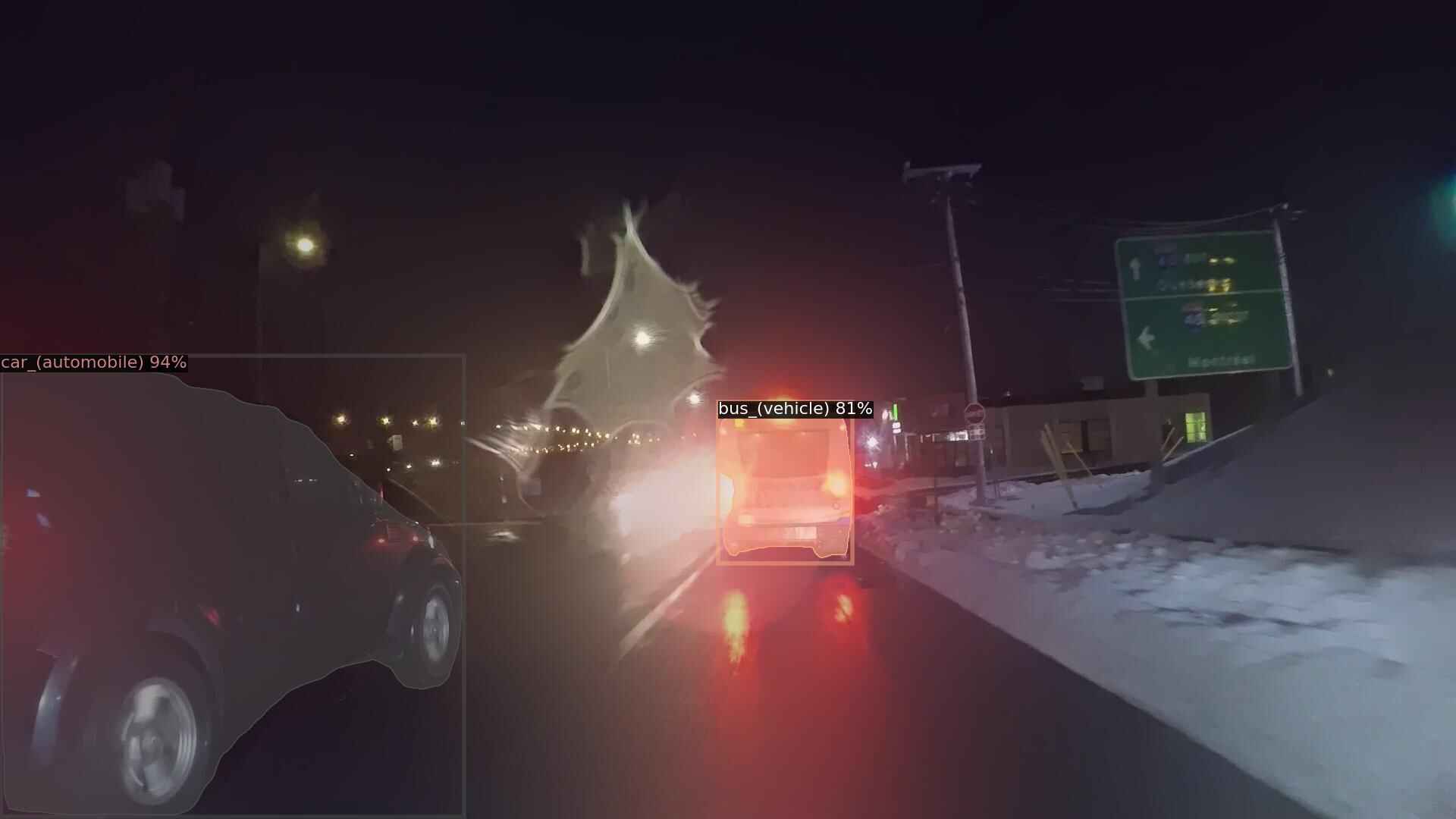}}
\end{subfigure}
\begin{subfigure}{1.3in}
{\includegraphics[width=1.3in,height=1.3in,keepaspectratio]{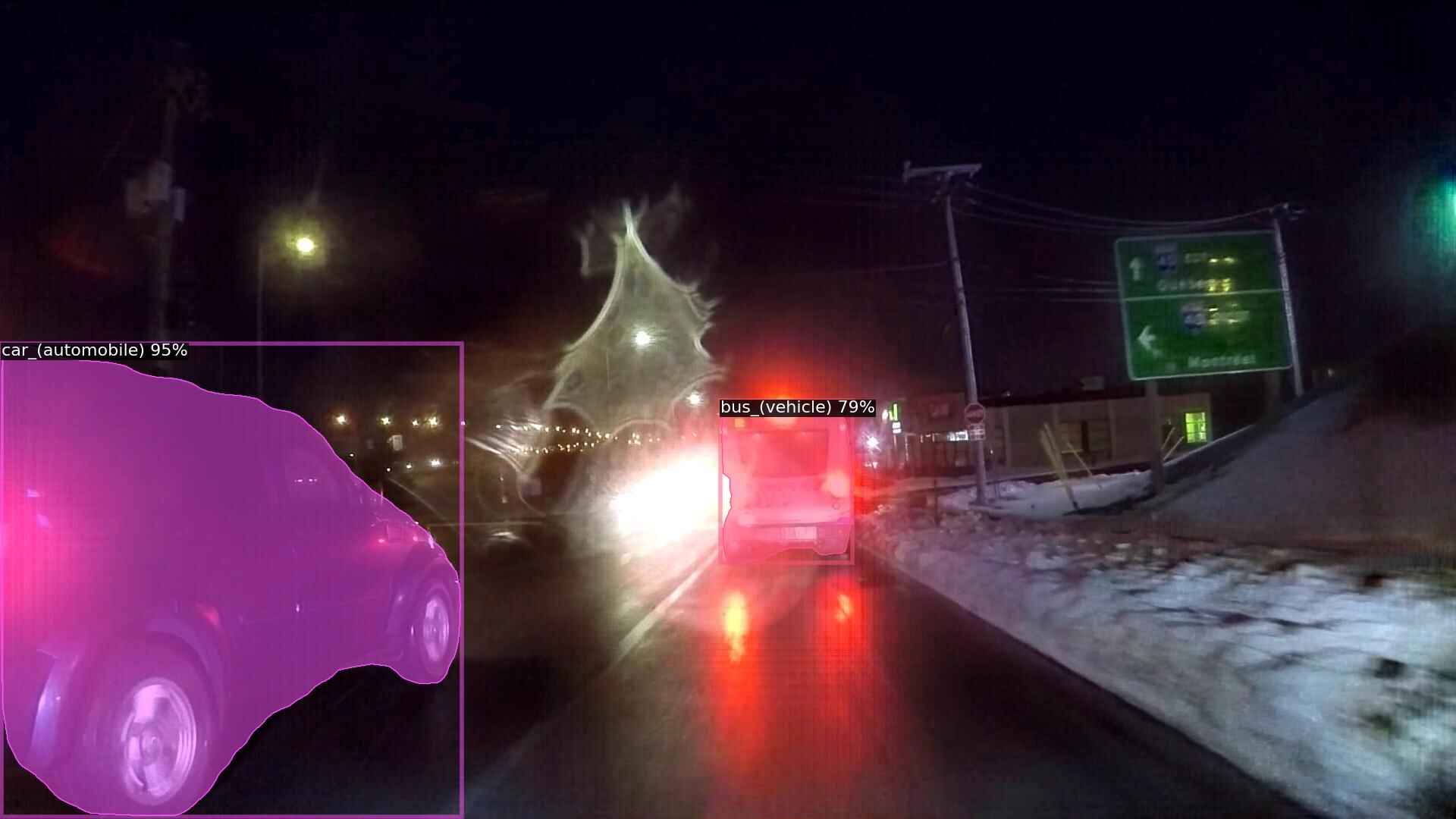}}
\end{subfigure}\\
\hspace{-0.04cm}
\raisebox{.9\height}{ \rotatebox[origin=]{90}{Bottom}}
\begin{subfigure}{1.3in}
{\includegraphics[width=1.3in,height=1.3in,keepaspectratio]{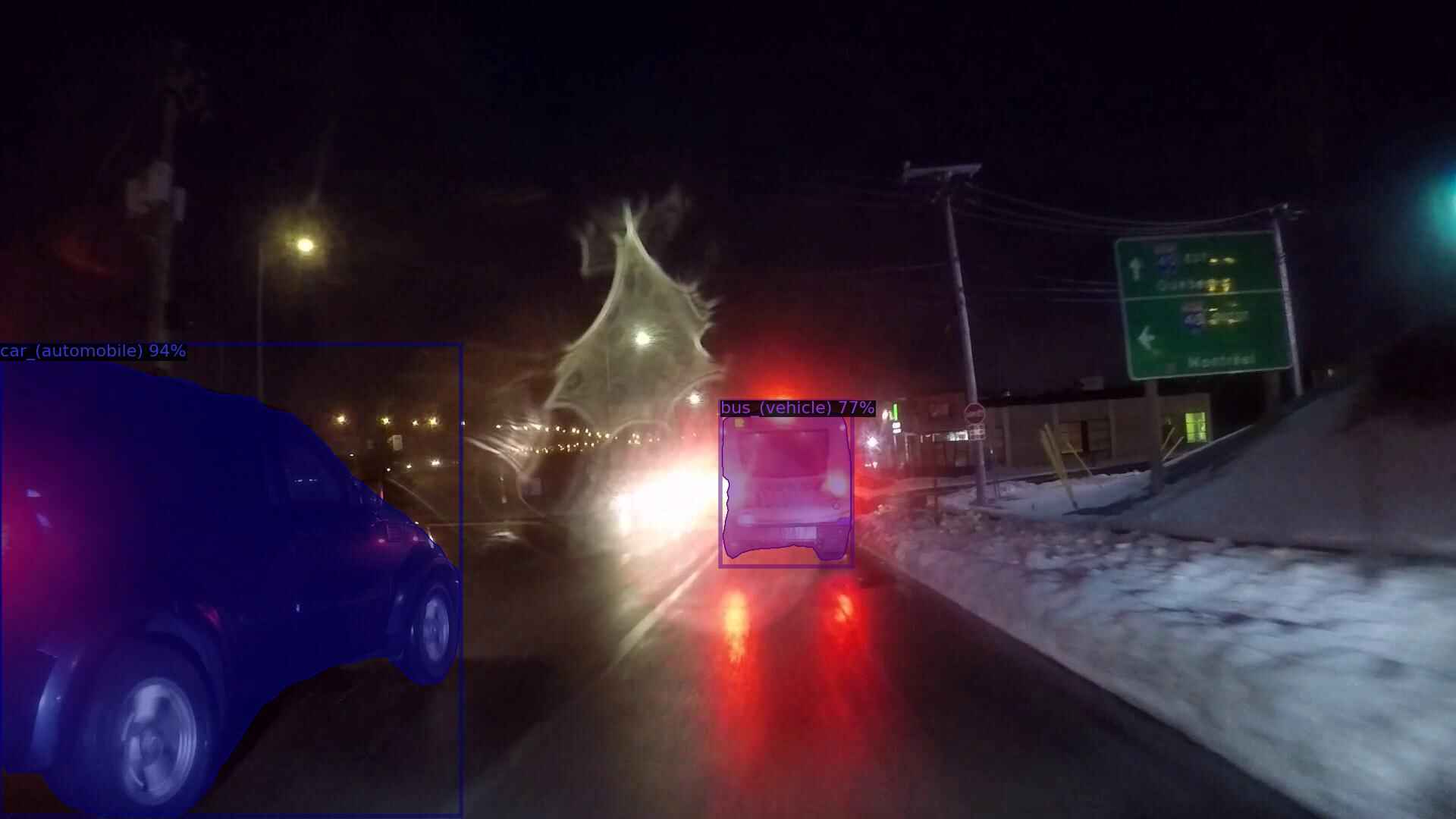}}
\end{subfigure}
\begin{subfigure}{1.3in}
{\includegraphics[width=1.3in,height=1.3in,keepaspectratio]{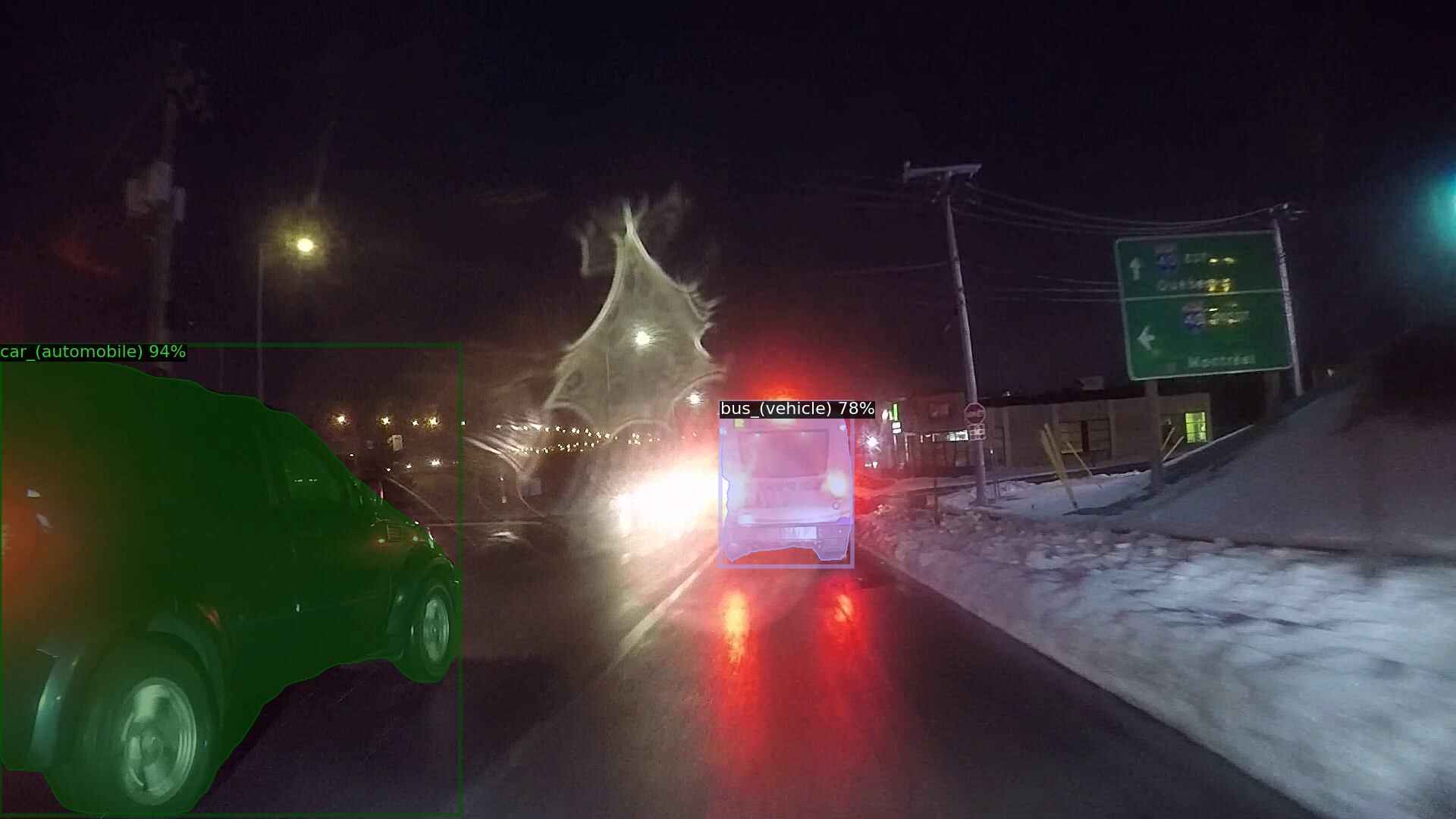}}
\end{subfigure}
\begin{subfigure}{1.3in}
{\includegraphics[width=1.3in,height=1.3in,keepaspectratio]{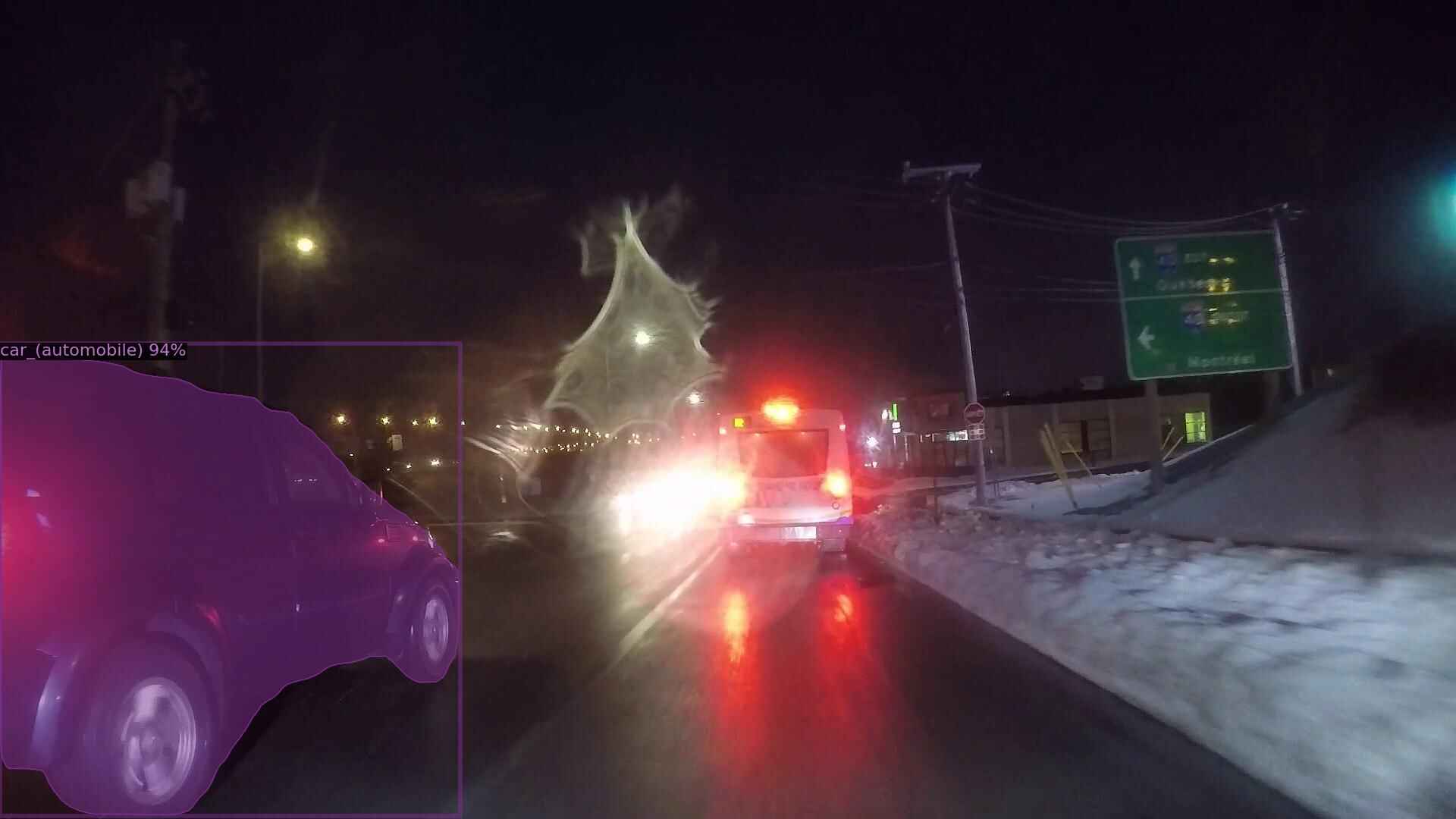}}
\end{subfigure}
\begin{subfigure}{1.3in}
{\includegraphics[width=1.3in,height=1.3in,keepaspectratio]{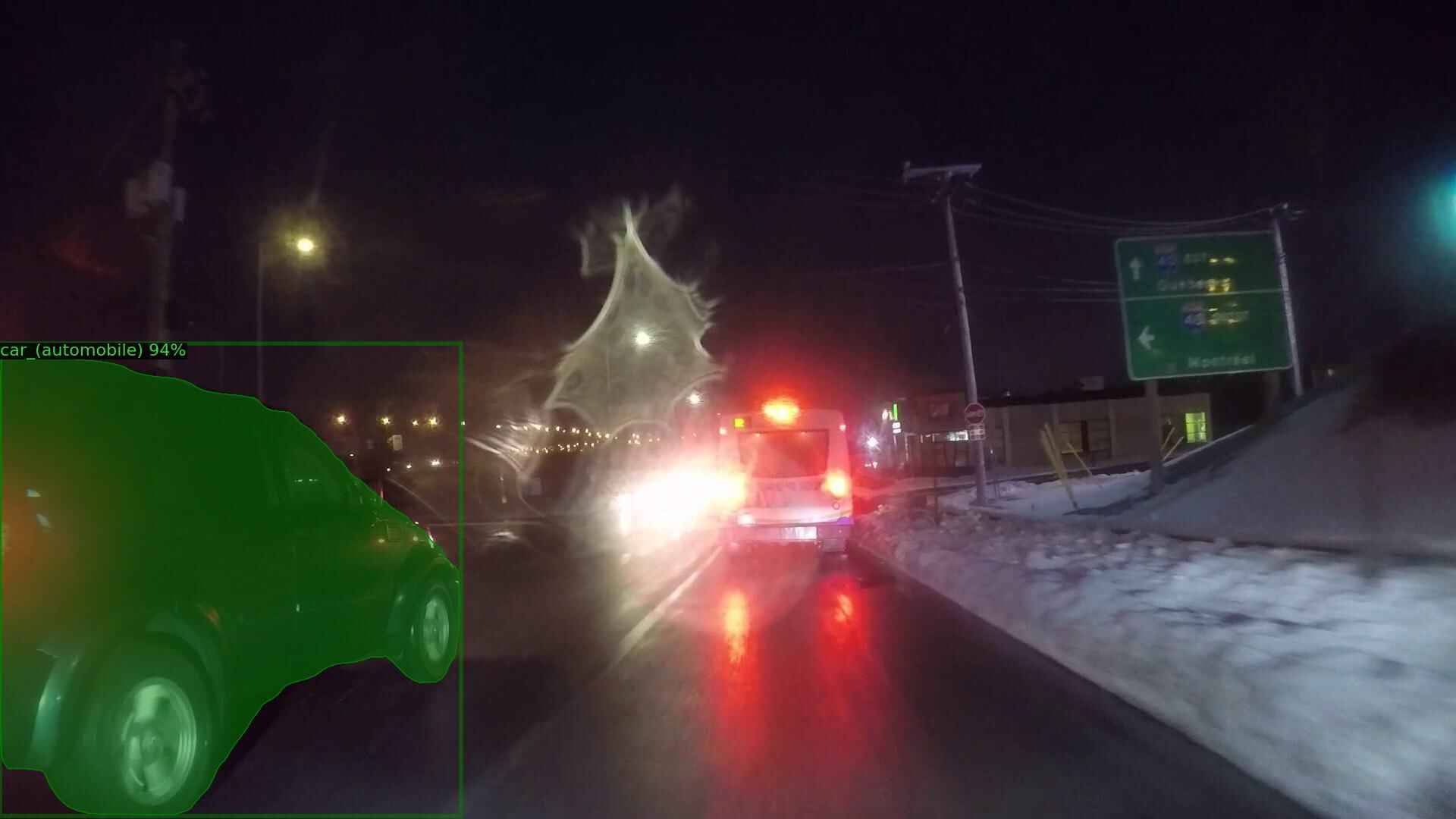}}
\end{subfigure}\\
\raisebox{.7\height}{ \rotatebox[origin=]{90}{Top}}
\begin{subfigure}{1.3in}
{\includegraphics[width=1.3in,height=1.3in,keepaspectratio]{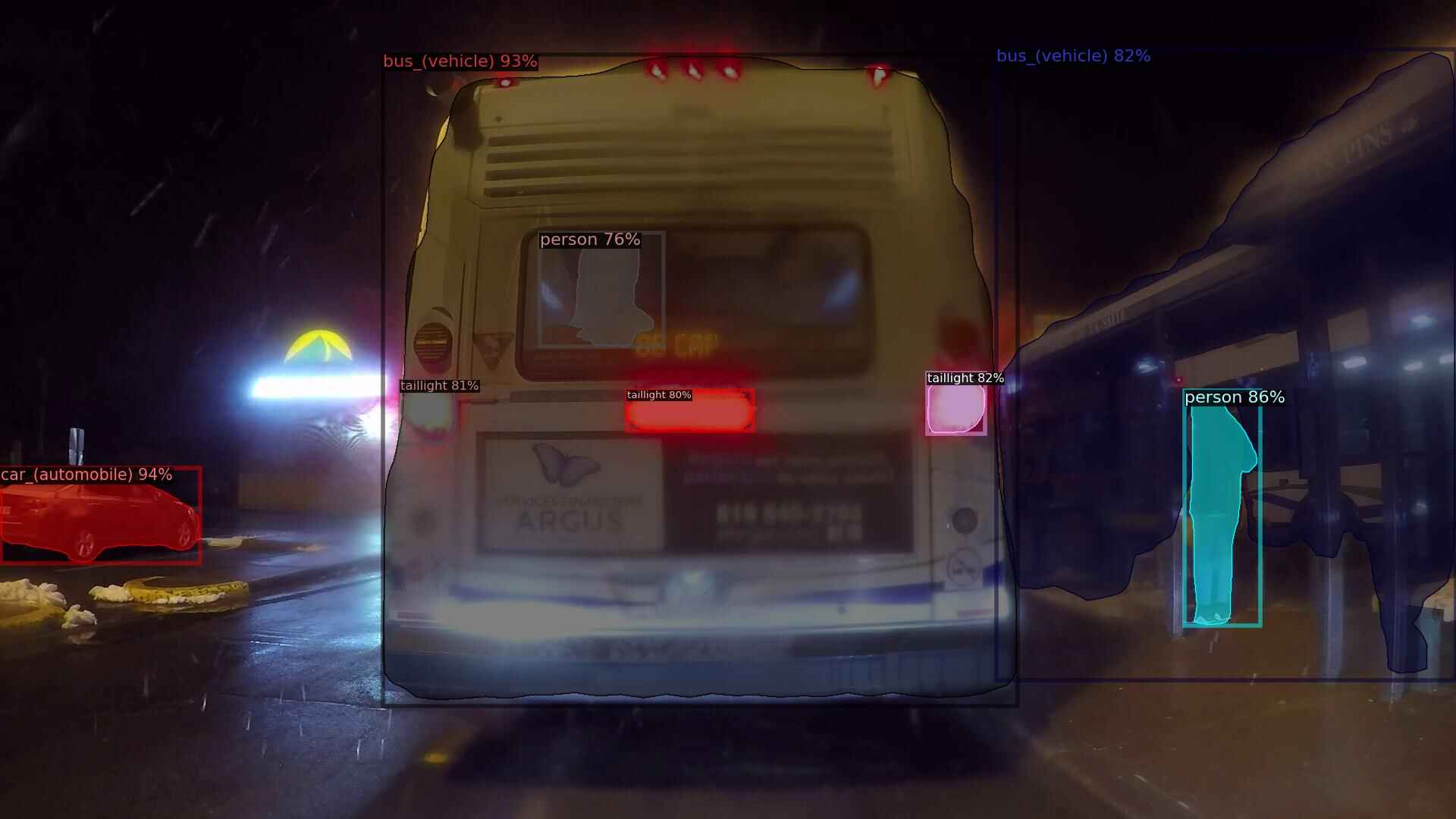}}
\end{subfigure}
\begin{subfigure}{1.3in}
{\includegraphics[width=1.3in,height=1.3in,keepaspectratio]{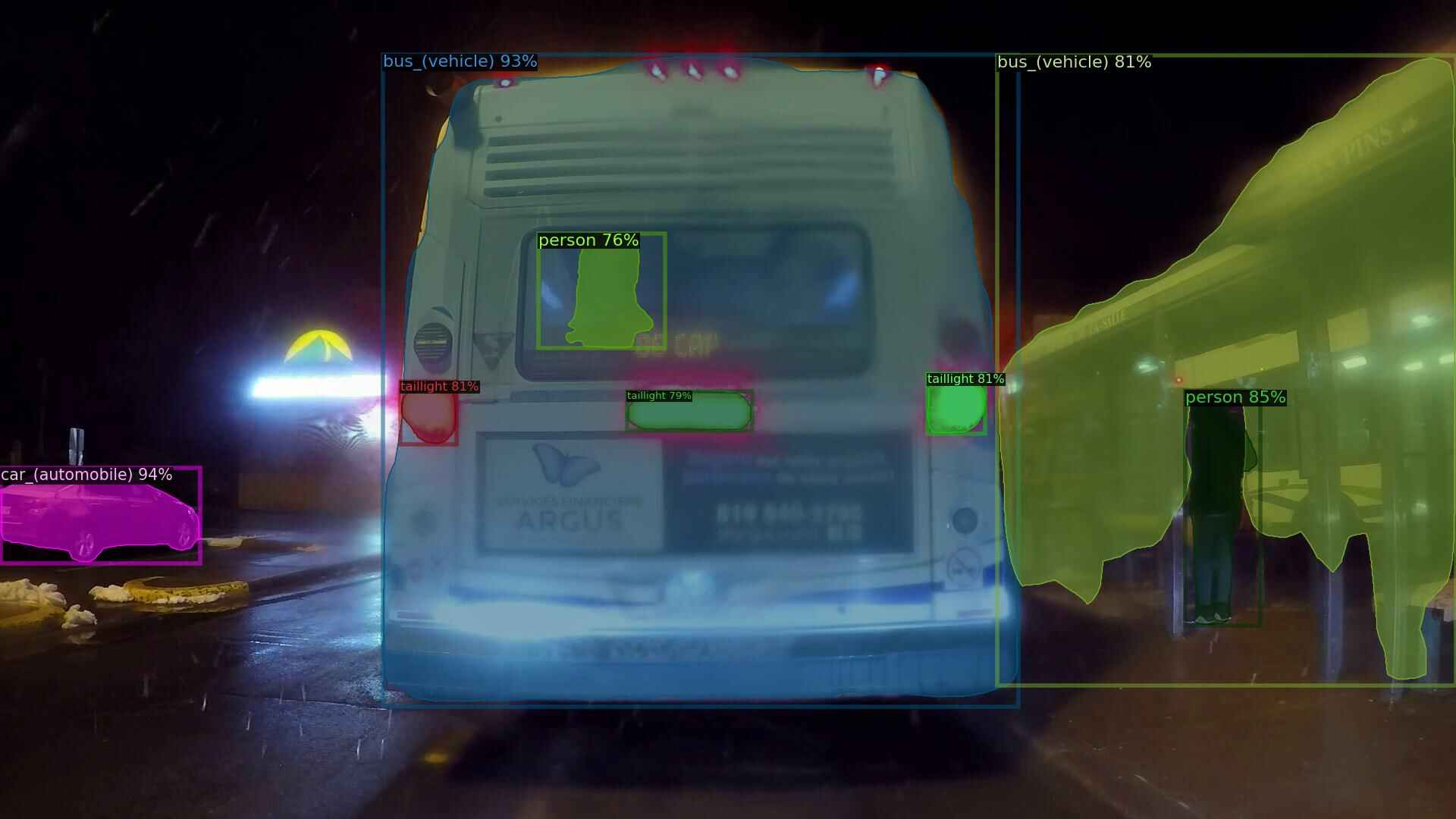}}
\end{subfigure}
\begin{subfigure}{1.3in}
{\includegraphics[width=1.3in,height=1.3in,keepaspectratio]{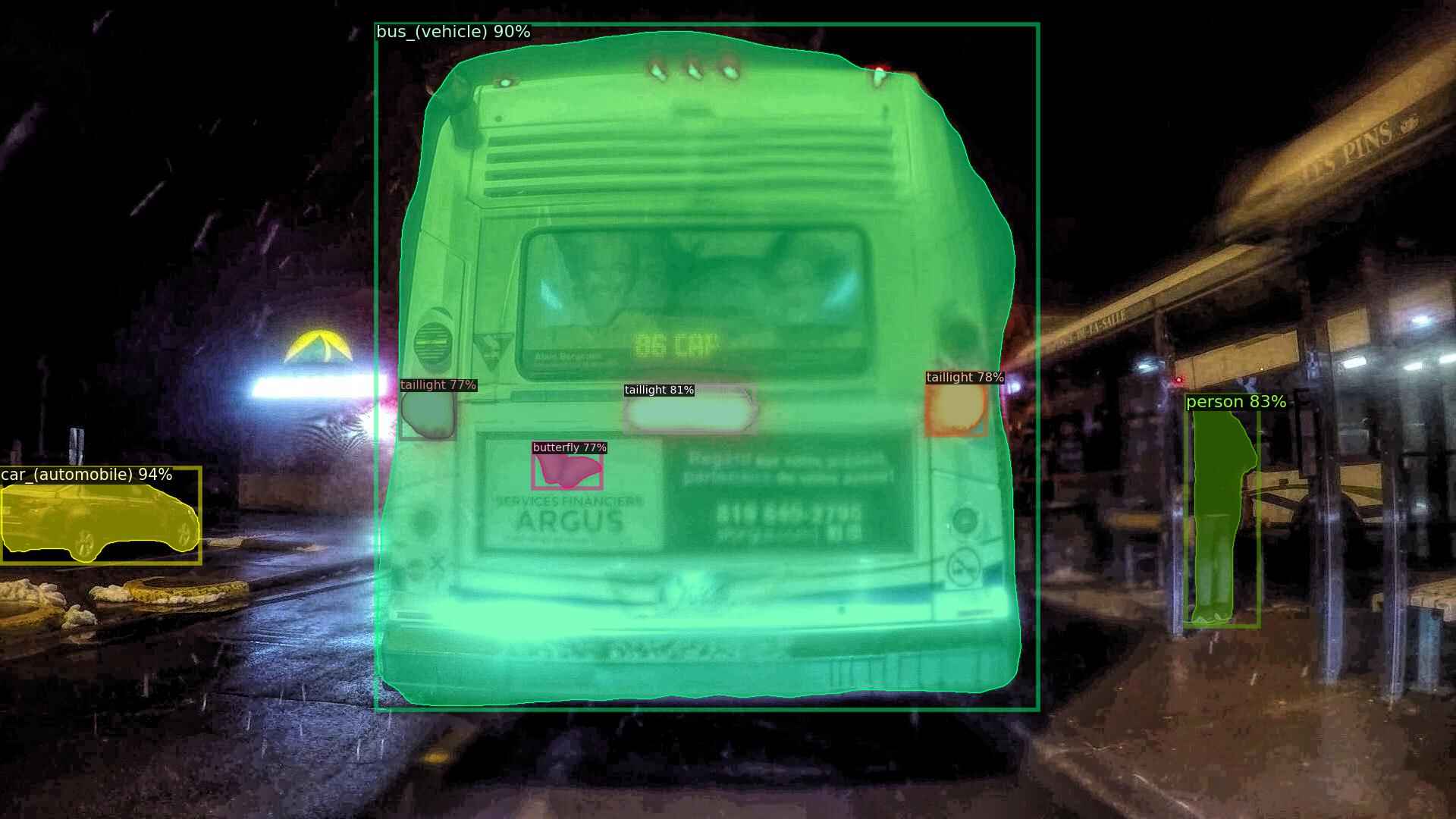}}
\end{subfigure}
\begin{subfigure}{1.3in}
{\includegraphics[width=1.3in,height=1.3in,keepaspectratio]{images/datapaper/171_reflection.jpg}}
\end{subfigure}
\begin{subfigure}{1.3in}
{\includegraphics[width=1.3in,height=1.3in,keepaspectratio]{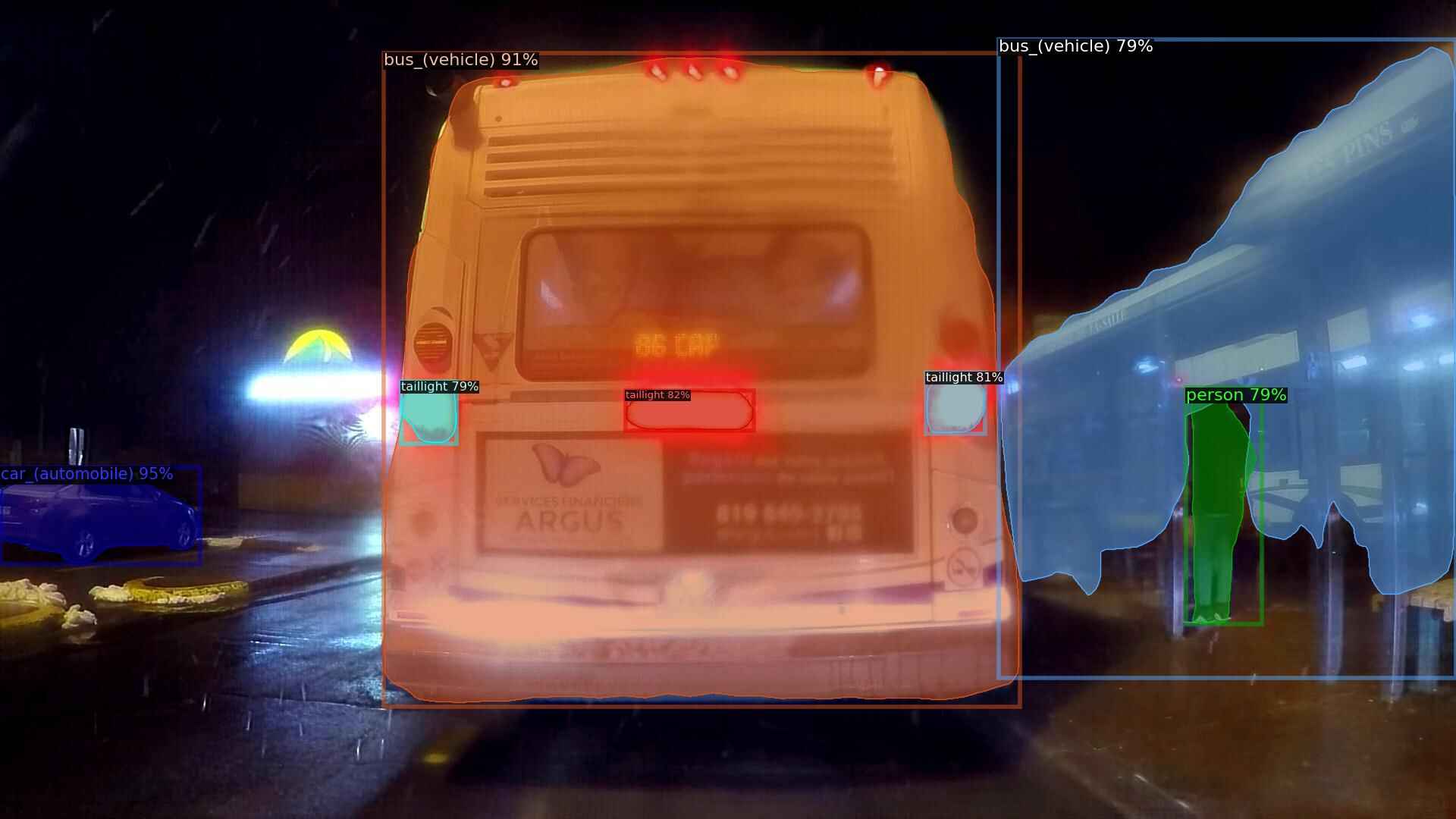}}
\end{subfigure}\\
\hspace{-0.04cm}
\raisebox{.9\height}{ \rotatebox[origin=]{90}{Bottom}}
\begin{subfigure}{1.3in}
{\includegraphics[width=1.3in,height=1.3in,keepaspectratio]{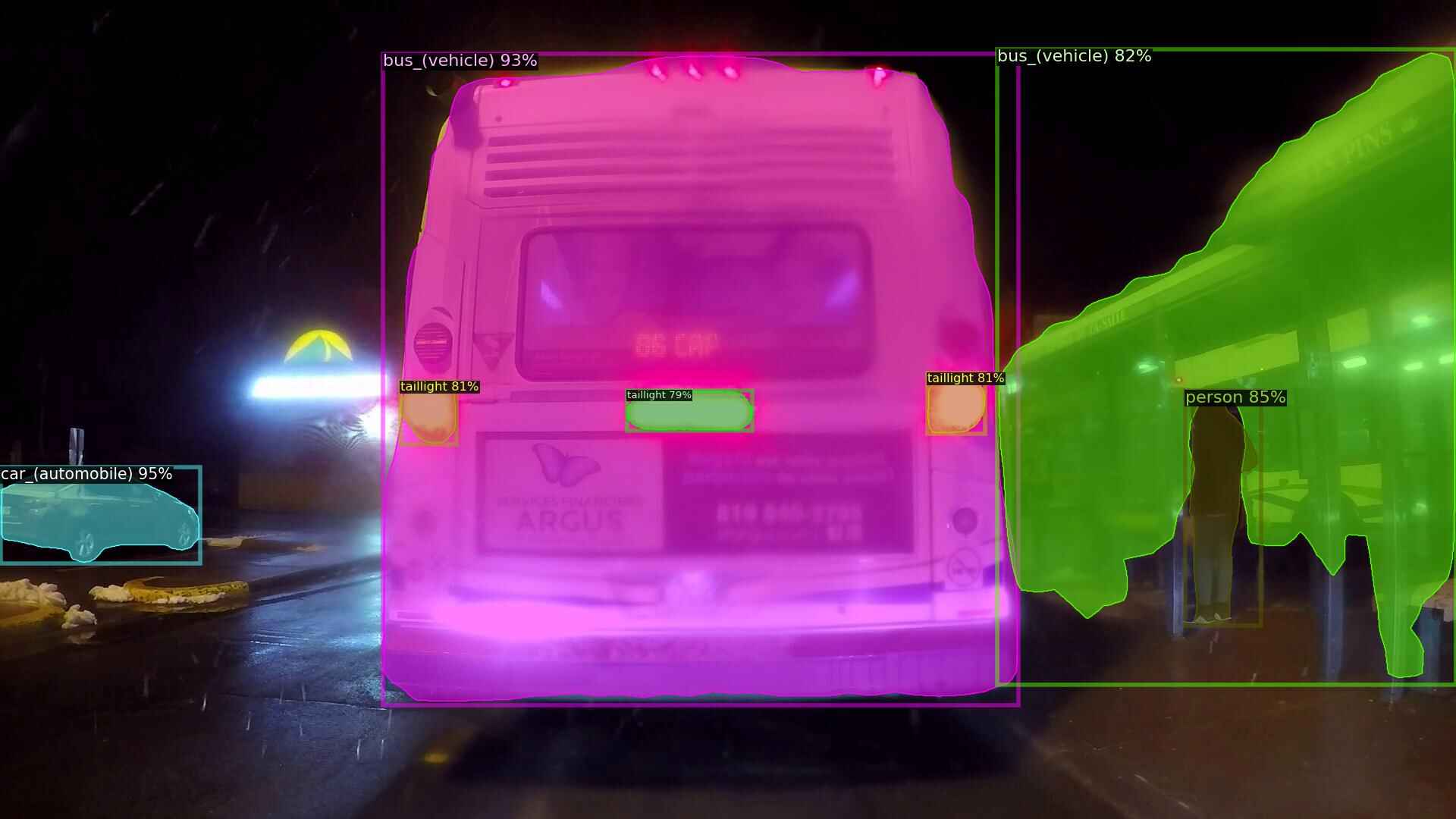}}
\end{subfigure}
\begin{subfigure}{1.3in}
{\includegraphics[width=1.3in,height=1.3in,keepaspectratio]{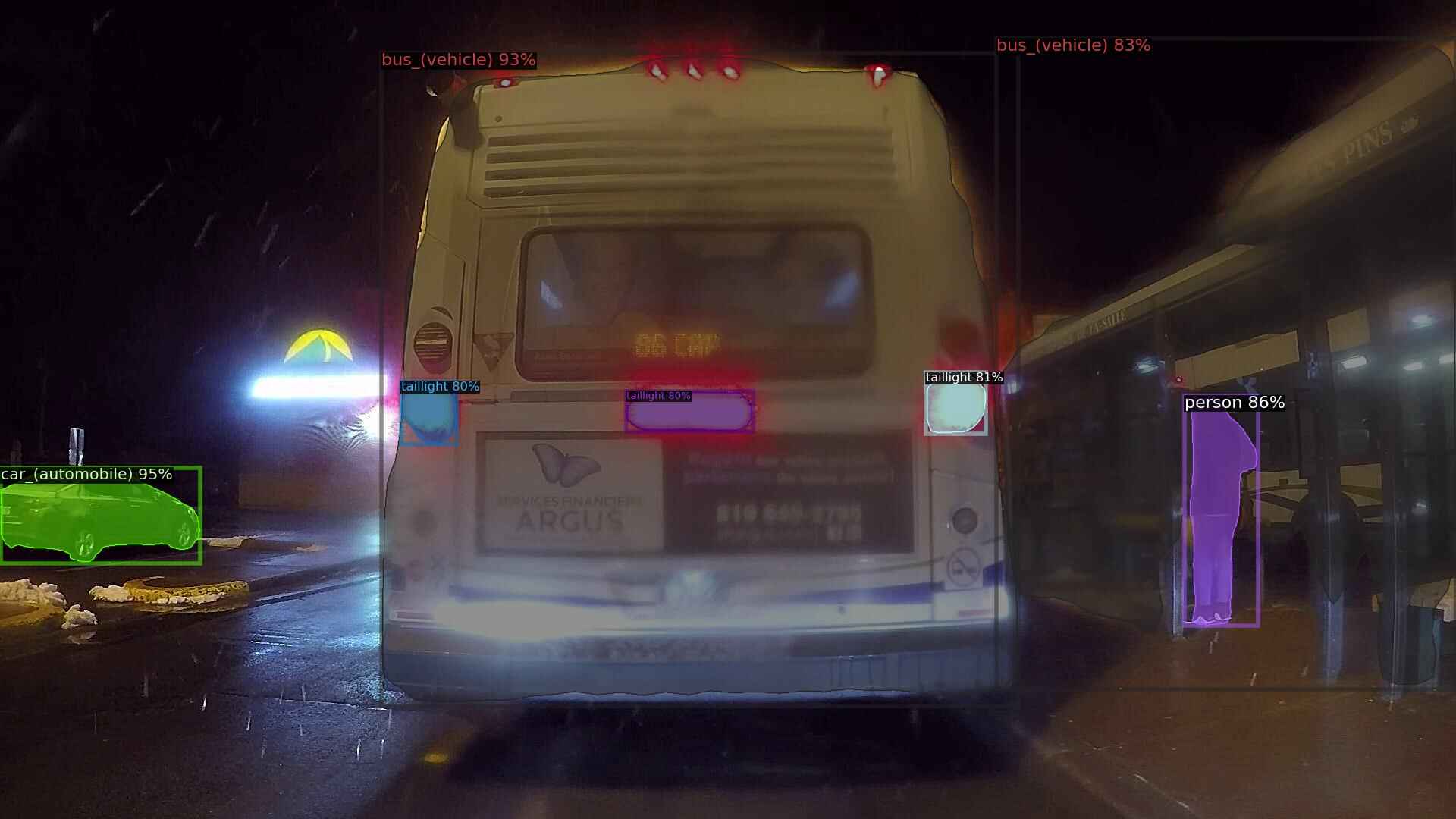}}
\end{subfigure}
\begin{subfigure}{1.3in}
{\includegraphics[width=1.3in,height=1.3in,keepaspectratio]{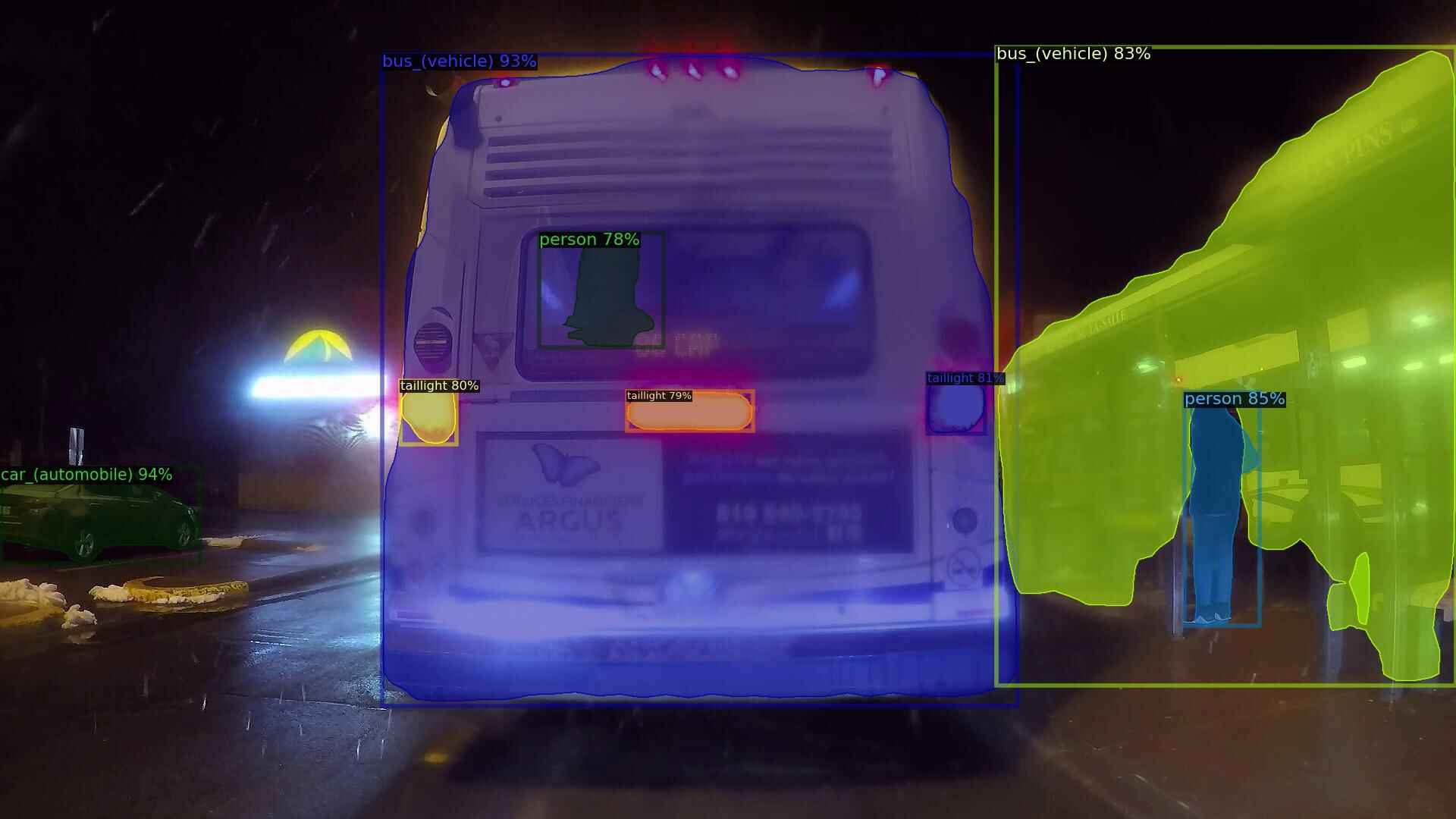}}
\end{subfigure}
\begin{subfigure}{1.3in}
{\includegraphics[width=1.3in,height=1.3in,keepaspectratio]{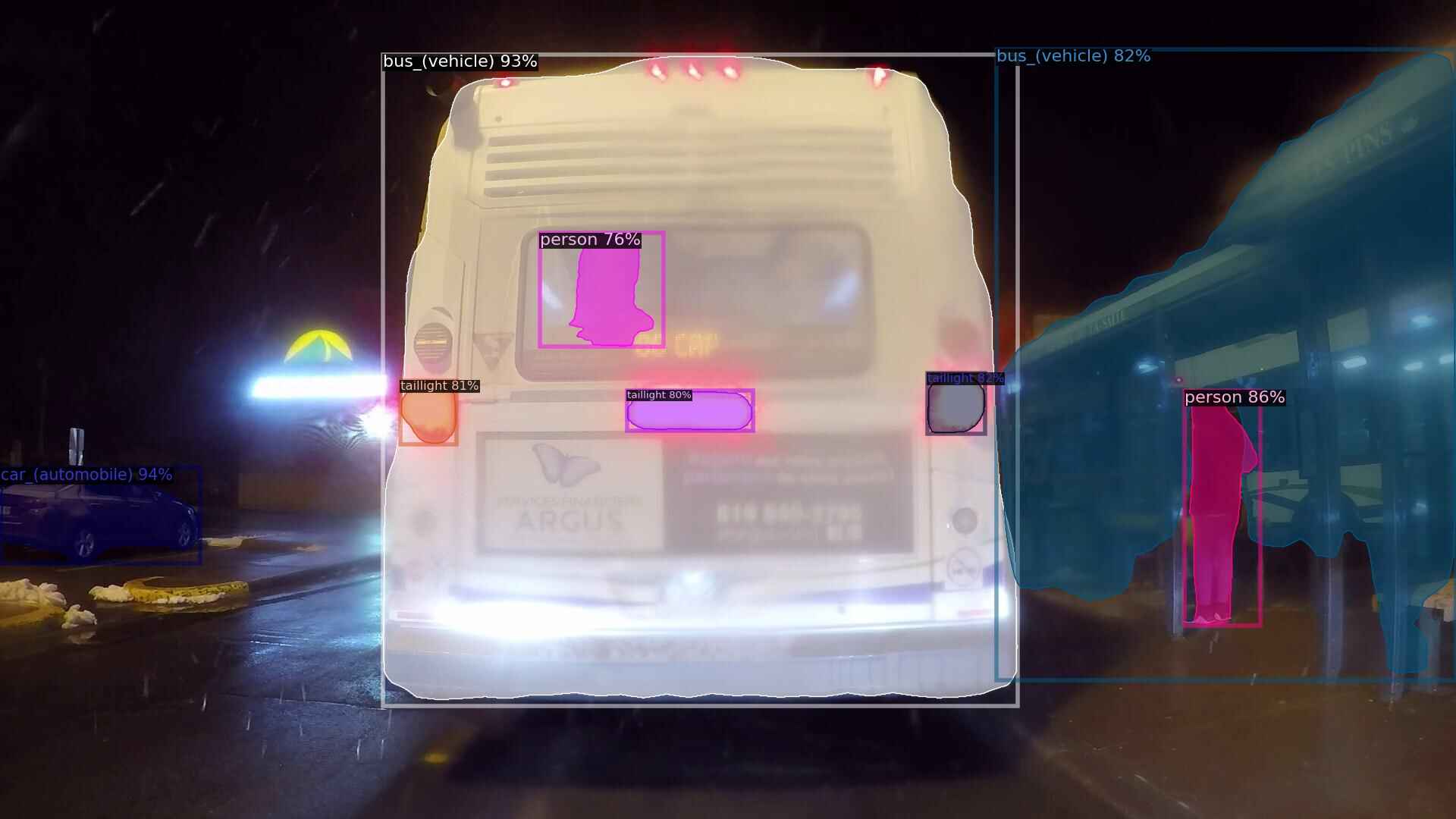}}
\end{subfigure}\\
\raisebox{.9\height}{ \rotatebox[origin=]{90}{TOP}}
\begin{subfigure}{1.3in}
{\includegraphics[trim={75 70 75 75},clip,width=1.3in,height=1.3in,keepaspectratio]{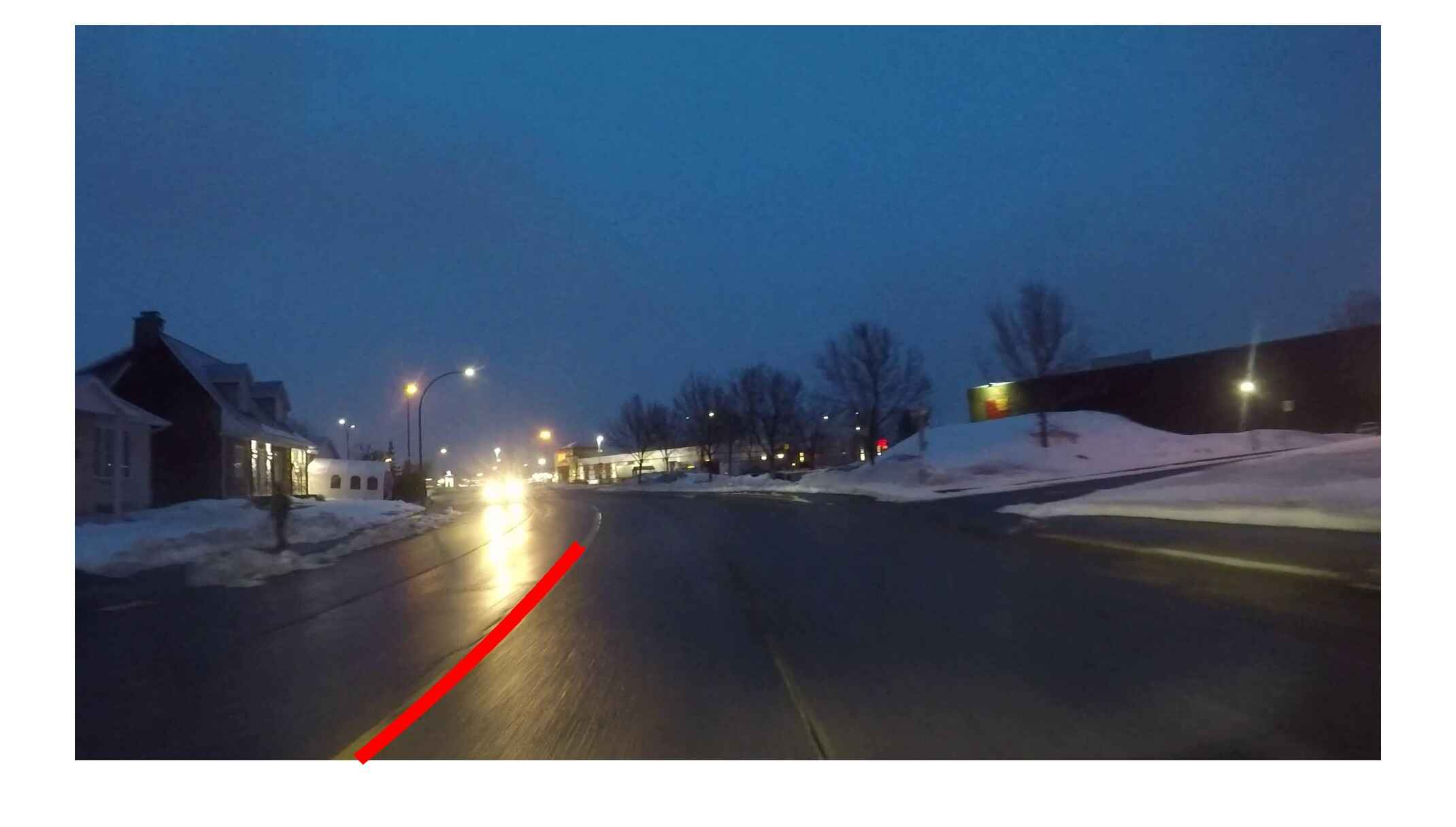}}
\end{subfigure}
\begin{subfigure}{1.3in}
{\includegraphics[trim={75 70 75 75 },clip,width=1.3in,height=1.3in,keepaspectratio]{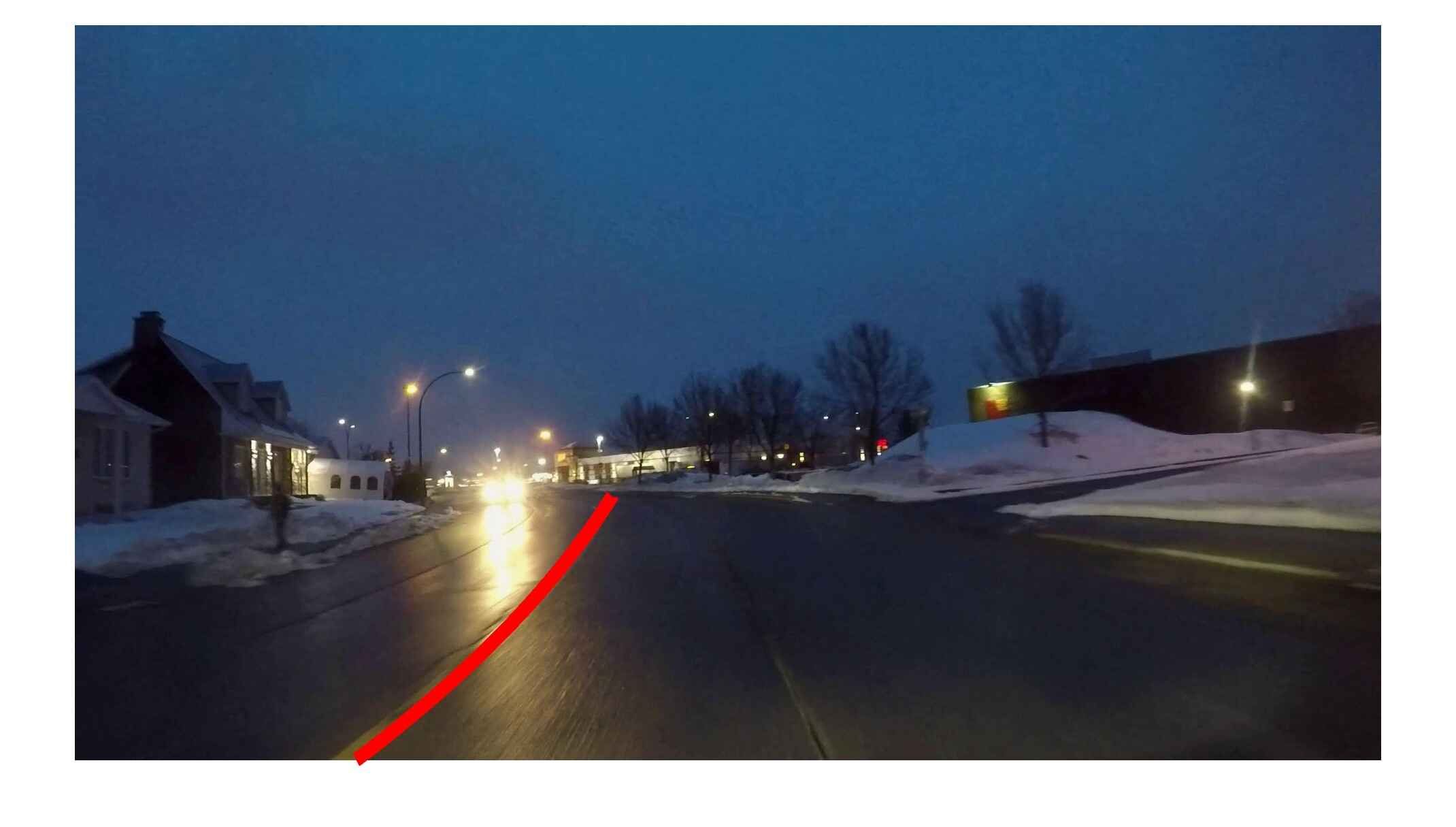}}
\end{subfigure}
\begin{subfigure}{1.3in}
{\includegraphics[trim={75 70 75 75 },clip,width=1.3in,height=1.3in,keepaspectratio]{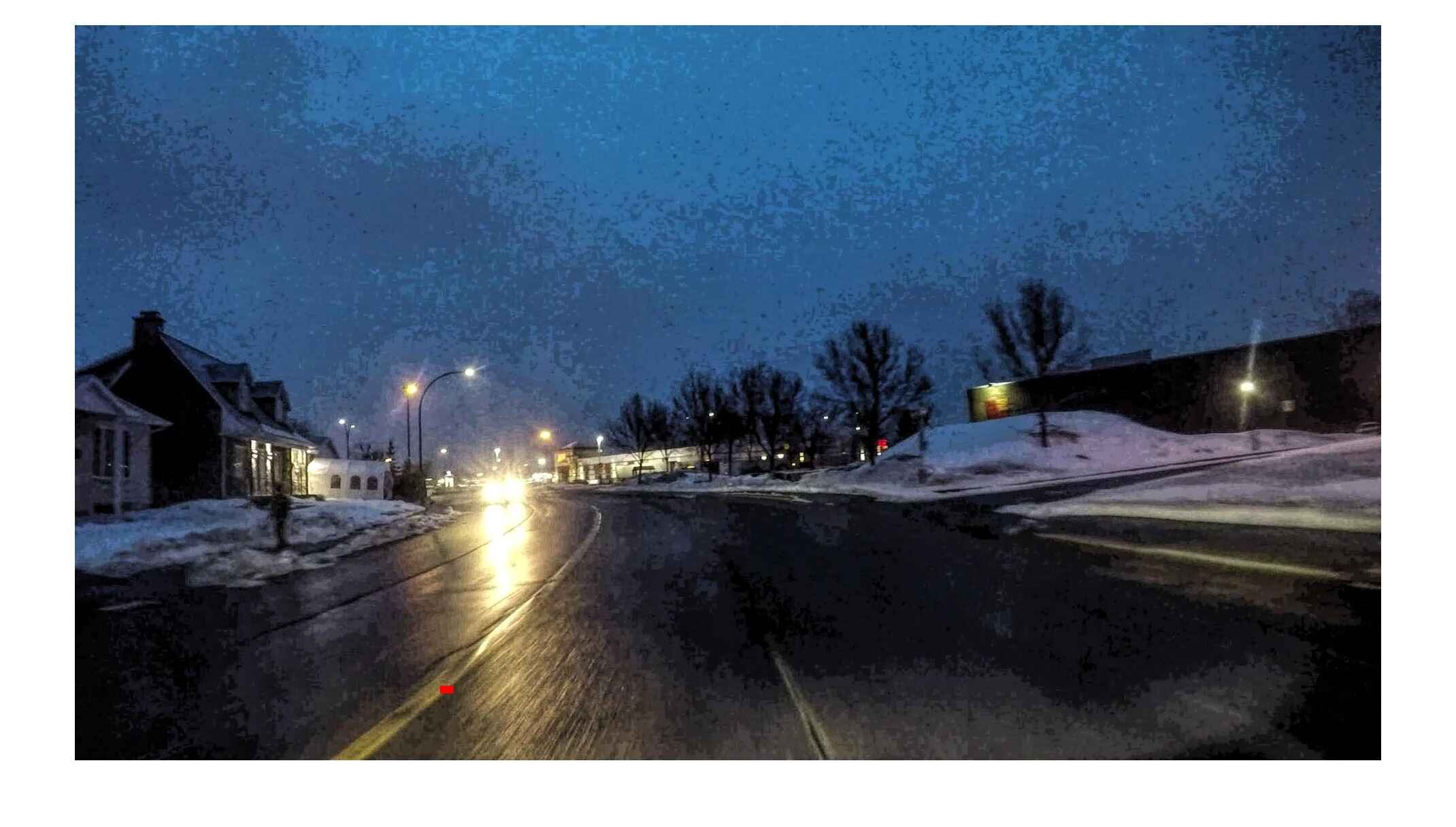}}
\end{subfigure}
\begin{subfigure}{1.3in}
{\includegraphics[trim={75 70 75 75},clip,width=1.3in,height=1.3in,keepaspectratio]{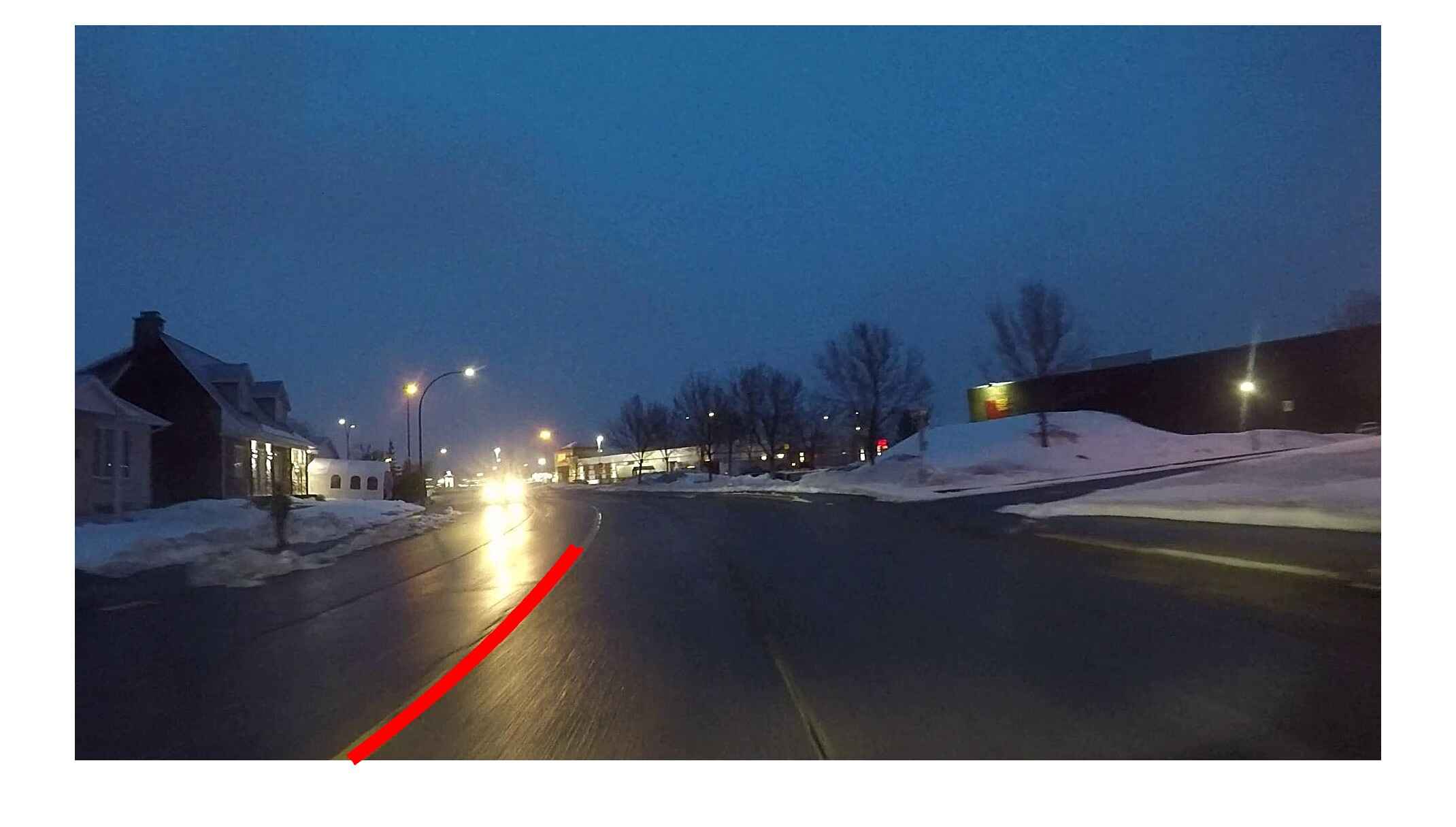}}
\end{subfigure}
\begin{subfigure}{1.3in}
{\includegraphics[trim={80 70 75 8},clip,width=1.3in,height=1.3in,keepaspectratio]{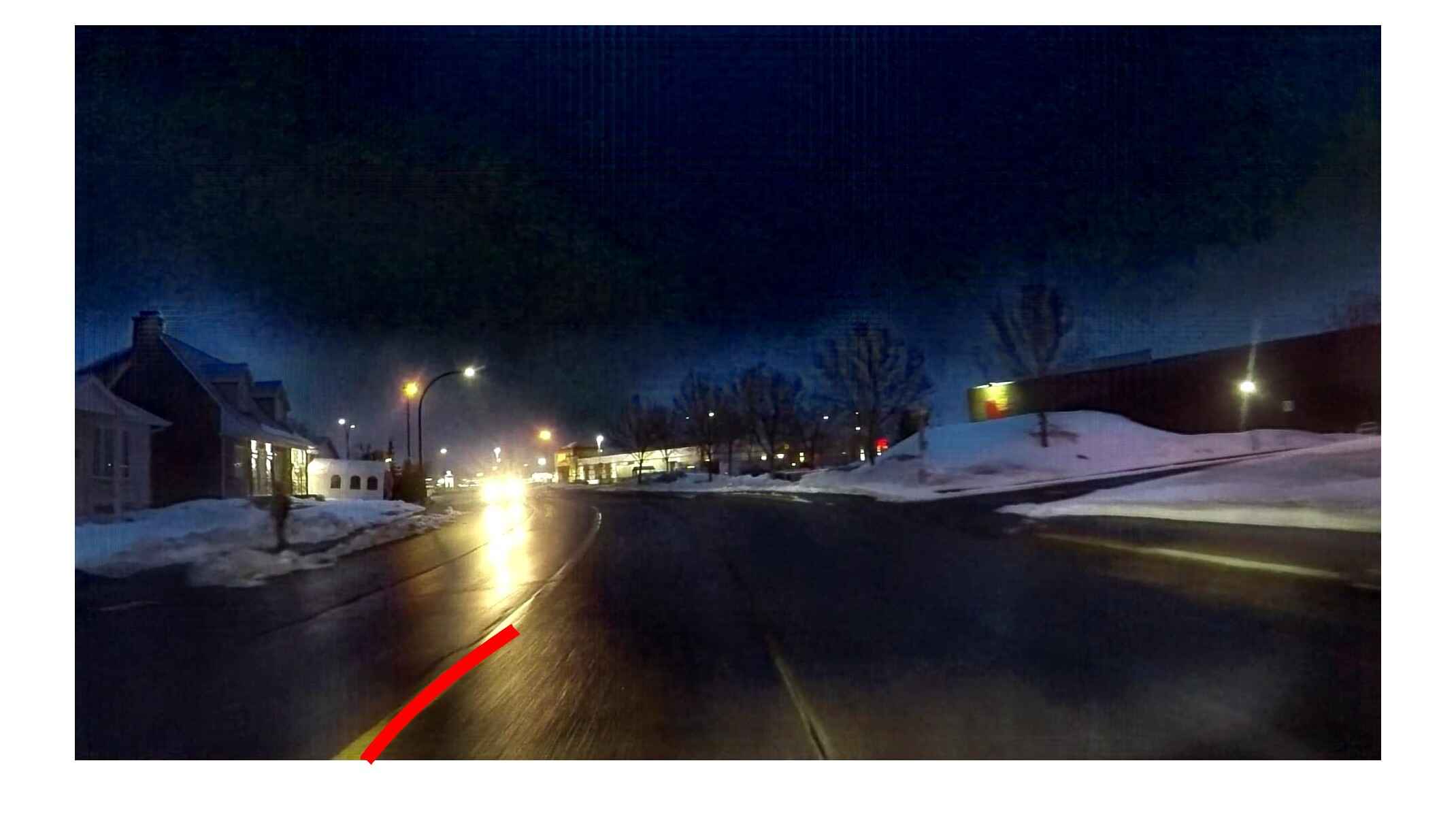}}
\end{subfigure}\\
\hspace{-0.08cm}
\raisebox{.9\height}{ \rotatebox[origin=]{90}{Bottom}}
\begin{subfigure}{1.3in}
{\includegraphics[trim={75 70 75 75},clip,width=1.3in,height=1.3in,keepaspectratio]{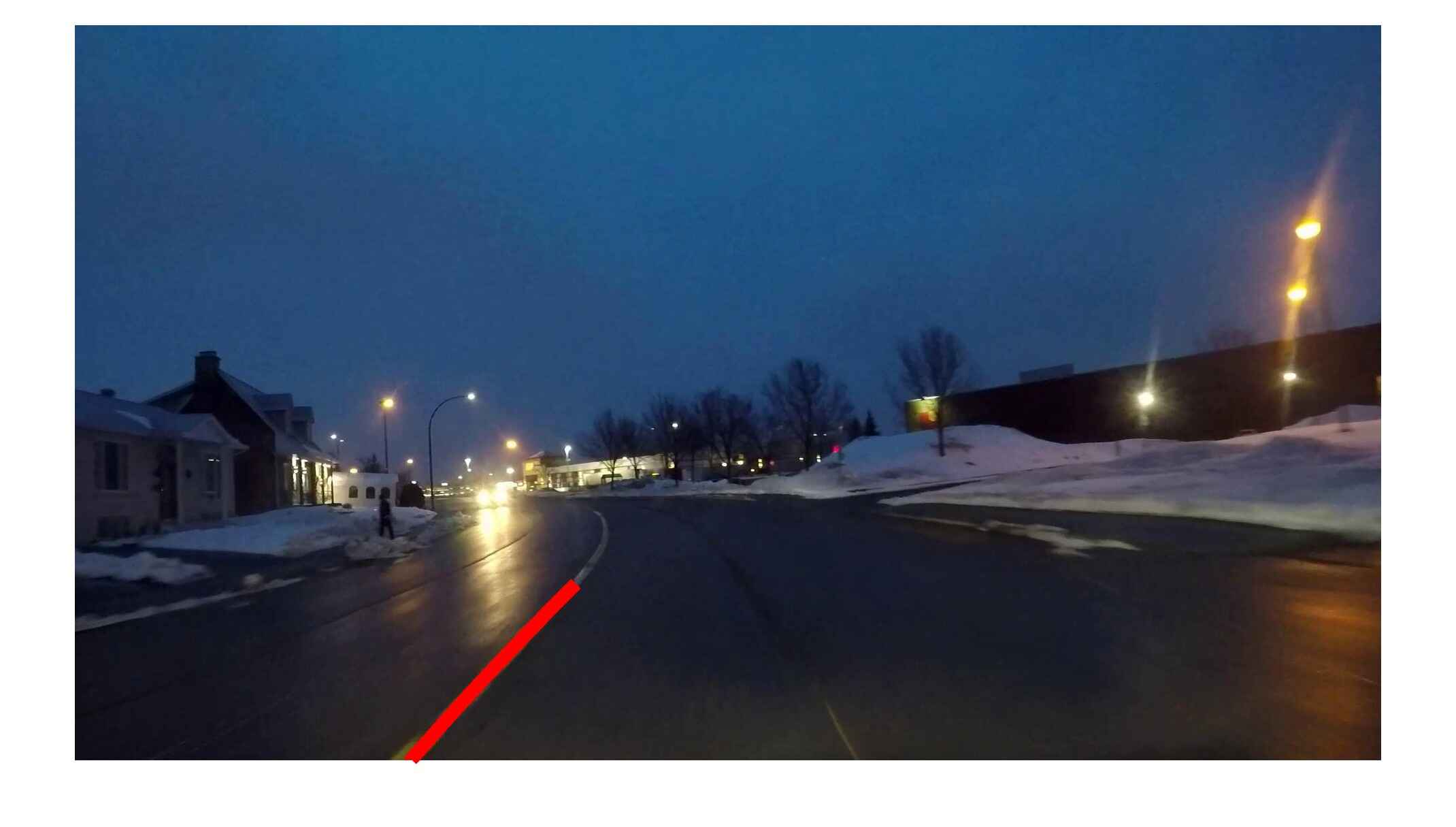}}
\end{subfigure}
\begin{subfigure}{1.3in}
{\includegraphics[trim={75 70 75 75},clip,width=1.3in,height=1.3in,keepaspectratio]{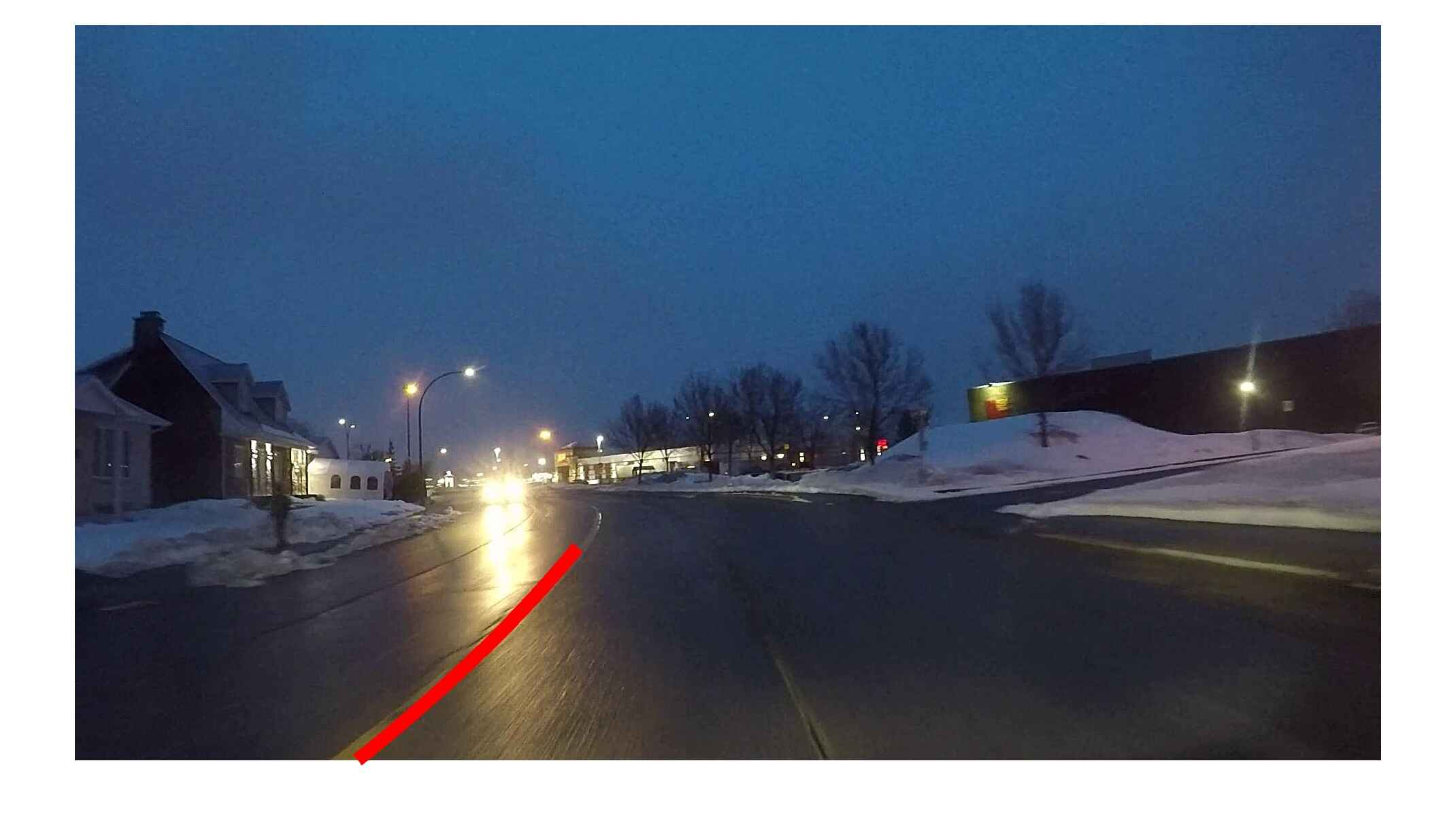}}
\end{subfigure}
\begin{subfigure}{1.3in}
{\includegraphics[trim={75 70 75 75},clip,width=1.3in,height=1.3in,keepaspectratio]{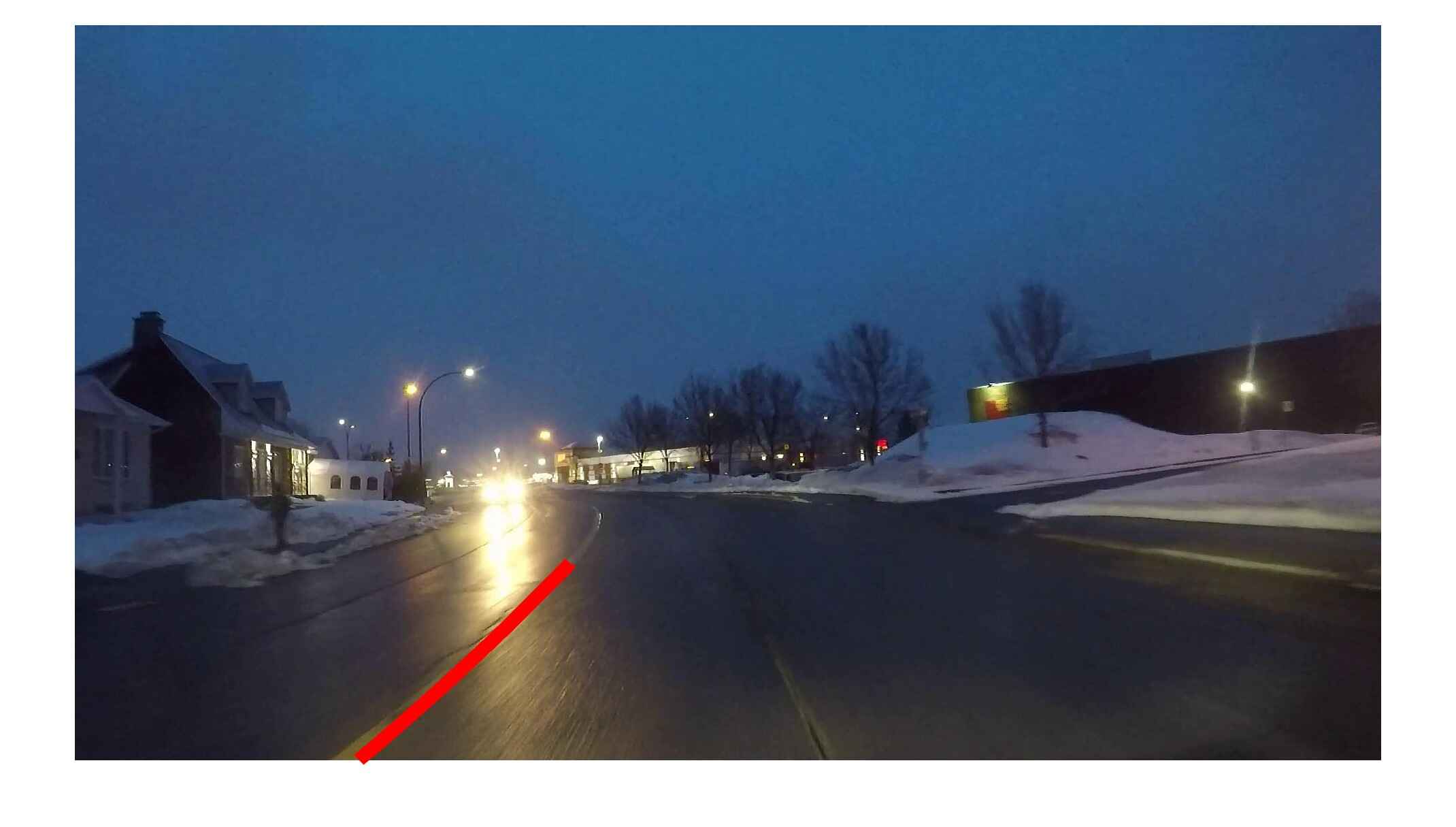}}
\end{subfigure}
\begin{subfigure}{1.3in}
{\includegraphics[trim={75 70 75 75},clip,width=1.3in,height=1.3in,keepaspectratio]{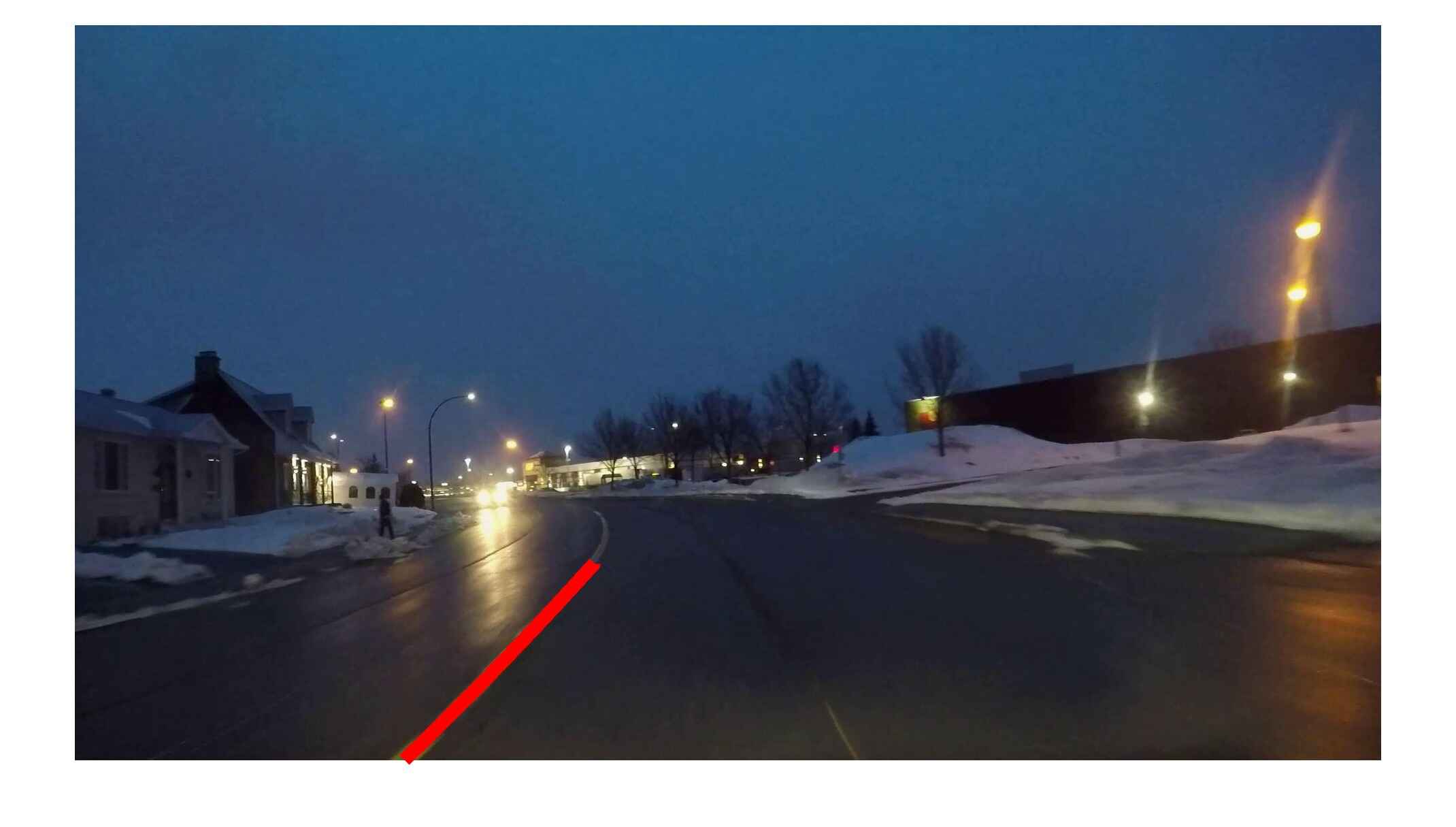}}
\end{subfigure}\\
\caption{Qualitative comparison of the randomly selected output of CV algorithms on a real dataset \cite{boisclair2022attention} used to produce the results in Fig. \ref{fig:Cv_comp_real}. (Top: Left to Right) Input, Proposed, Laplacian filter, Reflection removal \cite{Refl}, PFFNet \cite{mei2018pffn}. (Bottom: Left to Right) C2PNet \cite{zheng2023curricular}, Unsharp masking, Deblur \cite{Deblur_chen}, and Wiener filter. (Row 1-8) Object detection\textbackslash recognition. (Row 9-10) Lane detection. }
\label{fig:Cv_compviz_real}
\end{figure*}

In this section, we present the quantitative and qualitative evaluation of the proposed glare reduction method for improved performance in the perception tasks of autonomous vehicles. We used the tunnel scene for the evaluation as it exhibits high dynamic range and offers a realistic scenario with multiple sources for glare.  The selected dataset provides images of the same scene under uniform lighting conditions, crucial for estimating the ground truth for several perception tasks such as depth estimation, and object tracking. Additionally, the dataset provides sufficient glare variation through different exposure captures, providing ample data for robust evaluation of glare reduction methods. \\Additionally, we have compared the performance of the proposed method on a real dataset \cite{boisclair2022attention} captured from an autonomous car using a GoPro HERO5 camera. The testing dataset in this paper consists of a subset of images from \cite{boisclair2022attention} that contain strong sources of glare i.e all the nighttime images, made publicly available\footnote{\url{https://github.com/irh-ca-team-car/attention-data/tree/master/DSF/images}}.\\
We have compared the proposed method with a variety of techniques applicable to the glare reduction problem. Particularly, we explore five distinct categories of techniques aimed at reducing the impact of glare on the perception tasks of autonomous vehicles. These include basic deconvolution-based methods, such as the Wiener filter approach that does not account for saturated pixels. Furthermore, we investigate image-dehazing methods, specifically C2PNet \cite{zheng2023curricular} and PFFNet \cite{mei2018pffn}. Techniques for image enhancement, such as unsharp masking and Local Laplacian filtering, are also considered. Additionally, we assess the effectiveness of reflection removal  \cite{Refl} and image deblurring \cite{Deblur_chen} techniques. \\
The selection of these different methods is based on their applicability to the glare problem and varying computational complexity. For example, like glare, haze also adds the reflective component of light to different sensor locations. In the case of haze, light scatters off suspended particles or molecules in the atmosphere, leading to reduced contrast, color distortion, and loss of details in the image. Glare occurs when light scatters and reflects within the camera lens or optical system, resulting in the same unwanted artifacts.\\ Image Enhancement can be a simple and computationally efficient alternative to image deconvolution, and image dehazing, for mitigating the adverse effect of glare and improving the performance of the autonomous vehicle perception module.\\ Image deblurring techniques aim to correct images that have been degraded by blur or other factors that reduce image clarity. These methods, while effective at low to moderate exposure levels, do not perform well under conditions of intense glare. In \cite{Deblur_chen} pixel saturation is specifically considered and is, therefore, included in our comparison. Reflection removal methods target degradation caused by reflections within the camera lens or optical system. Reflection removal is particularly important in scenarios where glare is not due to external light sources but rather due to internal reflections in the camera.\\
In Fig. \ref{fig:Rcamercomp}, \ref{fig:Cv_comp_real}, and \ref{fig:Cv_comp} we demonstrate the effectiveness of the proposed joint GSFs estimation technique on perception tasks. In Fig. \ref{fig:Rcamercomp} the joint GSF is estimated for the camera lens combinations discussed in \ref{Joint GSF Estimation} and the dataset used was captured using the camera and lens specified in Section \ref{sec:testing frame work}. Notably, as depicted in Fig. \ref{fig:Rcamercomp}, our joint GSF estimation-based approach demonstrates consistent performance despite the addition of different camera-lens combinations involved in GSF estimation. Unlike Fig. \ref{fig:Rcamercomp}, where a uniformity across multiple camera-lens performance is visible, in Figure \ref{fig:Cv_comp_real} the Joint GSF method slightly outperforms the specific GSF estimated even for the same camera, GoPro HERO5, used in the data acquisition. This variation can be attributed to the differences in datasets. In Fig \ref{fig:Rcamercomp} the in-lab captured dataset is used that offers insignificant depth variation in comparison to the real dataset presented in Fig. \ref{fig:Cv_comp_real}. \\
The effectiveness of the joint GSF in comparison to a camera specific GSF becomes prominent in scenarios where the AVs perception markedly diminishes when the data is captured from a camera that is not included in specific GSF estimation, as depicted in Fig. \ref{fig:Cv_comp} Sony $\alpha$7R1 plot.
\\
Fig. \ref{fig:Cv_comp}, TABLE  \ref{summary}, and  Fig. \ref{fig:Cv_compviz} reveal noticeable variations in the performance of various glare reduction methods across different perception tasks. Due to the correlation between increasing brightness and heightened glare, under highest exposure values, the performance of glared input tends to deteriorate significantly. Nevertheless, across the spectrum of computer vision tasks,  the performance of various glare reduction methods exhibit distinctive trends, which are crucial to understanding their effective application in specific scenarios. Deconvolution-based methods, in particular the proposed method demonstrate a notable proficiency in managing glare across different tasks and stands out for its consistent high (above average metric value reported for the task) performance especially at higher exposure levels. This method adeptly preserves the characteristics of the input image while minimizing glare, a feature not as effectively mirrored by the Wiener Filter and Veiling glare removal \cite{Talvala}. \\
Dehazing methods present a contrast, especially in object recognition (OR), where PFFNet \cite{mei2018pffn}, manages to perform decently while C2PNet \cite{zheng2023curricular} completely fails. While these methods reduce overall brightness, which can be advantageous in high exposure settings, the trade-off often comes in the form of reduced clarity and the introduction of dark spots, notably impacting tasks like OD and DE. Interestingly, object tracking (OT), seems to be slightly less impacted by the aforementioned artifacts, where dehazing methods outperform computationally efficient, image enhancement methods at moderate to high glare levels, revealing their selective utility.\\
For image enhancement techniques, such as local Laplacian filtering and unsharp masking, we observe a divergence in effectiveness. While these methods enhance contrast, they often introduce significant artifacts in the presence of glare, which impact their performance as seen in their reduced efficiency in OD. However, in OR, especially at higher exposures, they demonstrate their ability to improve performance, marking their selective applicability based on the specific task and exposure conditions.\\
Deblur and Reflection removal methods hold their unique place in glare reduction. Deblur techniques, while effective at low to moderate exposure levels in some tasks, tend to falter as glare intensifies. Reflection removal, might be a reliable choice in consistent lighting conditions but fails in variable lighting environments.\\
Overall, the performance of glare reduction methods in computer vision tasks is not monolithic but varies significantly across different categories and within them. Deconvolution-based methods, led by the proposed method, generally excel in handling glare, particularly at higher exposures. This observation is further exemplified in scenarios like the Tunnel scene, where objects predominantly lie in darker dynamic ranges. However, it's crucial to note the performance drops at very high exposure levels, highlighting the challenge these methods face in extremely intense glare conditions. A summary of performance evaluation for the aforementioned glare reduction categories is presented in TABLE  \ref{summary}.
\begin{table*}[!h]
\centering
\caption{Performance of glare reduction methods across computer vision tasks at different exposure levels. $\color{arrowblue}\uparrow$ indicates effective performance, $\rightarrow$ indicates moderate performance, and $\color{arrowblue}\downarrow$ indicates decreased effectiveness/struggles.}
\begin{tabular}{|l|ccc|ccc|ccc|ccc|ccc|}
\hline
\textbf{Method Category} & \multicolumn{3}{c|}{\textbf{Object Detection}} & \multicolumn{3}{c|}{\textbf{Object Recognition}} & \multicolumn{3}{c|}{\textbf{Object Tracking}} & \multicolumn{3}{c|}{\textbf{Lane Detection}} & \multicolumn{3}{c|}{\textbf{Depth Estimation}} \\
 & Low & Mod & High & Low & Mod & High & Low & Mod & High & Low & Mod & High & Low & Mod & High \\ \hline
\textbf{Convolutional Methods} &  &  &  &  &  &  &  &  &  &  &  &  &  &  &  \\
Proposed Method & $\color{arrowblue}\downarrow$ & $\uparrow$ & $\color{arrowblue}\rightarrow$ & $\color{arrowblue}\uparrow$ & $\uparrow$ & $\color{arrowblue}\uparrow$ & $\color{arrowblue}\uparrow$ & $\uparrow$ & $\color{arrowblue}\uparrow$ & $\color{arrowblue}\uparrow$ & $\uparrow$ & $\color{arrowblue}\rightarrow$ & $\color{arrowblue}\uparrow$ & $\uparrow$ & $\color{arrowblue}\uparrow$ \\ \hline
Wiener Filter & $\color{arrowblue}\downarrow$ & $\rightarrow$ & $\color{arrowblue}\rightarrow$ & $\color{arrowblue}\uparrow$ & $\uparrow$ & $\color{arrowblue}\uparrow$ & $\color{arrowblue}\downarrow$ & $\rightarrow$ & $\color{arrowblue}\downarrow$ & $\color{arrowblue}\uparrow$ & $\uparrow$ & $\color{arrowblue}\downarrow$ & $\color{arrowblue}\uparrow$ & $\uparrow$ & $\color{arrowblue}\downarrow$ \\ \hline
Veiling Glare \cite{Talvala} & $\color{arrowblue}\downarrow$ & $\rightarrow$ & $\color{arrowblue}\rightarrow$ & $\color{arrowblue}\uparrow$ & $\uparrow$ & $\color{arrowblue}\uparrow$ & $\color{arrowblue}\downarrow$ & $\rightarrow$ & $\color{arrowblue}\downarrow$ & $\color{arrowblue}\uparrow$ & $\uparrow$ & $\color{arrowblue}\downarrow$ & $\color{arrowblue}\uparrow$ & $\uparrow$ & $\color{arrowblue}\downarrow$ \\ \hline
\textbf{Dehazing} &  &  &  &  &  &  &  &  &  &  &  &  &  &  &  \\
C2PNet \cite{zheng2023curricular} & $\color{arrowblue}\rightarrow$ & $\rightarrow$ & $\color{arrowblue}\downarrow$ & $\color{arrowblue}\downarrow$ & $\downarrow$ & $\color{arrowblue}\rightarrow$ & $\color{arrowblue}\uparrow$ & $\uparrow$ & $\color{arrowblue}\downarrow$ & $\color{arrowblue}\uparrow$ & $\uparrow$ & $\color{arrowblue}\downarrow$ & $\color{arrowblue}\downarrow$ & $\downarrow$ & $\color{arrowblue}\downarrow$ \\ \hline
PFFNet \cite{mei2018pffn}& $\color{arrowblue}\downarrow$ & $\uparrow$ & $\color{arrowblue}\downarrow$ & $\color{arrowblue}\rightarrow$ & $\uparrow$ & $\color{arrowblue}\uparrow$ & $\color{arrowblue}\rightarrow$ & $\uparrow$ & $\color{arrowblue}\downarrow$ & $\color{arrowblue}\uparrow$ & $\uparrow$ & $\color{arrowblue}\downarrow$ & $\color{arrowblue}\downarrow$ & $\rightarrow$ & $\color{arrowblue}\downarrow$ \\ \hline
\textbf{Image Enhancement} &  &  &  &  &  &  &  &  &  &  &  &  &  &  &  \\
Laplacian & $\color{arrowblue}\downarrow$ & $\rightarrow$ & $\color{arrowblue}\rightarrow$ & $\color{arrowblue}\rightarrow$ & $\uparrow$ & $\color{arrowblue}\uparrow$ & $\color{arrowblue}\rightarrow$ & $\rightarrow$ & $\color{arrowblue}\downarrow$ & $\color{arrowblue}\uparrow$ & $\uparrow$ & $\color{arrowblue}\uparrow$ & $\color{arrowblue}\rightarrow$ & $\uparrow$ & $\color{arrowblue}\uparrow$ \\
Unsharp Masking & $\color{arrowblue}\downarrow$ & $\rightarrow$ & $\color{arrowblue}\rightarrow$ & $\color{arrowblue}\downarrow$ & $\rightarrow$ & $\color{arrowblue}\uparrow$ & $\color{arrowblue}\uparrow$ & $\uparrow$ & $\color{arrowblue}\downarrow$ & $\color{arrowblue}\uparrow$ & $\uparrow$ & $\color{arrowblue}\downarrow$ & $\color{arrowblue}\downarrow$ & $\rightarrow$ & $\color{arrowblue}\downarrow$ \\ \hline
\textbf{Deblur} &  &  &  &  &  &  &  &  &  &  &  &  &  &  &  \\
Deblur \cite{Deblur_chen}& $\color{arrowblue}\downarrow$ & $\downarrow$ & $\color{arrowblue}\downarrow$ & $\color{arrowblue}\rightarrow$ & $\rightarrow$ & $\color{arrowblue}\uparrow$ & $\color{arrowblue}\downarrow$ & $\rightarrow$ & $\color{arrowblue}\downarrow$ & $\color{arrowblue}\uparrow$ & $\uparrow$ & $\color{arrowblue}\downarrow$ & $\color{arrowblue}\uparrow$ & $\uparrow$ & $\color{arrowblue}\downarrow$ \\ \hline
\textbf{Reflection Removal} &  &  &  &  &  &  &  &  &  &  &  &  &  &  &  \\
Reflection Removal \cite{Refl} & $\color{arrowblue}\downarrow$ & $\rightarrow$ & $\color{arrowblue}\rightarrow$ & $\color{arrowblue}\downarrow$ & $\downarrow$ & $\color{arrowblue}\downarrow$ & $\color{arrowblue}\rightarrow$ & $\rightarrow$ & $\color{arrowblue}\downarrow$ & $\color{arrowblue}\uparrow$ & $\uparrow$ & $\color{arrowblue}\downarrow$ & $\color{arrowblue}\uparrow$ & $\uparrow$ & $\color{arrowblue}\uparrow$ \\ \hline
\end{tabular}
\label{summary}
\end{table*}
\\
\color{black}Previous research, as demonstrated in  \cite{buckler2017reconfiguring}, has highlighted that within the conventional Image Signal Processing (ISP) pipeline, two pivotal stages significantly affect the performance of machine vision algorithms: demosaicing and image encoding. Consequently, in our study, we employ three distinct image encoding techniques namely gamma, log, and linear encoding, as elaborated in Section \ref{sec:transfer-functions}.\\ 
\begin{figure*}[!h]
\centering
\begin{subfigure}{6.5in}
{\includegraphics[trim={0 38 140 105 },clip,width=6.5in,height=6.5in,keepaspectratio]{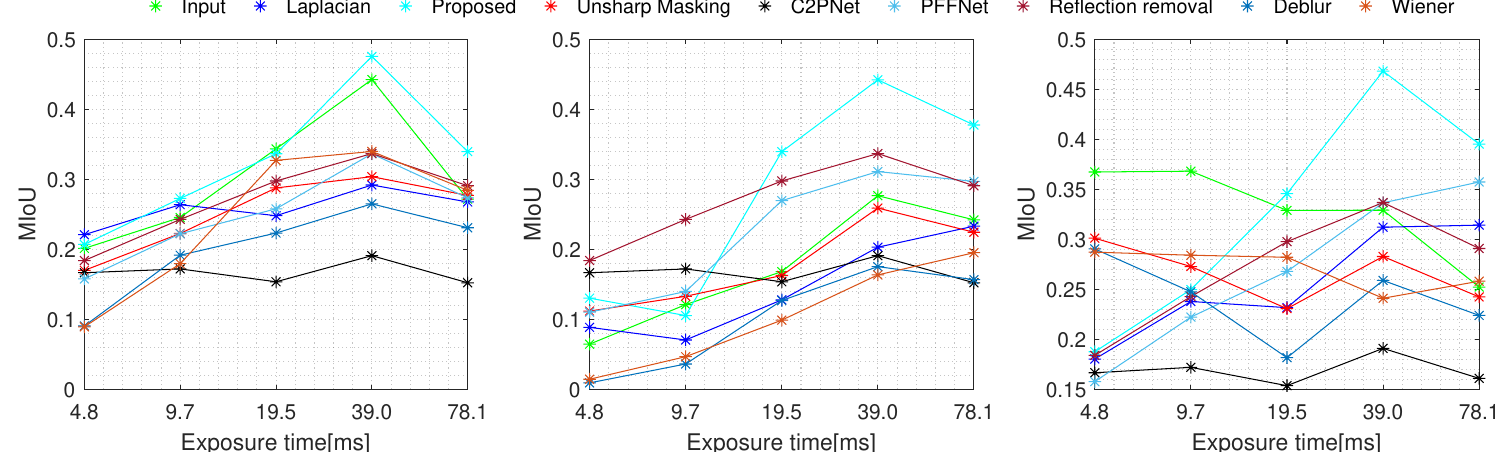}}
\end{subfigure}
\caption{Performance variation in object detection for different image encodings.(Left to Right) Gamma, linear, and log encodings.}
\label{fig:TF}
\end{figure*}
This deliberate selection allows us to comprehensively assess the effectiveness of various glare reduction methods while mitigating potential bias associated with the choice of encoding method. Although, the proposed approach displays a substantial superiority over other glare reduction methods, particularly evident in the object detection task, as illustrated in Fig. \ref{fig:TF} and \ref{fig:TFVIZ}, it is noteworthy that among the various image encodings tested, gamma encoding emerges as distinctly superior. It is essential to highlight that all the machine vision techniques evaluated in this paper were pre-trained using a large dataset comprising gamma-encoded images. Consequently, there exists a bias toward gamma encoding. Therefore, we have chosen gamma encoding in the assessment of glare reduction methods.\\
\begin{figure*}[!h]
\centering
\raisebox{.7\height}{ \rotatebox[origin=]{90}{Gamma}}
\begin{subfigure}{1.3in}
{\includegraphics[width=1.3in,height=1.0in,keepaspectratio]{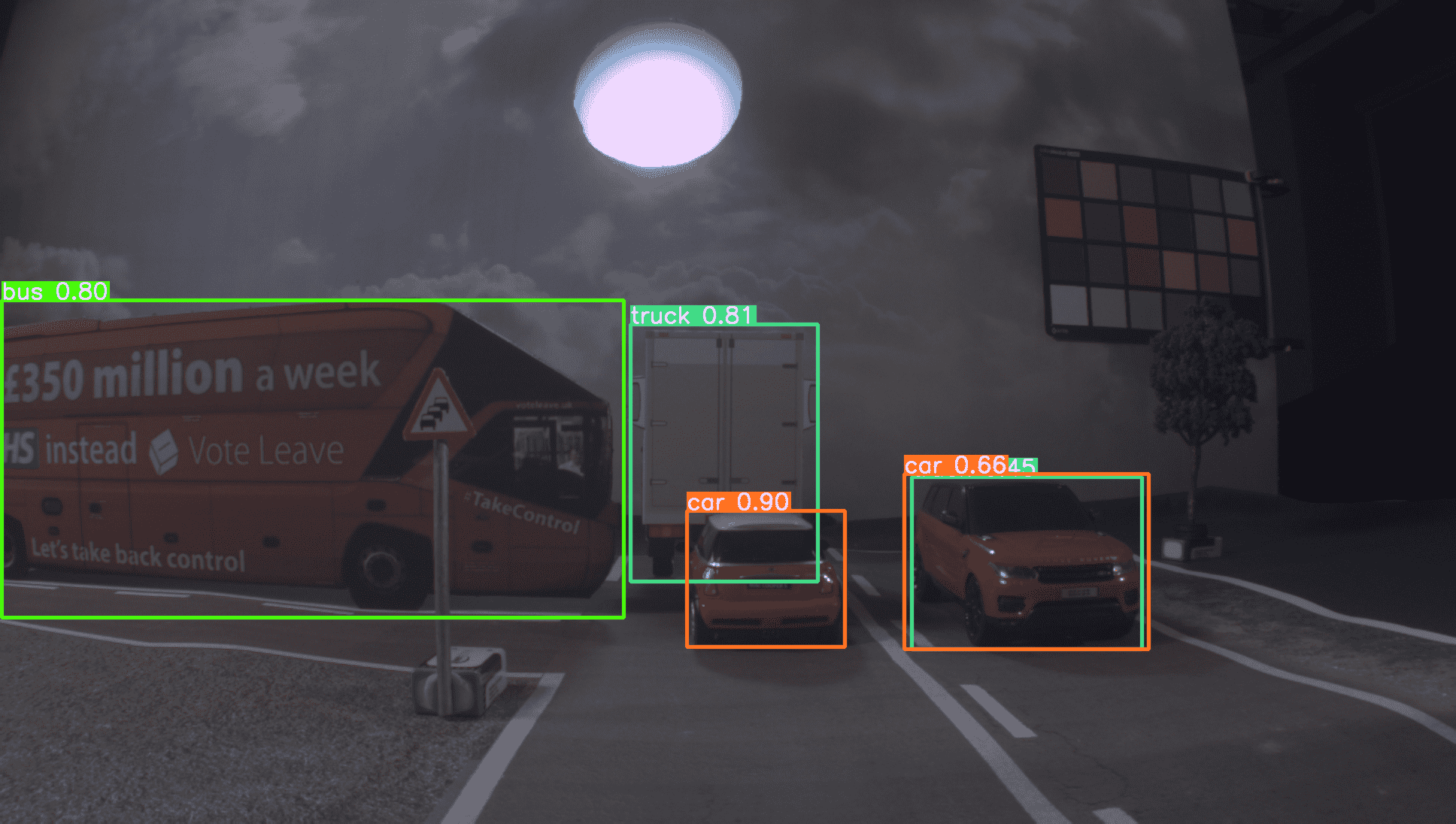}}
\end{subfigure}
\begin{subfigure}{1.3in}
{\includegraphics[width=1.3in,height=1.0in,keepaspectratio]{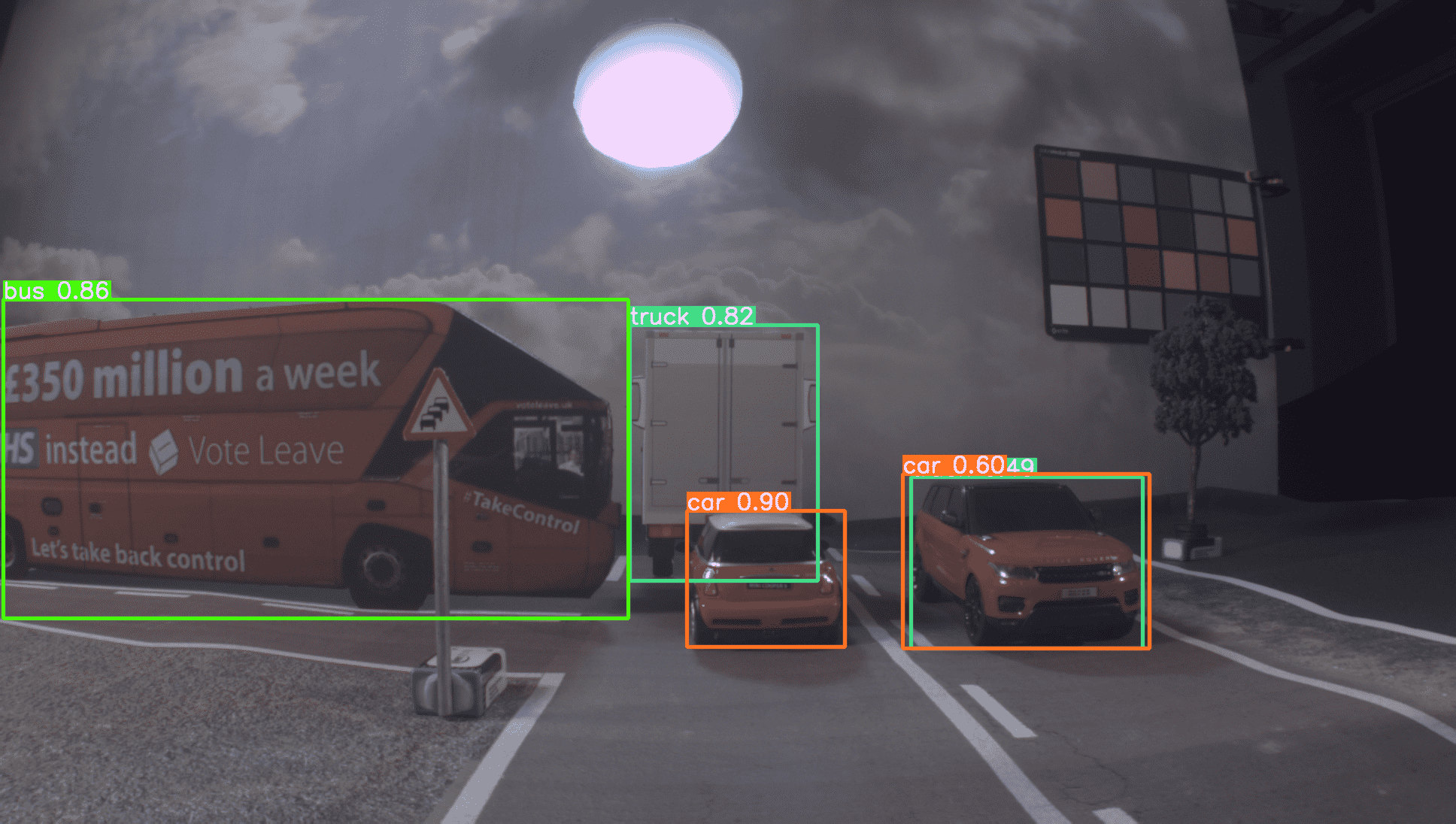}}
\end{subfigure}
\begin{subfigure}{1.3in}
{\includegraphics[width=1.3in,height=1.0in,keepaspectratio]{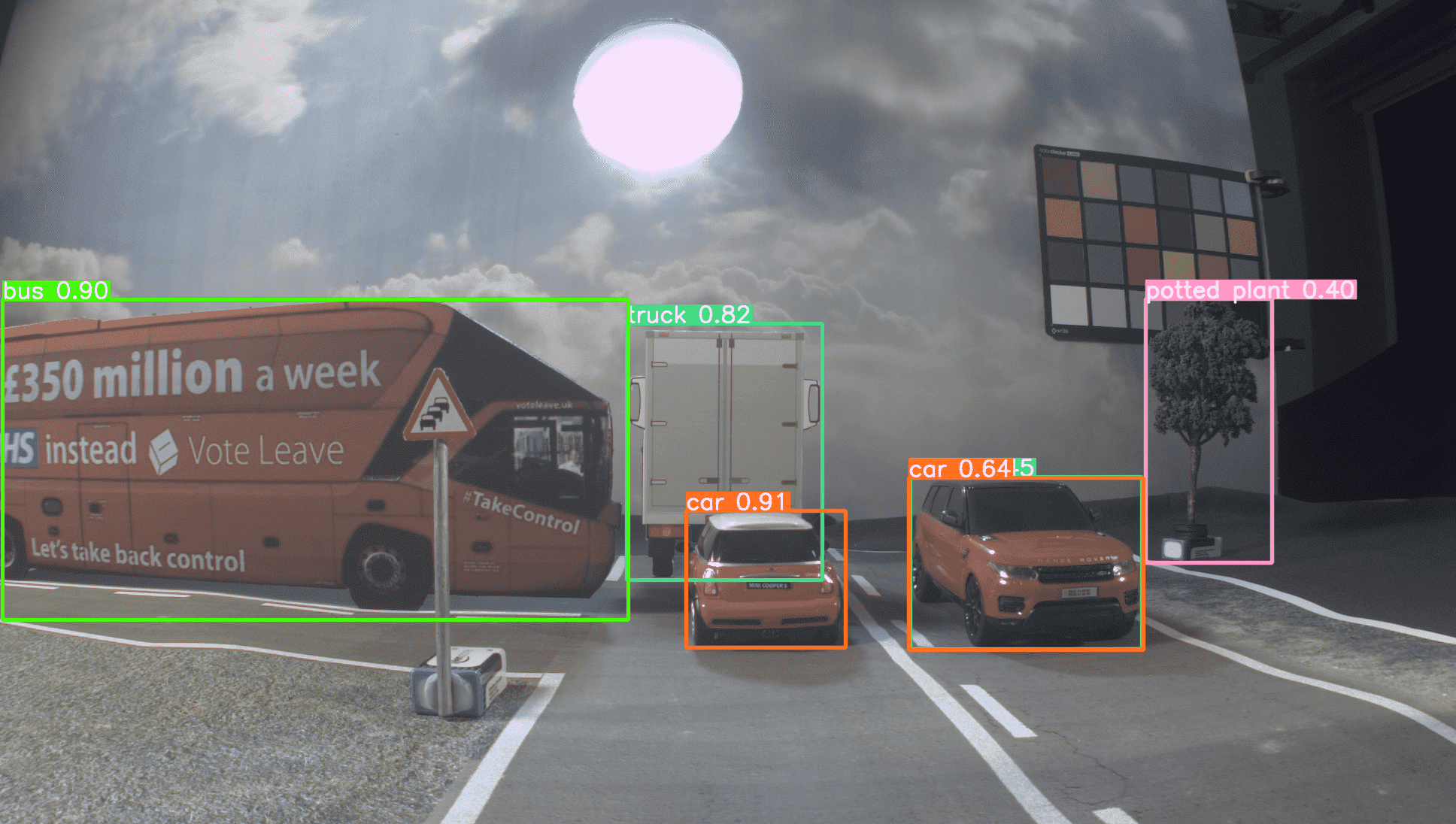}}
\end{subfigure}
\begin{subfigure}{1.3in}
{\includegraphics[width=1.3in,height=1.0in,keepaspectratio]{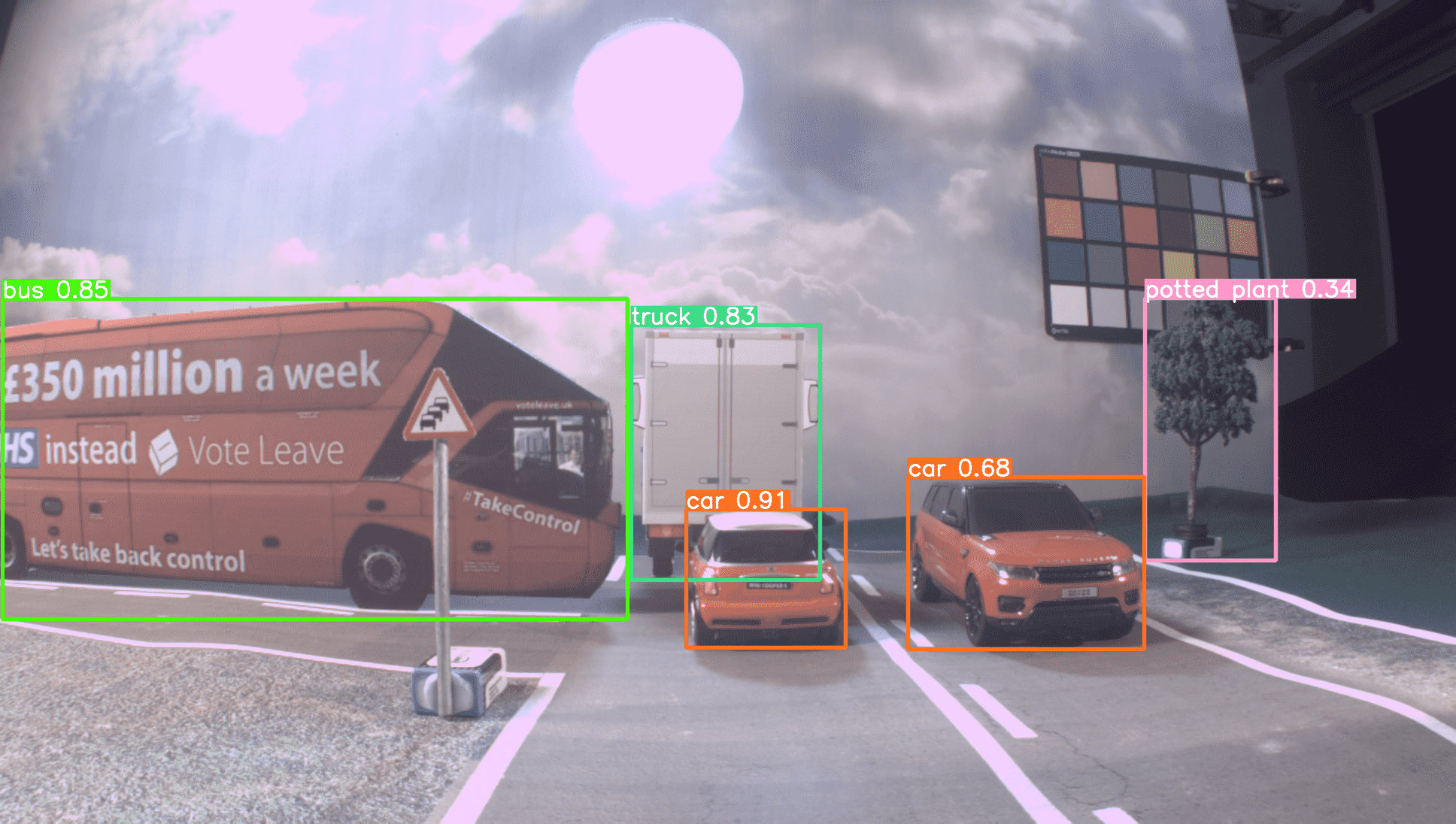}}
\end{subfigure}
\begin{subfigure}{1.3in}
{\includegraphics[width=1.3in,height=1.0in,keepaspectratio]{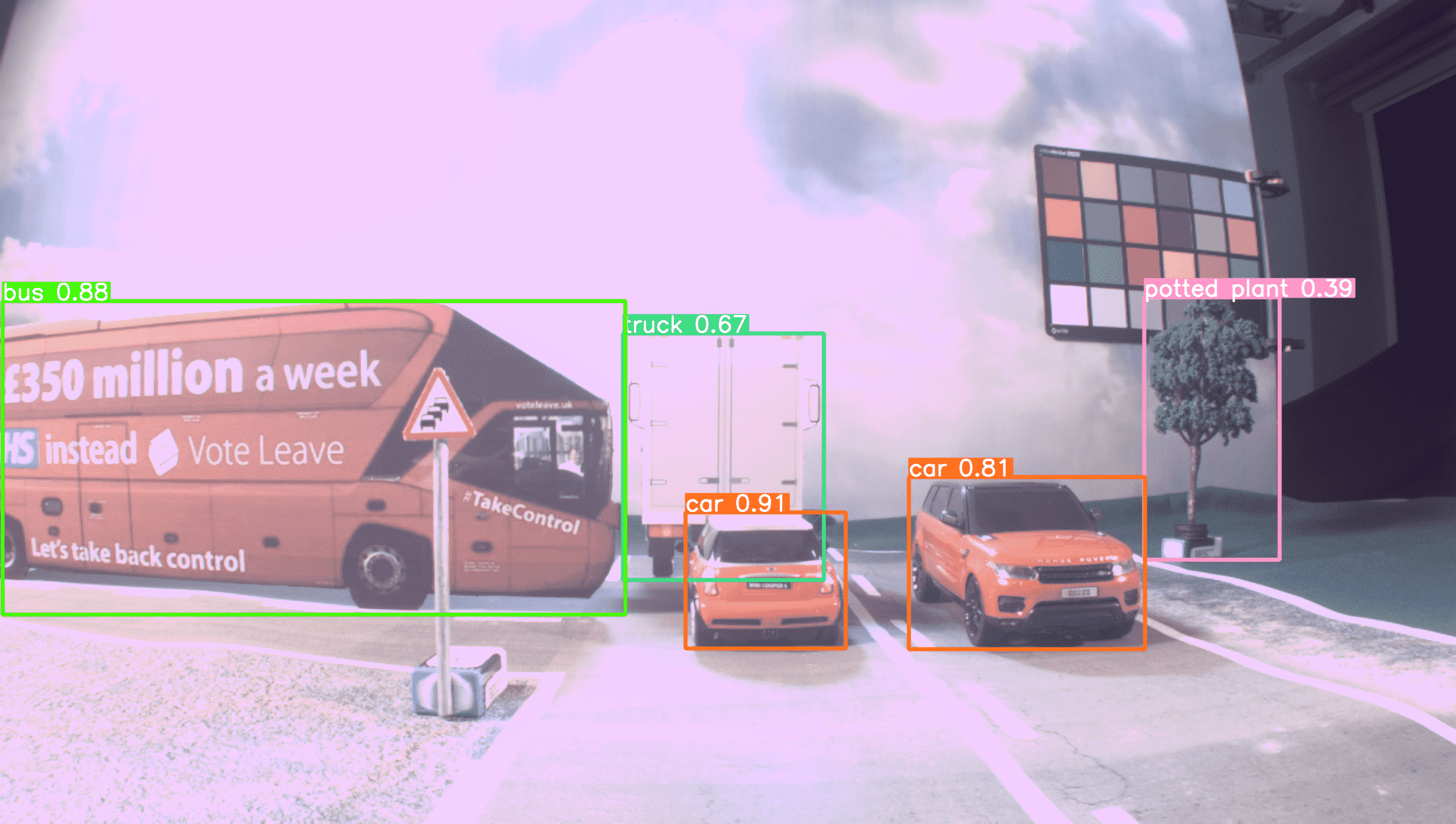}}
\end{subfigure}\\
\raisebox{.7\height}{ \rotatebox[origin=]{90}{Linear}}
\begin{subfigure}{1.3in}
{\includegraphics[width=1.3in,height=1.0in,keepaspectratio]{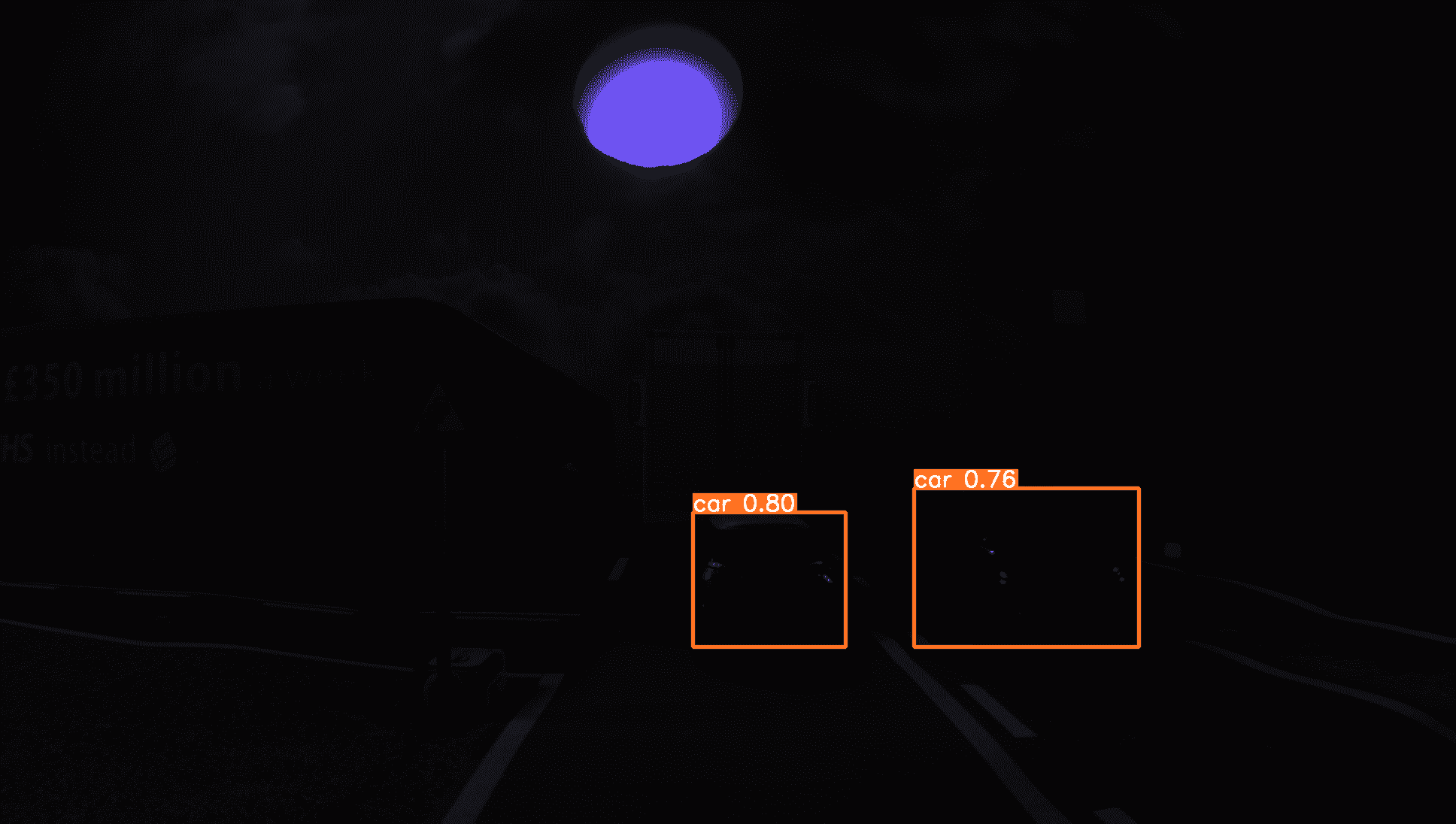}}
\end{subfigure}
\begin{subfigure}{1.3in}
{\includegraphics[width=1.3in,height=1.0in,keepaspectratio]{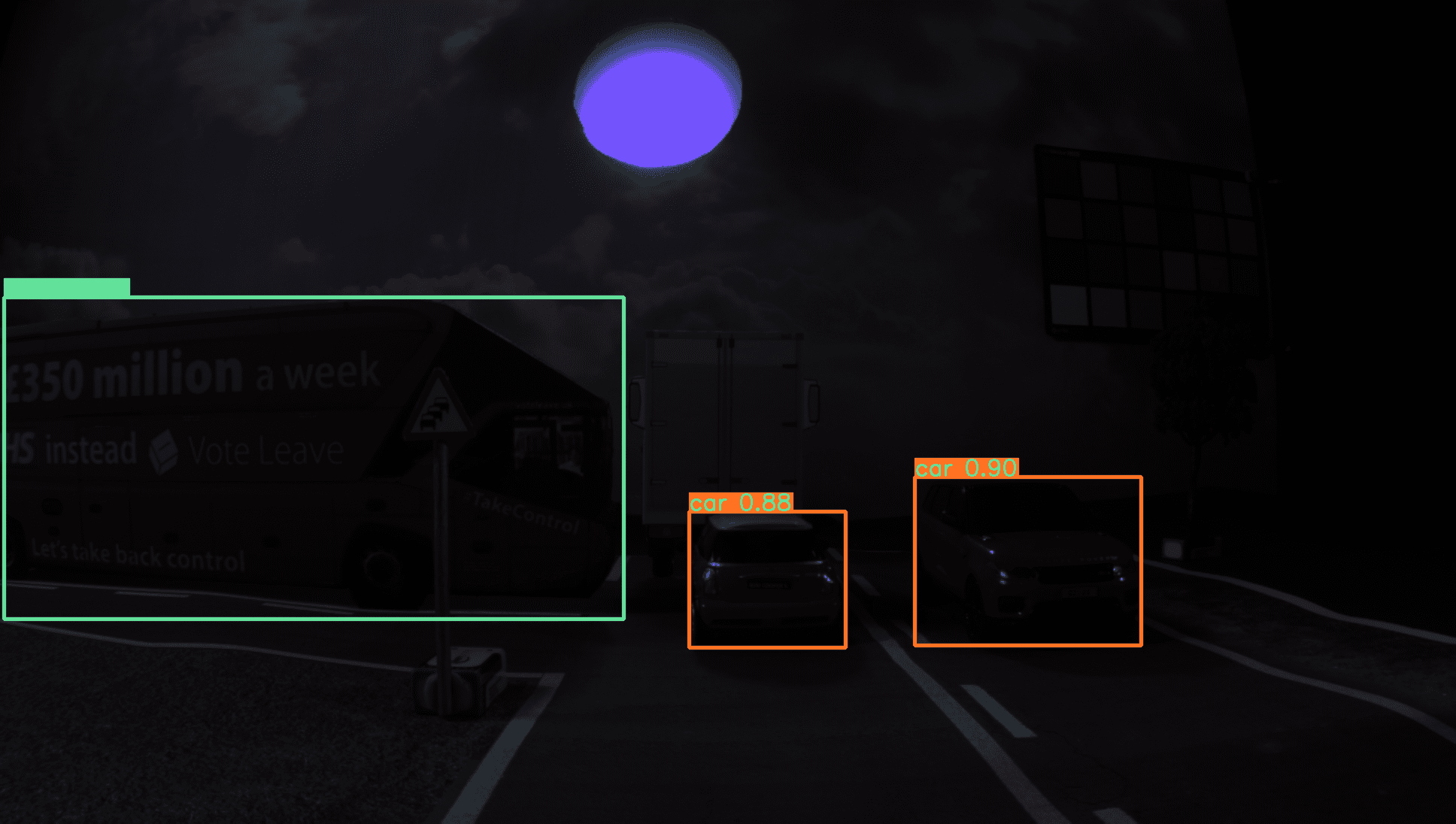}}
\end{subfigure}
\begin{subfigure}{1.3in}
{\includegraphics[width=1.3in,height=1.0in,keepaspectratio]{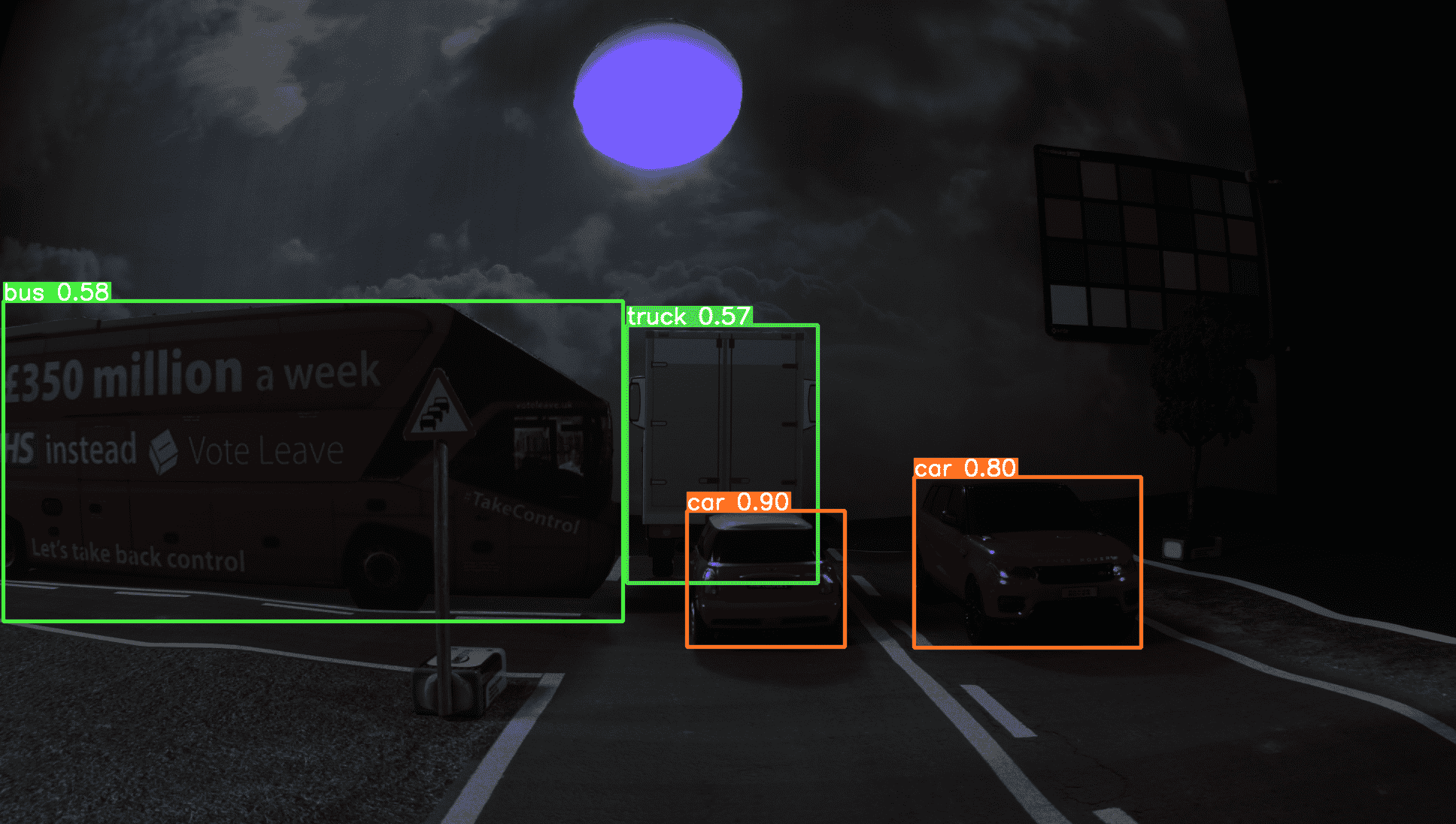}}
\end{subfigure}
\begin{subfigure}{1.3in}
{\includegraphics[width=1.3in,height=1.0in,keepaspectratio]{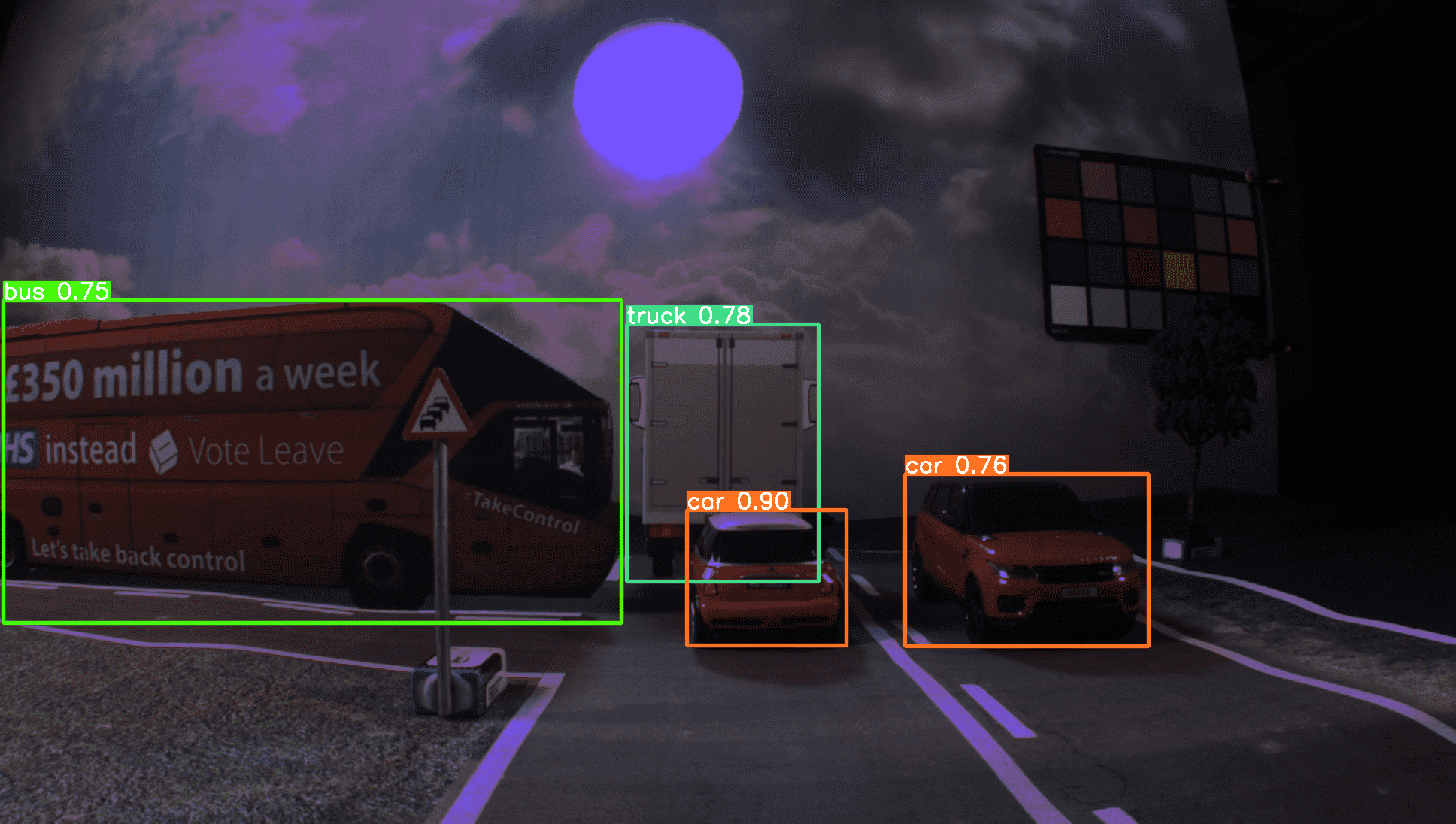}}
\end{subfigure}
\begin{subfigure}{1.3in}
{\includegraphics[width=1.3in,height=1.0in,keepaspectratio]{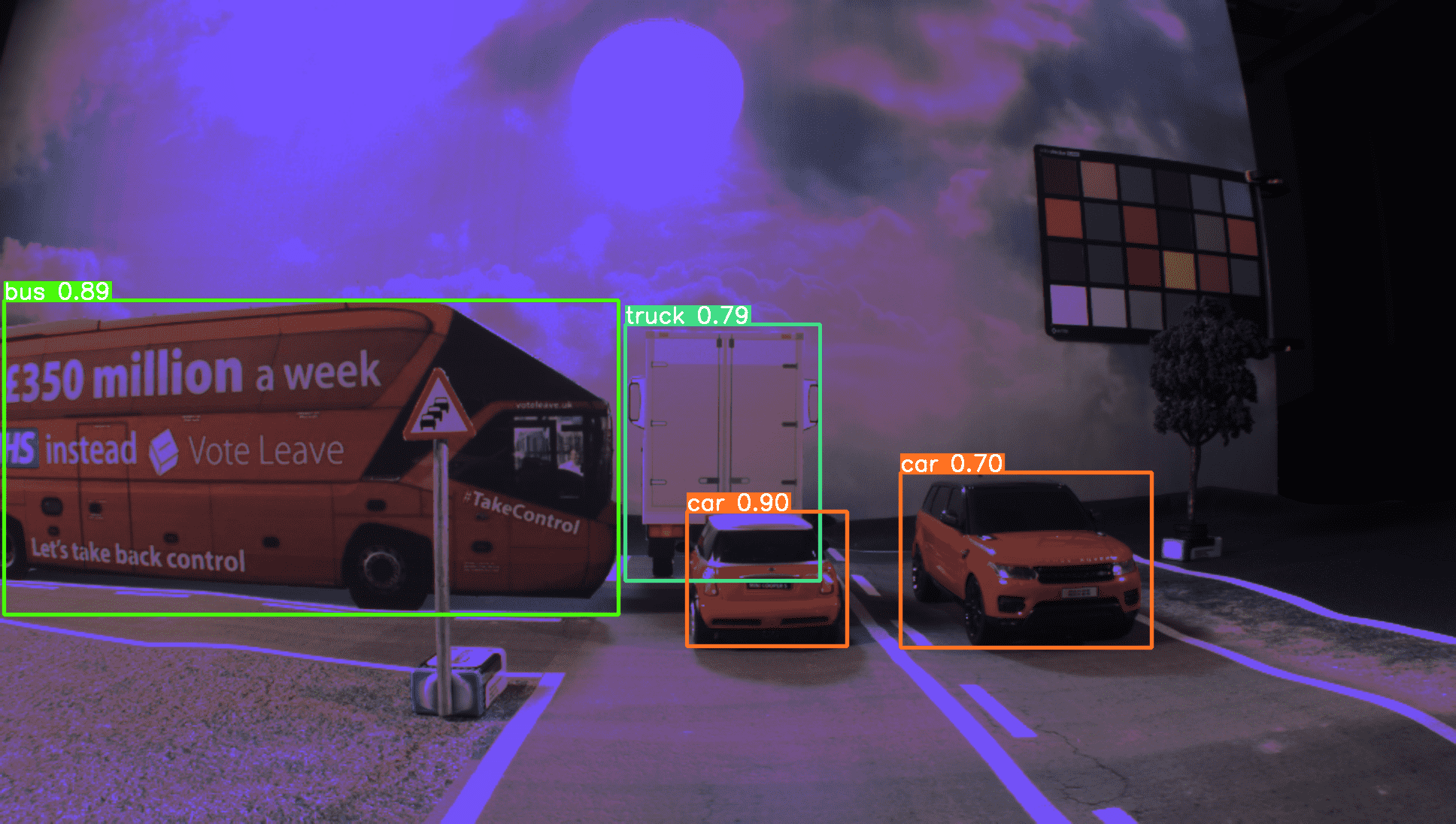}}
\end{subfigure}\\
\hspace{-0.2cm}
\raisebox{.7\height}{ \rotatebox[origin=]{90}{Log}}
\begin{subfigure}{1.3in}
{\includegraphics[width=1.3in,height=1.0in,keepaspectratio]{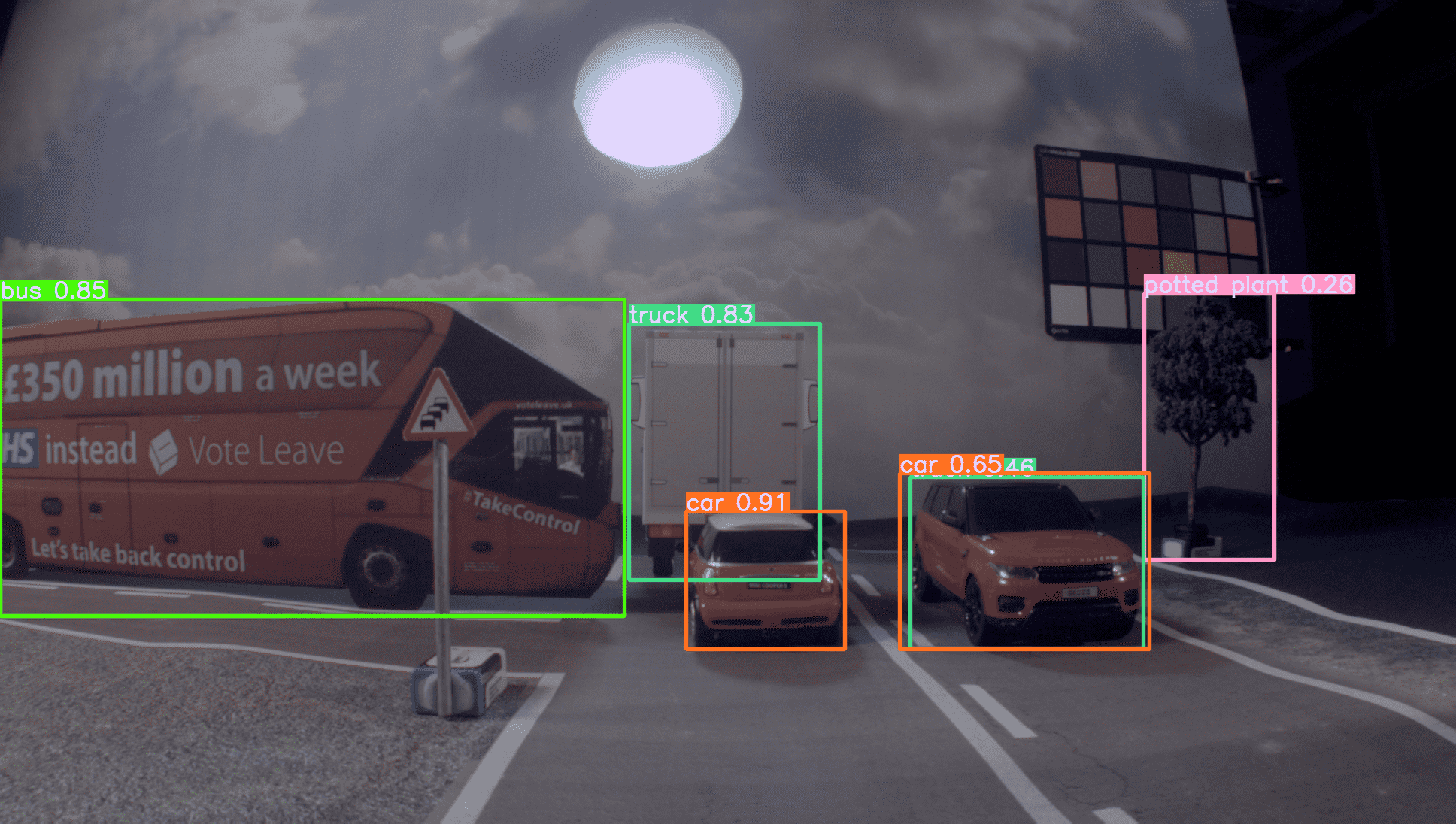}}
\end{subfigure}
\begin{subfigure}{1.3in}
{\includegraphics[width=1.3in,height=1.0in,keepaspectratio]{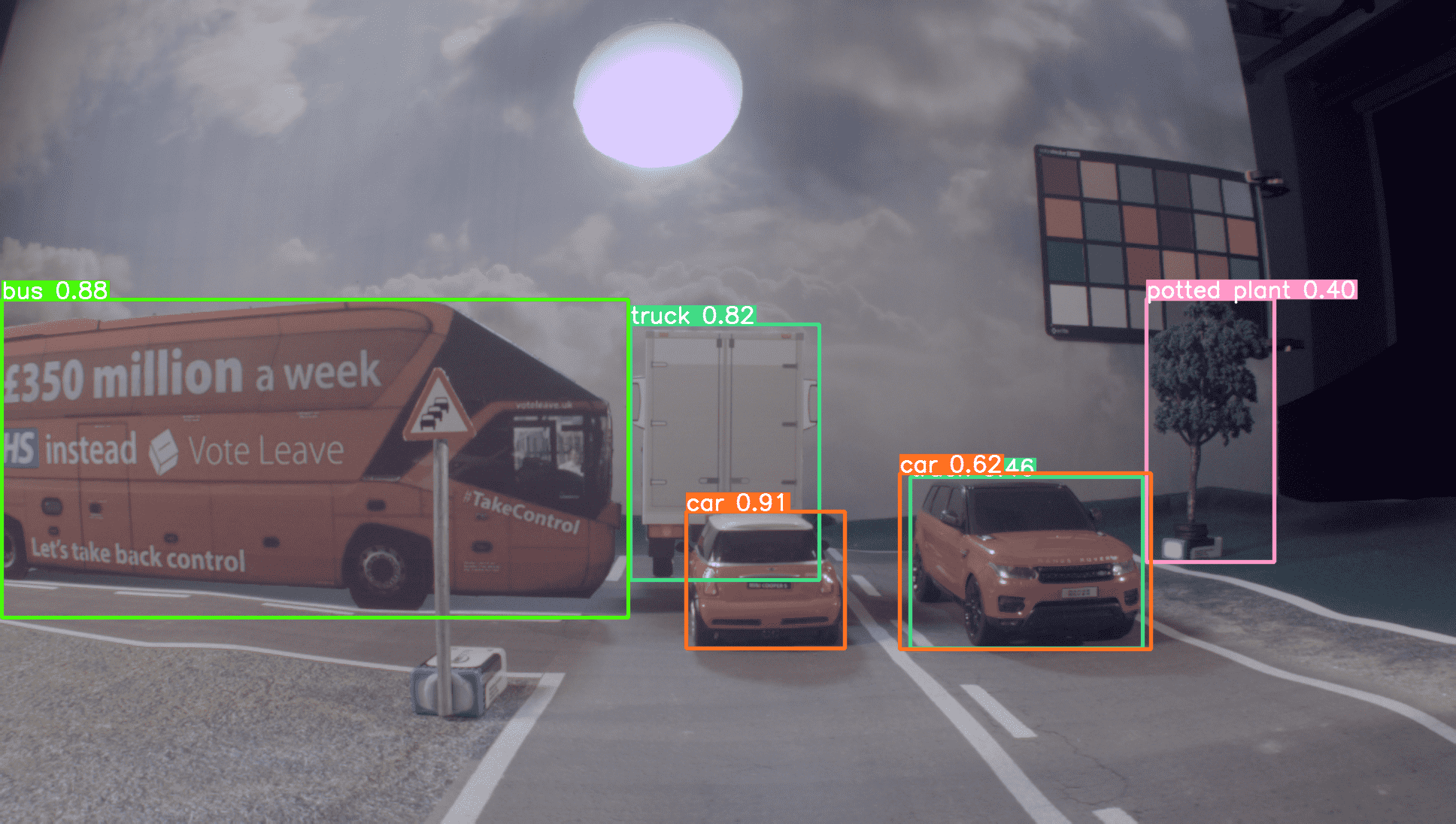}}
\end{subfigure}
\begin{subfigure}{1.3in}
{\includegraphics[width=1.3in,height=1.0in,keepaspectratio]{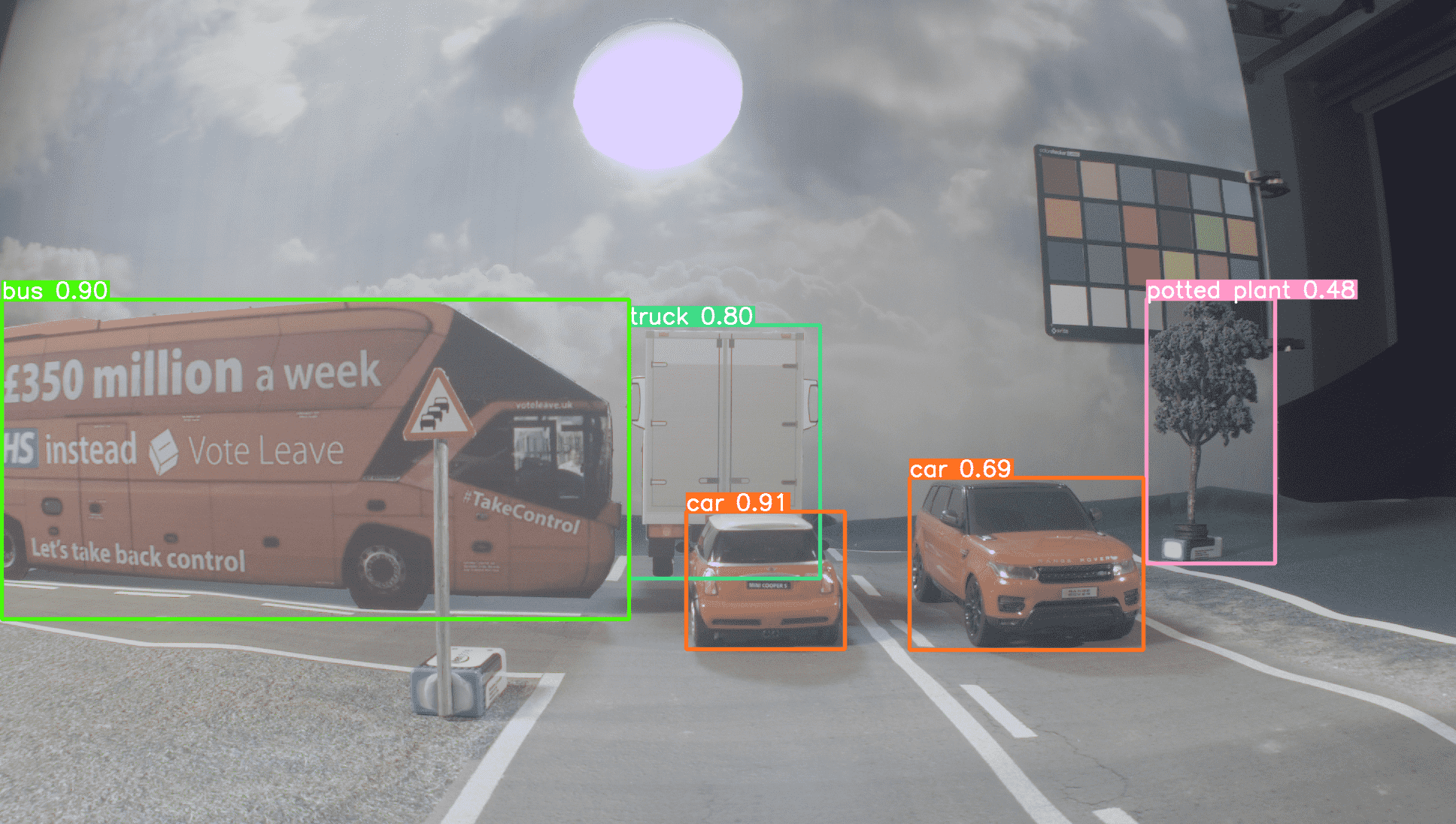}}
\end{subfigure}
\begin{subfigure}{1.3in}
{\includegraphics[width=1.3in,height=1.0in,keepaspectratio]{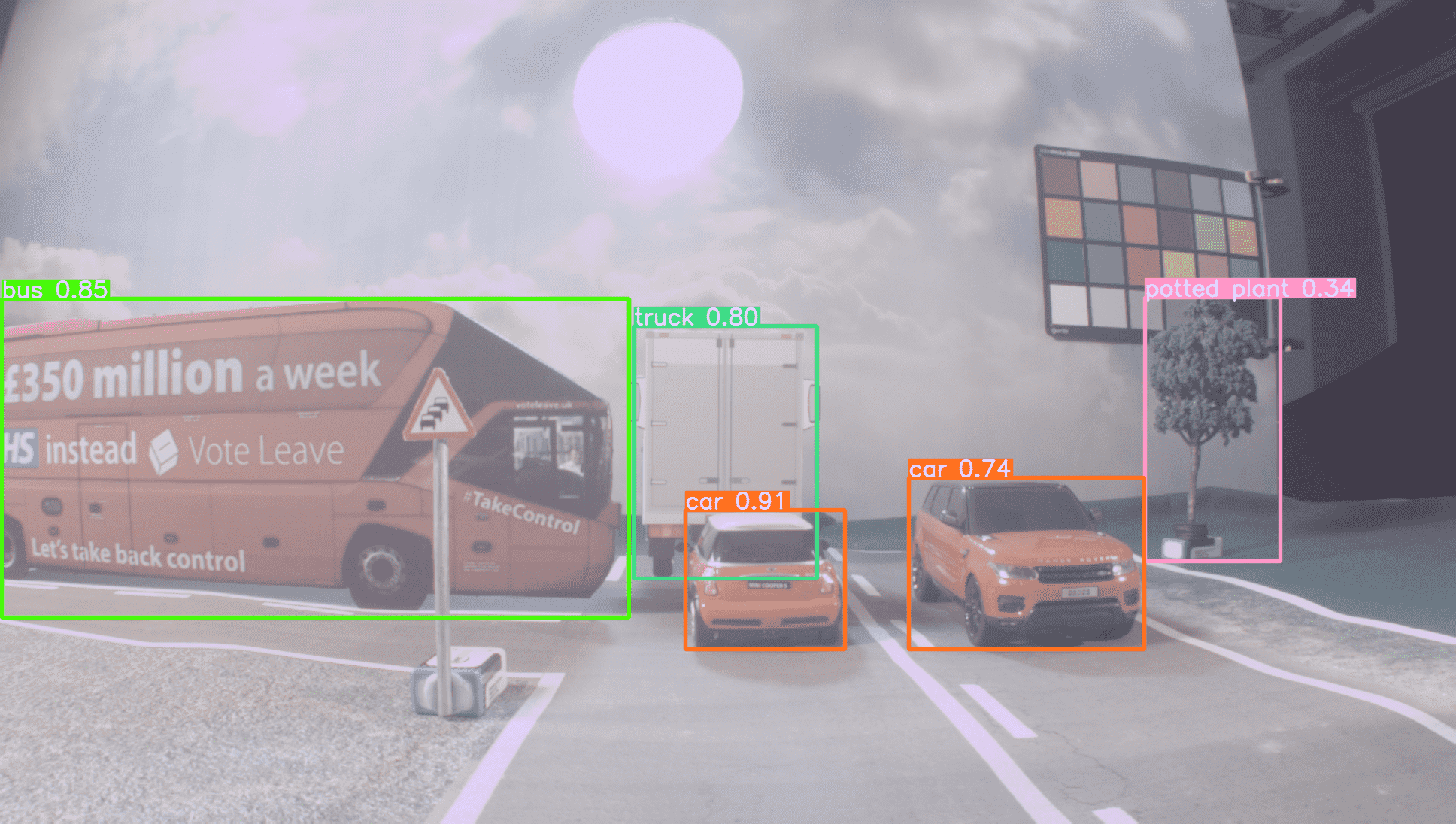}}
\end{subfigure}
\begin{subfigure}{1.3in}
{\includegraphics[width=1.3in,height=1.0in,keepaspectratio]{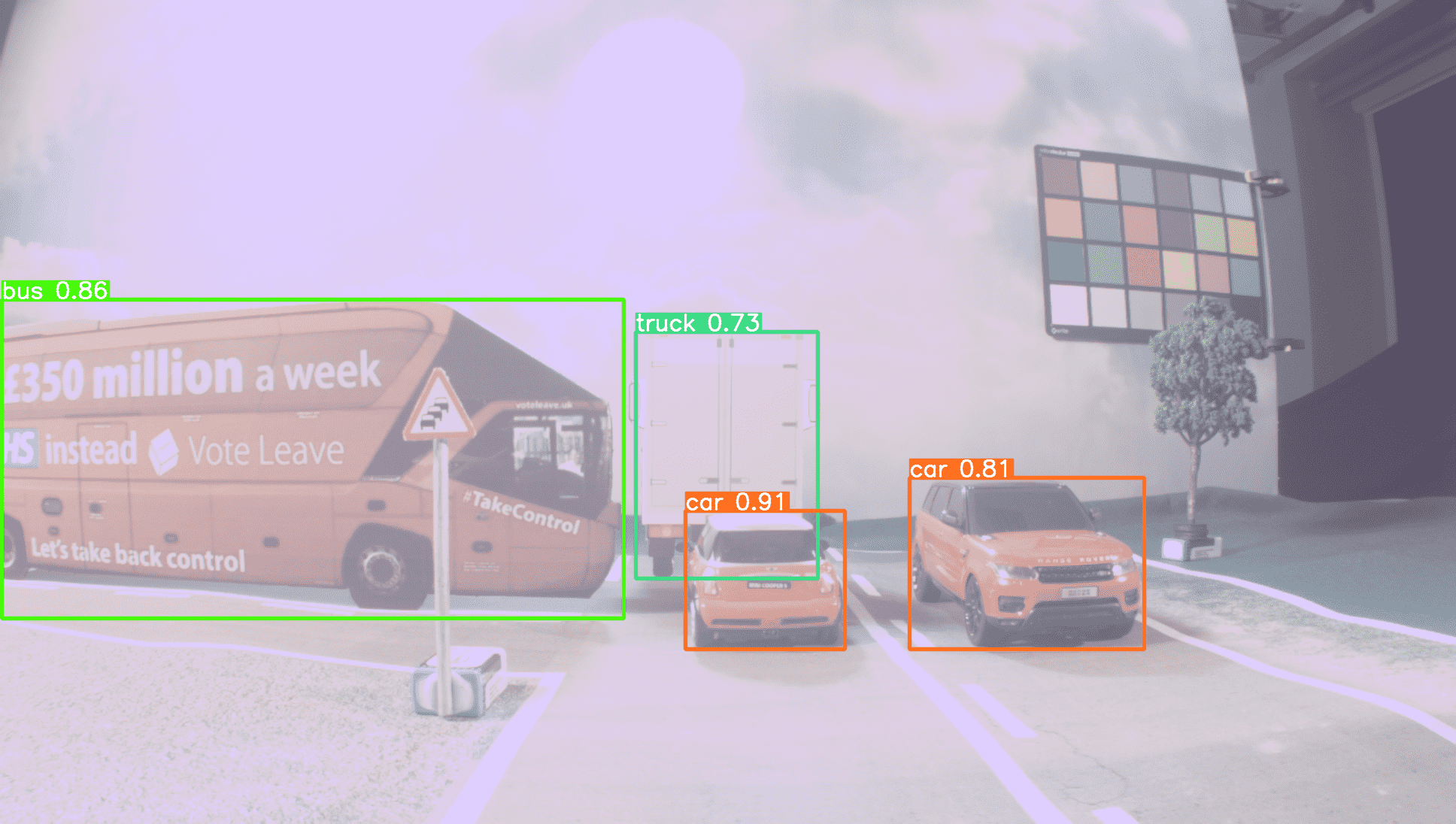}}
\end{subfigure}\\
\caption{Sample output images of object detection algorithm used to produce the results presented in Fig. \ref{fig:TF} plots. Exposure time increases from left to right matching exposure in Fig. \ref{fig:TF}.}
\label{fig:TFVIZ}
\end{figure*}
For a comprehensive assessment of our glare reduction technique, it's imperative to implement testing on a dataset procured from an actual AV, thereby affirming its efficiency across diverse scenarios. Figures \ref{fig:Cv_comp_real}, \ref{fig:Cv_compviz_real}, and Table \ref{real_data_summary}, offer both quantitative and qualitative comparisons between our method and existing ones. This comparative analysis is confined to select perception tasks, constrained by the availability of ground truth in the original dataset. The choice of tasks depends on the feasibility of manual ground truth annotation. As indicated in Figures \ref{fig:Cv_comp_real} and \ref{fig:Cv_compviz_real}, our method notably surpasses current techniques. However, the observed reduction in performance disparity between our method and existing ones, in both controlled experimental setup (Fig. \ref{fig:Cv_comp}) and real-world autonomous driving scenarios (Fig. \ref{fig:Cv_comp_real}), is attributable to the smaller size and limited glare instances in the real dataset.\\ 
\begin{table*}[!h]
\caption{ Quantitative comparison of glare reduction methods on the selected images presented in Fig. \ref{fig:Cv_compviz_real}, evaluating performance across different CV metrics.}
\begin{tabular}{|l|lllllll|l|}
\hline
\multirow{3}{*}{Mehtods {[}De-glare{]}}            & \multicolumn{7}{l|}{Object detection and Object recognition}                                                                                                                                                          & Lane detection   \\ \cline{2-9} 
                                                   & \multicolumn{7}{l|}{Detection-Confidence score {[}Avg.{]}}                                                                                                                                                            & Detected  points \\ \cline{2-9} 
                                                   & \multicolumn{1}{l|}{Person}   & \multicolumn{1}{l|}{Car}      & \multicolumn{1}{l|}{Bus}      & \multicolumn{1}{l|}{Traffice light} & \multicolumn{1}{l|}{Street sign} & \multicolumn{1}{l|}{Tail-light} & Total      &                  \\ \hline
Input                                              & \multicolumn{1}{l|}{3-83\%}   & \multicolumn{1}{l|}{7-91\%}   & \multicolumn{1}{l|}{1-31\%}   & \multicolumn{1}{l|}{1-77\%}         & \multicolumn{1}{l|}{0-0\%}       & \multicolumn{1}{l|}{3-81\%}     & 15-60.5\%  & 316              \\ \hline
Proposed                                           & \multicolumn{1}{l|}{3-82.6\%} & \multicolumn{1}{l|}{7-91.1\%} & \multicolumn{1}{l|}{2-56.6\%} & \multicolumn{1}{l|}{1-76\%}         & \multicolumn{1}{l|}{1-37.5\%}    & \multicolumn{1}{l|}{3-80.3\%}   & 17-70.68\% & 388              \\ \hline
Laplacian filter                                   & \multicolumn{1}{l|}{1-27.6\%} & \multicolumn{1}{l|}{6-77.8\%} & \multicolumn{1}{l|}{3-82.3\%} & \multicolumn{1}{l|}{0-0\%}          & \multicolumn{1}{l|}{0-0\%}       & \multicolumn{1}{l|}{3-78.6\%}   & 13-58.55\% & 12               \\ \hline
Reflection removal                                 & \multicolumn{1}{l|}{1-29.3\%} & \multicolumn{1}{l|}{6-77.7\%} & \multicolumn{1}{l|}{2-58.6\%} & \multicolumn{1}{l|}{1-75\%}         & \multicolumn{1}{l|}{0-0\%}       & \multicolumn{1}{l|}{3-78.3\%}   & 13-53.15\% & 313              \\ \hline
PFFNet \textbackslash{}cite\{mei2018pffn\}         & \multicolumn{1}{l|}{2-54\%}   & \multicolumn{1}{l|}{6-78\%}   & \multicolumn{1}{l|}{2-56.6\%} & \multicolumn{1}{l|}{0-0\%}          & \multicolumn{1}{l|}{1-38\%}      & \multicolumn{1}{l|}{3-80.6\%}   & 14-51.2\%  & 193              \\ \hline
C2PNet \textbackslash{}cite\{zheng2023curricular\} & \multicolumn{1}{l|}{2-56.6\%} & \multicolumn{1}{l|}{7-91.5\%} & \multicolumn{1}{l|}{2-56.6\%} & \multicolumn{1}{l|}{1-77\%}         & \multicolumn{1}{l|}{1-37.5\%}    & \multicolumn{1}{l|}{3-80.3\%}   & 16-66.58\% & 313              \\ \hline
Unsharp masking                                    & \multicolumn{1}{l|}{2-58\%}   & \multicolumn{1}{l|}{7-91\%}   & \multicolumn{1}{l|}{2-57\%}   & \multicolumn{1}{l|}{0-0\%}          & \multicolumn{1}{l|}{0-0\%}       & \multicolumn{1}{l|}{3-80.3\%}   & 14-47.71\% & 313              \\ \hline
Deblur                                             & \multicolumn{1}{l|}{3-83.3\%} & \multicolumn{1}{l|}{7-91\%}   & \multicolumn{1}{l|}{1-31\%}   & \multicolumn{1}{l|}{1-78\%}         & \multicolumn{1}{l|}{0-0\%}       & \multicolumn{1}{l|}{3-80\%}     & 15-60.55\% & 290              \\ \hline
Wiener filter                                      & \multicolumn{1}{l|}{3-83\%}   & \multicolumn{1}{l|}{7-91.1\%} & \multicolumn{1}{l|}{1-31\%}   & \multicolumn{1}{l|}{1-76\%}         & \multicolumn{1}{l|}{0-0\%}       & \multicolumn{1}{l|}{3-81\%}     & 15-60.35\% & 290              \\ \hline
\end{tabular}
\label{real_data_summary}
\end{table*}
\color{black}
\vspace{-.5cm}
\section{Conclusions}
In this paper, we introduce a glare reduction technique designed to improve perception in autonomous vehicles (AV). This approach utilizes a combined GSF, derived through offline calibration, to estimate true radiance in glare-affected saturated pixels. It addresses two main challenges of deconvolution-based methods: accurate radiance prediction in saturated areas and application restriction only to a singular data source due to GSF limitations. The paper also examines the impact of glare on various autonomous vehicle perception tasks, emphasizing the importance of glare reduction. The evaluation demonstrates that the proposed method surpasses the best performing glare mitigation strategy on a real AV's captured dataset in various perception tasks. Specifically, it shows superior performance in object detection by 5.15\%, object recognition by 18.16\%, and lane detection by 1.03\%, which represent an average improvement of 8.11\%. 
%
%
\bibliographystyle{IEEEbib}
\bibliography{strings.bib}
\end{document}